\documentclass[manuscript,screen]{acmart}

\renewcommand\footnotetextcopyrightpermission[1]{} 
\usepackage{color}
\usepackage{amsmath}
\usepackage{amsfonts}
\usepackage{booktabs, multicol, multirow}
\usepackage{fontawesome5}
\usepackage{xcolor}
\usepackage{threeparttable}
\usepackage{relsize}
\usepackage{makecell}
\usepackage{verbatim}
\usepackage{diagbox}
\usepackage{subfigure}
\usepackage{bm}
\usepackage{mathtools}
\AtBeginDocument{%
  \providecommand\BibTeX{{%
    \normalfont B\kern-0.5em{\scshape i\kern-0.25em b}\kern-0.8em\TeX}}}

\acmSubmissionID{CSUR-2022-0640}

\begin{document}

\title{A Survey of Robustness and Safety of 2D and 3D Deep Learning Models Against Adversarial Attacks}
\author{Yanjie Li}
\email{yanjie.li@connect.polyu.hk}
\orcid{0000-0001-8859-8331}
\affiliation{%
  \institution{The Hong Kong Polytechnic University}
  \streetaddress{11, Yuk Choi Road, Hung Hom, KLN}
  \country{Hong Kong}
}

\author{Bin Xie}
\email{xiebin.sc@gmail.com}
\orcid{0000-0001-5118-3570}
\affiliation{%
  \institution{The Hong Kong Polytechnic University}
  \streetaddress{11, Yuk Choi Road, Hung Hom, KLN}
  \country{Hong Kong}
}

\author{Songtao Guo}
\email{guosongtao@cqu.edu.cn}
\affiliation{%
  \institution{Chongqing University}
  \streetaddress{Chongqing}
  \country{China}
}

\author{Yuanyuan Yang}
\email{yuanyuan.yang@stonybrook.edu}
\orcid{0000-0001-7296-9222}
\affiliation{%
  \institution{Stony Brook University}
  \streetaddress{Stony Brook, NY}
  \country{USA}
}

\author{Bin Xiao}
\email{csbxiao@comp.polyu.edu.hk}
\orcid{0000-0003-4223-8220}
\authornote{Corresponding author}
\affiliation{
  \institution{The Hong Kong Polytechnic University}
  \streetaddress{11, Yuk Choi Road, Hung Hom, KLN}
  \country{Hong Kong}
}

\renewcommand{\shortauthors}{Yanjie Li, et al.}

\begin{abstract}
Benefiting from the rapid development of deep learning, 2D and 3D computer vision applications are deployed in many safe-critical systems, such as autopilot and identity authentication. However, deep learning models are not trustworthy enough because of their limited robustness against adversarial attacks. The physically realizable adversarial attacks further pose fatal threats to the application and human safety. Lots of papers have emerged to investigate the robustness and safety of deep learning models against adversarial attacks. To lead to trustworthy AI, we first construct a general threat model from different perspectives and then comprehensively review the latest progress of both 2D and 3D adversarial attacks. We extend the concept of adversarial examples beyond imperceptive perturbations and collate over 170 papers to give an overview of deep learning model robustness against various adversarial attacks. To the best of our knowledge, we are the first to systematically investigate adversarial attacks for 3D models, a flourishing field applied to many real-world applications. In addition, we examine physical adversarial attacks that lead to safety violations. Last but not least, we summarize present popular topics, give insights on challenges, and shed light on future research on trustworthy AI.

\end{abstract}

\ccsdesc[500]{Security and privacy}
\ccsdesc[500]{Computing methodologies~Machine learning}
\ccsdesc[500]{Computing methodologies~Computer vision}

\keywords{Deep learning; 3D computer vision; Adversarial attack; Robustness;}

\maketitle

\section{Introduction}
    The significant strides of deep learning (DL) algorithms have driven considerable technological progress in computer vision (CV) tasks, which are widely deployed in various safety-critical and mission-critical systems like identity authentication and self-driving vehicles. These applications depend on the assumption that these deep learning models are \textit{trustworthy} and robust against small perturbations, which means these models can produce consistent predictions when noise exists. However, studies have shown that deep-learning models are vulnerable to \textit{adversarial examples} (AEs), which makes the deep-learning models produce false predictions by crafting elaborate imperceptive or semantic-preserving perturbations. To realize trustworthy AI, continuous efforts have been spent on improving the model's robustness and safety against adversarial attacks and finding these models' robustness upper bound by constructing stronger adversarial attacks. This is a relentless race about adversarial attacks and defenses. This paper thoroughly surveys the latest progress of this adversarial attack and defense competition from the adversary's perspective.

    This paper primarily discusses the phenomenon of adversarial attacks in the realm of computer vision tasks. While most computer vision tasks tend to focus on image processing, recent attention has been directed toward 3D tasks. 3D data can supplement 2D data by providing depth information about an image, thereby allowing for a more reliable and detailed analysis of targets, and can be utilized in various applications. For instance, autonomous vehicles often rely on a combination of cameras and lidar to perceive their surroundings. However, due to the unordered nature of 3D data, direct application of 2D adversarial attacks is not feasible. 3D adversarial attacks have been proposed, building upon the principles of 2D adversarial attacks but with specific designs tailored to the characteristics of 3D data. For example, 3DAdv \cite{xiao2018generating} is based on the C\&W attack but with novel distance metrics. 3Dhacker \cite{tao20233dhacker} is based on the boundary attack but fuses the point cloud in the spectral domain rather than via coordinate-wise average operation. Due to the close relationship between 2D and 3D computer vision tasks and the shared theoretical foundation of 2D and 3D attacks, this survey summarizes the latest progress of adversarial attacks in both 2D and 3D computer vision.

    The aim of this survey is to systematize the latest progress of 2D and 3D adversarial attacks to help researchers construct stronger AEs to evaluate model robustness, design robust models, and ensure safety in real-world applications. To select literature for systematic review, we first clarify the review scope and thoroughly search publications in top computer vision or security conferences and journals within the past several years. Then, we select high-cited or representative work related to 2D and 3D adversarial attacks. These articles are categorized and compared based on the targets, scenarios, and methods for comprehensiveness. The main contributions of this work are as follows:
    \begin{itemize}
      \item We summarize the latest studies on the adversarial robustness and safety of 2D and 3D deep learning models against adversarial attacks. Over 170 papers in recent years have been collated and compared. Moreover, We divide them into 2D and 3D attacks according to the data characteristics and application scenarios.
      \item  For 2D adversarial attacks, we extend the meaning of adversarial attack from imperceptible perturbation to semantic-preserving perturbation, such as color space distortion and spatial transformation distortion. We classify these attacks according to methodologies and compare their pros and cons to give a full-scope view. 
      \item  3D data is increasingly leveraged in safety-critical fields like self-driving. However, the 3D model robustness is hardly reviewed. To help design robust 3D deep learning models, we are the first to organize 3D adversarial attacks systematically and classify them according to their algorithms.
      \item When deep learning models are deployed in the real world, safety is the primary premise. We comprehensively examine related works of adversarial attacks for safe-critical missions, especially for camera-based and Lidar-based self-driving cars and face recognition.
      \item At the end of this review, we summarize the present hot research topics into several points and identify their challenges, such as improving the attack transferability, generating semantic perturbation, and evaluating the robustness of 3D deep learning models. We also provide some useful advice for future research directions.
    \end{itemize}
    
\begin{figure}[h]
      \centering
      \includegraphics[scale=0.49]{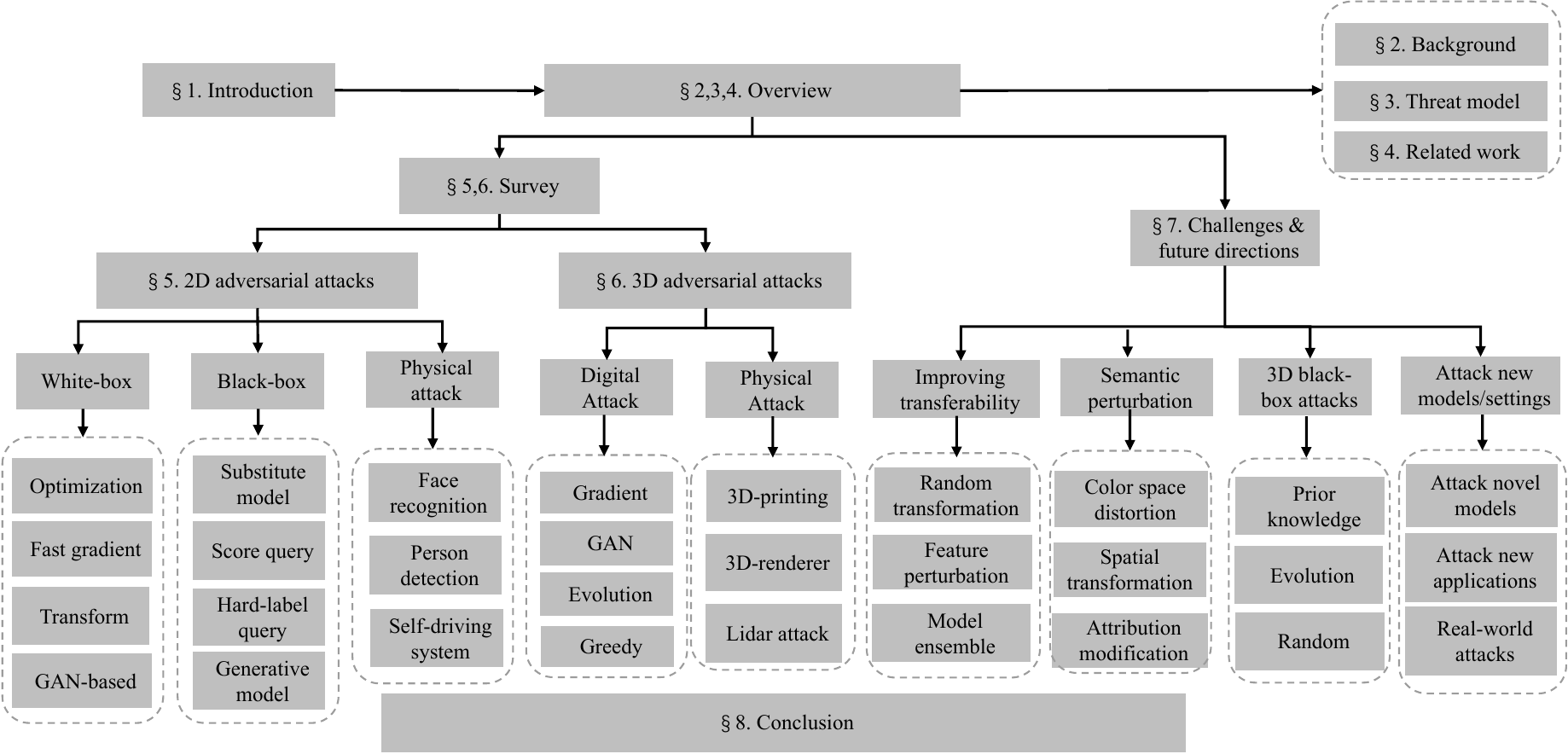}
      \caption{The structure of this survey}
      \label{figure_structure}
      \Description{The structure of this paper.}
    \end{figure}

    The structure of this survey is shown in Figure.\ref{figure_structure}. In Sec.\ref{background}, we clarify the basic concepts of deep learning and computer vision. In Sec.\ref{threat model}, we build a general threat model for deep-learning-based computer vision systems by examining the attack surface from different perspectives. In Sec.\ref{related work}, we summarize the latest related reviews. In Sec.\ref{2D_attack} and \ref{3D attack}, we summarize the representative works that evaluate the robustness and safety of deep learning models by 2D and 3D adversarial examples, respectively. In each section, we first introduce our taxonomy based on attack methodology or target applications, then summarize the representative digital-world and physical-world adversarial attacks. Finally, in Sec.\ref{Sec_challenge}, we identify the present obstacles in the adversarial attacks, coupled with some viewpoints for future studies.
    
    To unify the symbols of different articles, Table.\ref{tab: notation and abbr} shows some common notations in this survey.
    
    \begin{table}[htb]
      \caption{Common notations used in this survey}
      \label{tab: notation and abbr}
      \begin{tabular}{llll}
        \toprule
        Notations & Description & Notations & Description\\
        \midrule
        $D_{tr}, D_{te}$ & Training\verb|\|testing dataset & $x$ & Normal inputs \\
        $x',x^{adv}$ & Modified\verb|\|adversarial inputs &  $\delta$ & Adversarial perturbations \\
        $y$ & Ground truth labels &   $y'$ & Adversarial target labels\\
        $\mathcal{F}$ &Classification model & $\hat{\mathcal{F}}$ & Substitute model \\        
        $g$ & The model's gradient upon $x$ &  $\hat{g}$ & The estimated model gradient  \\
        $l_p$ & The $p$-norm distance & $Z_i(x)$ & Output logits of $i^{th}$-to-last layer\\
        $\theta$ & Model parameters  & $\mathcal{L}(x',y)$ & Loss function\\
        $\mathcal{G}$ & The GAN's generator & $\mathcal{D}$ & The GAN's discriminator \\
        $\mathcal{P}$ & Original point cloud  & $\mathcal{P'}$ & Adversarial point cloud\\   
        \bottomrule
      \end{tabular}
    \end{table}

     \section{Background}
     \label{background}

        Deep learning is a popular representation learning algorithm for its outstanding performance on image classification and other tasks. It can learn complex functions through a composition of superficial but non-linear layers. 
        Suppose $x$ is an image or a point cloud, $\mathcal{F}_\theta$ is a deep learning model with model parameters $\theta$. The goal of the object classification task is to find $\theta$ that can minimize the difference between the ground truth label $y$ and the prediction $\mathcal{F}_{\theta}(x)$, that is 
            \begin{equation}
            \label{eq1}
            arg \min_\theta \mathcal{L}(\mathcal{F}_{\theta}(x), y),
            \end{equation}   
        where $\mathcal{L}$ is a loss function to measure the entropy between the  $\mathcal{F}_{\theta}(x)$ and y. The $\theta$ is usually optimized through gradient descent algorithms, such as Adam. After training, the deep learning model is deployed into real-world applications.           

       Many deep learning model variants have been proposed to boost performance and adapt to different task characteristics, such as multi-layer perceptions, stacked autoencoders, convolutional neural networks, deep brief networks, and vision transformers. These models are widely used in various 2D computer vision tasks, such as object detection, image segmentation, image classification, action recognition, and motion tracking. Recently, with the boom of 3D data in self-driving systems and other applications, deep learning for 3D computer vision has attracted much attention. Various models came out, such as MLP-based models (e.g. PointNet \cite{qi2017pointnet}), convolutional models (e.g. Pointwise-CNN \cite{ha2018pointwise}), graph-based convolutional models (e.g. DGCNN \cite{wang2019dynamic}), and transformer-based models (e.g. PCT \cite{guo2021pct}). The major applications of the deep learning model in 3D computer vision can be categorized into three tasks: 3D object detection and tracking, 3D object classification, and 3D object segment. Because most 3D deep learning models are extensions of 2D models, many security threats in the 2D domain also exist in the 3D domain.

  \section{Related work}
    \label{related work}
    To boost this adversarial machine learning arms race and mitigate the risk of adversarial attacks, some previous surveys have tried to summarize the latest relevant studies. For example, Serban \cite{serban2020adversarial}
    collected adversarial attacks in the object recognition task. However, they did not include 3D adversarial attacks and semantic attacks like color space perturbations. Miller \textit{et al.} \cite{miller2020adversarial} and Machado \textit{et al.} \cite{machado2021adversarial} reviewed the recent progress of adversarial machine learning, but from the defense rather than the adversary perspective. Other reviews regard the adversarial attack as part of the AI attacks without a separate detailed investigation. For example, Liu \textit{et al.} \cite{liu2018survey} classified AI attacks into integrity, availability, and confidentiality attacks according to the traditional taxonomy of security violation. Papernot \textit{et al.} \cite{papernot2018sok} and Liu \textit{et al.} \cite{liu2020privacy} categorized attacks based on the deep learning pipeline, dividing them into training and test phase attacks. He \textit{et al.} \cite{He2022TowardsSecurity} categorized the AI attacks into adversarial attacks, poisoning attacks, model extraction, and inversion attacks. Moreover, these surveys lack summaries of the latest research trends, such as the growing interest in transferable adversarial examples, semantic perturbations, 3D adversarial attacks, and physical-realizable adversarial attacks. In addition, although attack-agnostic robustness evaluation has emerged in recent years, the adversarial attack is still one of the most efficient and reliable ways to evaluate model robustness. There is still a lack of a systematic review to collate these developments and discuss future directions in light of these latest efforts.

    \section{Threat model}
    \label{threat model}
\begin{figure}[t]
                  \centering
                  \includegraphics[scale=0.55]{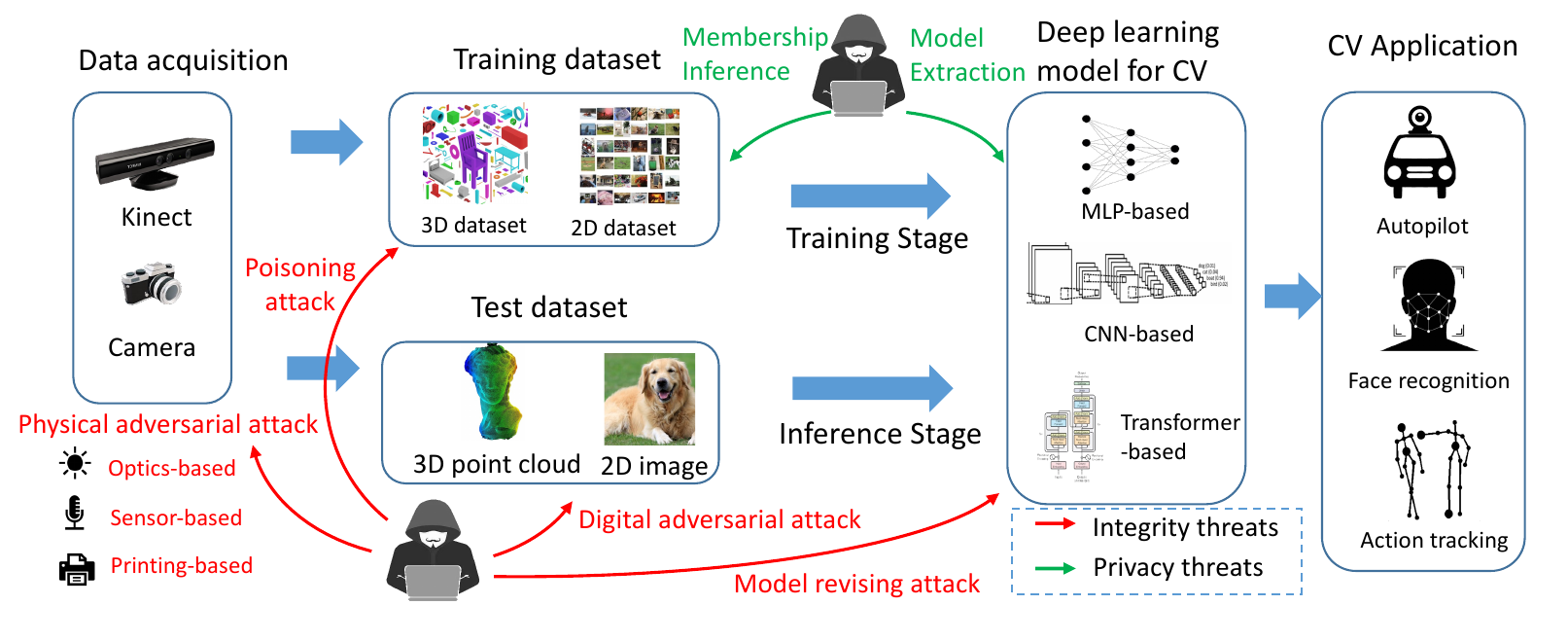}
                  \caption{An overview of the attack surface when deep learning models are deployed into computer vision applications.}
                  \label{attackSurface}
                  \Description{The structure of this paper.}
                \end{figure}

        The threat model evaluates the possible risk and security level of the system by describing the target, capability, and knowledge of the attacker. By establishing a full-scale threat model, we can understand the security problems of the deep learning system more comprehensively. We first identify the complete and thorough attack surface when the deep neural networks are deployed into computer vision tasks, as shown in Figure.\ref{attackSurface}. Then we analyze the adversary's goals, capabilities, and knowledge to construct a threat panorama.
        \subsection{The attack surface of deep learning models}
        The attack surface is composed of all possible vulnerabilities in a system that an unauthorized user can access. As mentioned above, deep learning applications train the model according to the model performance on the training dataset and then deploy the model to real-world applications for various 2D or 3D tasks. In this process, the attack surface includes the training and test dataset and the deep learning model. 
            \subsubsection{Training dataset.} Collecting a training dataset with satisfactory quality requires a great amount of effort and money. Meanwhile, some datasets, like medical diagnostic records and personal genomic data, may contain sensitive information. Therefore, the privacy and confidentiality of these training datasets are essential. \textit{Membership inference attacks} \cite{shokri2017membership} and \textit{data reconstruction attacks} \cite{salem2020updates} can leak the attribute or information of training data. In addition, more capable attackers may be able to modify the training datasets. Under this assumption, \textit{backdoor attacks} and \textit{poisoning attacks} are proposed. The backdoor attacks make the model behave normally for clean samples but behave incorrectly for the same samples with specific stickers by adding triggers in the training process. The poisoning attacks modify the sample's label in the training phase so that the model classifies the clean sample incorrectly.
            
            \subsubsection{Test dataset.} The main threat to the test dataset is \textit{adversarial attack}. The adversarial attack can make the model produce wrong results by adding invisible or semantic perturbations to the input data. In recent years, physical adversarial attacks have gained more and more attention. Through 2D patches \cite{duan2020adversarial}, 3D printing \cite{athalye2018synthesizing}, optical illumination \cite{gnanasambandam2021optical} or sensor injection \cite{cao2019adversarial}, the attacker makes the camera or radar collect modified 2D or 3D data and output incorrect classification results.
            It is worth mentioning that this paper divides the adversarial attack into 2D and 3D attacks according to the data and model type rather than the attack media. For example, Athalye \textit{et al.} \cite{athalye2018synthesizing} rendered 2D adversarial images onto the 3D-printed turtles. Although the turtles are 3D objects, the final input of the 2D model is the turtle image collected by the camera, so it is still a 2D adversarial attack.
            
            \subsubsection{Deep learning model.} Training deep learning models requires a lot of resources and time, so a well-trained deep learning model successfully applied to practice has great value. However, there are already some \textit{model extraction attacks} that can steal sensitive information such as model parameters and decision boundaries. In addition, the model itself may also be subject to \textit{model revising attack}. For example, an adversary can make the model misclassify by only flipping several bits of the model \cite{rakin2021t} or poisoning the open-source pre-trained weights \cite{kurita-etal-2020-weight}.

        This paper mainly discusses the adversarial attack because it poses a more practical threat than other attacks. It does not need to have knowledge about the training dataset and even the model architecture (for black-box attacks) and, therefore, can be executed in the inference stage. In addition, generating strong AEs efficiently can improve the effectiveness and efficiency of adversarial training, which is essential for training robust models. 
            
        \subsection{Adversarial goal}
        According to the attacker's goal and the level of security violation, deep learning attacks can be divided into \textbf{confidentiality attack, integrity attack, and availability attack}.  The main goal of the confidentiality attack is to leak the privacy of the model or data, such as the model inverse attack and membership inference attack. The primary aim of the integrity attack is to maliciously change the model's output by modifying the training data, the test data, or even the model itself, such as backdoor attacks, adversarial attacks, and weight modification attacks. The availability attack aims to make the ML services unavailable to legal users. A typical example of availability attacks is to make the dataset \textit{not exploitable} by maliciously poisoning the whole training dataset. For example, the Shortcut attack \cite{yu2022availability} uses linearly separable noises assigned with target labels to mislead the neural network, resulting in the trained model with low accuracy on the test dataset. The current security research in deep learning focuses on confidentiality and integrity.
        
        \subsection{Adversarial capabilities and knowledge}
        The attacker's capabilities refer to the amount of information the attackers have and the actions they can perform. In the context of deep learning systems, the attacker's abilities, from weak to strong, can be classified as follows: only being able to access hard labels and revising test data; being able to access model output confidence; having the ability to access model parameters and training data; modifying training data; modifying model parameters. 
        
        According to the attackers' understanding of models and data, the adversarial knowledge can be classified into \textbf{black-box, gray-box, and white-box models}. There is no clear dividing line between the three. Generally, the attacks that know the model's internal structure and parameters are called white-box attacks. The attacks that can only access the classification results are called black-box attacks, and the attacks in between are called gray-box attacks. In reality, adversarial knowledge is usually determined by adversarial ability, so the two are closely related.

\section{Adversarial attacks for 2D deep learning models}
\label{2D_attack}
The adversarial attack, one of the most threatening attacks, has attracted much research interest. In this and the next section, we will discuss the adversarial attacks in the deep learning-based computer vision system, in which the adversary's goal is to cause the deep learning models to produce wrong predictions. We divide them into 2D and 3D attacks according to the different target model types. The 2D attacks mainly target the 2D models, while the 3D attacks mainly target the 3D model by contaminating 3D data like point clouds.

\subsection{Catalog of 2D adversarial attacks} 
   Although the 2D adversarial attack has been summarized in some previous literature \cite{ serban2020adversarial, machado2021adversarial, He2022TowardsSecurity}, they usually organize from a certain aspect, lacking systematization and the latest research progress. Therefore, there is a need to reorganize the latest studies, especially the emerging ones that cannot be classified into any previous categories, such as color space perturbations and other semantic perturbations. This paper summarizes the most recent progress to help readers keep abreast of the newest tendency. Our survey classified 2D adversarial attacks from various dimensions. 
    \subsubsection{Classifying according to different distance metrics} When the term \textit{adversarial example} was first proposed in 2013 \cite{szegedy2013intriguing}, its definition is leveraging crafted imperceptible perturbation to fool the deep neural network, that is 
    \begin{equation}
        \min_{x'} \left \| {x'-x} \right\|_p \text{, subject to } \mathcal{F}(x') \neq \mathcal{F}(x) \text{ and } x'\in [0,1]^m,
    \end{equation} where $p\in \{0,1,2,\infty\}$. Besides \textbf{$l_p$-norm distance}, recently there are also other distance metrics are proposed to constrain the perturbation size, such as \textbf{color space distance \cite{shamsabadi2020colorfool,laidlaw2019functional}, geodesic distance \cite{fawzi2015manitest}, Wasserstein distance \cite{zheng2019distributionally}}, or even \textbf{without any distance limitation} \cite{song2018constructing}. 

    \subsubsection{Classifying according to physical achievability} Based on whether they can be realized in the physical world, 2D adversarial attacks can be divided into \textbf{digital attack} and \textbf{physical attacks}. The digital adversarial attacks assume that the adversary can directly modify the digital images, while the physical adversarial attacks suppose that the attacker cannot immediately revise the neural network's input and, therefore, revise the real-world objects instead. Physical attacks are more difficult than digital attacks because of limited perturbation space and various environmental variables, such as different viewpoints, distance, and background illumination.   
    \subsubsection{Classifying according to adversary's knowledge level}  As mentioned in the previous section, according to the adversary's knowledge extent, 2D digital adversarial attacks can be divided into \textbf{white-box, gray-box, and black-box attacks}. 
    Different knowledge extent can result in disparate choices of attack methods because of different difficulties. Therefore, we introduce the white-box attacks and black-box attacks separately.     
    
    \subsubsection{Classifying according to the definition of perturbation} 
    According to the perturbation's difference, Gilmer \textit{et al.} \cite{gilmer2018motivating} classified adversarial perturbations into \textbf{imperceptible, content-preserving, and unconstrained perturbation}. The first one sets the original image as the starting point and adds invisible noises to this image. This constraint makes the AE almost the same as the clean one in human eyes and usually uses $l_p$-norm as their distance metrics. For content-preserving perturbation, the adversary retains the semantics of the original image and misleads the classifier at the same time by methods like changing colors \cite{hosseini2018semantic, laidlaw2019functional} and spatially transforms the images \cite{fawzi2015manitest}. These attacks usually use semantic distance to restrain the perturbation. For unconstrained input, the adversary induces erroneous results from the neural networks through any examples. For instance, Song \textit{et al.} \cite{song2018constructing} generates AEs from scratch. 

    Based on the above observation, we classify 2D adversarial attacks into digital white-box and black-box adversarial attacks, and physical adversarial attacks, as shown in Figure.\ref{figure_structure}.

    \begin{table}
    \scriptsize
    \centering
      \caption{Summary of main \textbf{white-box} adversarial attacks in 2D CV tasks sorted by the algorithm and published year}
       \label{tab: summary of 2D digital AE}%
      \begin{threeparttable}
      \setlength{\tabcolsep}{3mm}{
\begin{tabular}{p{6.835em}cp{1.28em}p{1.945em}p{6.11em}p{3.945em}p{2.835em}p{3em}cp{11.665em}}
\toprule
\multirow{2}[4]{*}{Attacks} & \multicolumn{1}{c}{\multirow{2}[4]{*}{Year}} & \multicolumn{2}{p{3.225em}}{Threat model} & \multirow{2}[4]{*}{Algorithm} & \multirow{2}[4]{*}{Distance} & \multicolumn{3}{p{8.115em}}{Performance} & \multirow{2}[4]{*}{Key idea} \\
\cmidrule{3-4}\cmidrule{7-9}\multicolumn{1}{c}{} &       & Goal  & Knowl.\tnote{*} & \multicolumn{1}{c}{} & \multicolumn{1}{c}{} & Efficiency & Camouflage & \multicolumn{1}{p{2.28em}}{ ASR \tnote{**}} & \multicolumn{1}{c}{} \\
\midrule
L-BGFS \cite{szegedy2013intriguing} & 2014  & T     & $\square$ & Optimization & $L_2$ & Costly & Invisible & 100.0\% & Box constrained L-BGFS \\
JSMA \cite{papernot2016limitations} & 2016  & T     & $\square$ & Optimization & $L_0$ & Efficient & Slight & 97.2\% & $L_0$ + greedy \\
UAP \cite{moosavi2017universal} & 2017  & U     & $\square$ & Optimization & $L_1,L_\infty$ & Medium & Invisible & 93.7\% & Universal perturbation \\
C\&W \cite{carlini2017towards} & 2017  & T\&U  & $\square$ & Optimization & $L_0,L_2,L_\infty$ & Medium & Invisible & 100.0\% & Substitute loss function \\
EAD \cite{chen2018ead} & 2017  & T\&U  & $\square$ & Optimization & $L_1,L_2$ & Costly & Invisible & 100.0\% & $L_1$ + $L_2$ distance \\
DAG \cite{xie2017adversarial} & 2017  & U     & $\square$, & Optimization & $L_\infty$ & Efficient & Invisible & 69.0\% & Segmentation \\
SV-UAP \cite{khrulkov2018art} & 2018  & U     & $\square$ & Optimization & $L_2, L_\infty$ & Medium & Marked & 60.0\% & Singular vector \\
GD-UAP\cite{mopuri2018generalizable} & 2018  & U     & $\square$,$\color{gray}\blacksquare$ & Optimization & $L_\infty$ & Medium & Invisible & 83.5\% & Data-free UAP \\
RobustAdv \cite{luo2018towards} & 2018  & U     & $\square$,$\blacksquare$ & Optimization & Percep. & Costly & Slight & 98.5\% & Perceptual sensitivity \\
TAP \cite{zhou2018transferable} & 2018  & U     & $\square$ & Optimization & $L_2$ & Efficient & Marked & 46.7\% & Low-pass filter \\
DAA\cite{zheng2019distributionally} & 2019  & U     & $\square$ & Optimization & $L_\infty$ & Medium & Marked & 45.0\% & Advl-data distribution \\
SparseFool  \cite{modas2019sparsefool} & 2019  & U     & $\square$ & Optimization & $L_1$ & Medium & Slight & 100.0\% &  Projected $L_1$-DeepFool \\
DF-UAP\cite{zhang2020understanding} & 2020  & T\&U  & $\square$ & Optimization & $L_\infty$ & Efficient & Slight & 96.2\% & Targeted data-free UAP \\
sC\&W \cite{zhang2020smooth} & 2020  & T\&U  & $\square$ & Optimization & Smooth & Medium & Marked & 98.0\% & Smooth perturbation \\
GreedyFool \cite{dong2020greedyfool} & 2020  & U     & $\square$ & Optimization & $L_0$ & Costly & Invisible & 94.6\% & Greedy algorithm \\
SSAH \cite{luo2022frequency} & 2022  & T\&U  & $\square$ & Optimization & Low Freq. & Medium & Invisible & 99.8\% & High-frequency limitation \\
FGSM \cite{goodfellow2014explaining} & 2015  & U     & $\square$ & Fast Gradient  & $L_\infty$ & Efficient & Marked & 72.3\% & One-step gradient attack \\
Deepfool \cite{moosavi2016deepfool} & 2016  & U     & $\square$ & Fast Gradient  & $L_1,L_2,L_\infty$ & Medium & Invisible & 90.0\% & Boundary dist. estimation \\
BIM\&ILCM \cite{kurakin2018adversarial} & 2016  & T\&U  & $\square$ & Fast Gradient  & $L_\infty$ & Medium & Slight & 90.0\% & Iteration FGSM \\
MI-FGSM \cite{dong2018boosting} & 2017  & T\&U  & $\square$, & Fast Gradient  & $L2, L_\infty$ & Efficient & Slight & 100.0\% & Momentum \\
PGD \cite{madry2017towards} & 2017  & U     & $\square$ & Fast Gradient  & $L_2, L_\infty$ & Medium & Slight & 99.2\% & Projection grad. descent \\
R-FGSM \cite{tramer2017ensemble} & 2018  & T\&U  & $\square$ & Fast Gradient  & $L_\infty$ & Efficient & Slight & 64.8\% & Random FGSM \\
BPDA\&EOT \cite{athalye2018obfuscated} & 2018  & U     & $\square$ & Fast Gradient  & $L_2, L_\infty$ & Efficient & Invisible & 100.0\% & Attack obfuscated grad. \\
MDI2FGSM\cite{xie2019improving} & 2019  & T\&U  & $\square$ $\blacksquare$ & Fast Gradient  & $L_\infty$ & Costly & Slight & 62.2\% & Random transformation \\
TI-BIM \cite{dong2019evading}  & 2019  & U     & $\square$ & Fast Gradient  & $L_\infty$ & Costly & Slight & 82.0\% & Translation kernels \\
DDN \cite{rony2019decoupling} & 2019  & U     & $\square$ & Fast Gradient  & $L_2$ & Efficient & Slight & 100.0\% & Adjustable step size \\
HP-UAP\cite{zhang2021universal} & 2021  & T\&U  & $\square$,$\blacksquare$ & Fast Gradient  & $L_\infty$ & Efficient & Invisible & 91.1\% & Frequency filter \\
NAG-UAP \cite{mopuri2018nag} & 2018  & U     & $\square$ & GAN   & $L_\infty$ & Costly & Slight & 94.1\% & Generative UAP \\
ATN \cite{baluja2018learning} & 2018  & T     & $\square$ & GAN   & $L_2$ & Efficient & Slight & 95.9\% & GAN+reranking \\
UnresGM\cite{song2018constructing} & 2018  & T     & $\square$ & GAN   & $-$   & Costly & Marked & 84.0\% & GAN from scratch \\
GAP \cite{poursaeed2018generative} & 2018  & T\&U  & $\square$ & GAN   & $L_2, L_\infty$ & Efficient & Slight & 74.1\% & UAP and img-dependent \\
AdvAttGAN\cite{joshi2019semantic} & 2019  & U     & $\square$ & GAN   & $-$   & Costly & Marked & 98.0\% & Modify facial attribute \\
SemAdv \cite{qiu2020semanticadv} & 2020  & T     & $\square$,$\blacksquare$ & GAN   & $-$   & Costly & Marked & 67.7\% & Modify visual attribute \\
SparseGAN \cite{he2022transferable} & 2022  & U     & $\square$ & GAN   & $L_0$ & Efficient & Invisible & 58.4\% & Perturbation decoupling \\
Manitest \cite{fawzi2015manitest} & 2015  & U     & $\square$ & Spatial Trans. & Geodesic & Costly & Invisible & 35.6\% & Geodesics on manifold \\
SimpleTrans \cite{engstrom2018rotation} & 2017  & U     & $\square$ $\blacksquare$ & Spatial Trans. & $-$   & Medium & Invisible & 90.0\% & Simple transformation \\
Manifool \cite{kanbak2018geometric} & 2018  & U     & $\square$ & Spatial Trans. & Geodesic & Efficient & Invisible & 75.0\% & Iterative method \\
stAdv \cite{xiao2018spatially}  & 2018  & T     & $\square$ & Spatial Trans. & T.V.  & Efficient & Slight & 99.6\% & Flow field \\
Adef \cite{alaifari2018adef} & 2019  & T     & $\square$ & Spatial Trans. & $L_2$ & Medium & Invisible & 99.0\% & Iterative deforming \\
3DRender \cite{zeng2019adversarial} & 2019  & U     & $\square$ & Spatial Trans. & $L_2$ & Costly & Marked & 90.7\% & 3D renderer \\
PSI \cite{zheng2019distributionally} & 2019  & U     & $\square$ & Spatial Trans. & Wasserstein & Efficient & Invisible & 91.7\% & Wasserstein distance \\
EdgeFool \cite{shamsabadi2020edgefool} & 2020  & U     & $\square$ & Spatial Trans. & Smooth & Costly & Marked & 99.0\% & Image enhancement \\
Chroma-shift \cite{aydin2021imperceptible}  & 2021  & T\&U  & $\square$ & Spatial Trans. & $-$   & Efficient & Invisible & 96.1\% & YUV colorspace \\
FilterFool \cite{shamsabadi2021semantically} & 2021  & U     & $\square$ & Spatial Trans. & SSIM  & Costly & Slight & 48.3\% & Mimic filter \\
Semantic \cite{hosseini2018semantic} & 2018  & U     & $\square$ & Color Trans. & $-$   & Costly & Marked & 94.3\% & HSV colorspace \\
Blind-Spot \cite{zhang2018limitations} & 2019  & U     & $\square$ & Color Trans. & kNN   & Efficient & Marked & 100.0\% & Blind Spot \\
ReColorAdv \cite{laidlaw2019functional} & 2019  & U     & $\square$ & Color Trans. & Smooth & Medium & Slight & 97.0\% & RGB/CIELUV colorspace \\
cAdv\&tAdv \cite{bhattad2019unrestricted} & 2020  & T     & $\square$ & Color Trans. & $-$   & Medium & Marked & 99.7\% & Colorization and texture \\
PerC attack\cite{zhao2020towards} & 2020  & U     & $\square$ & Color Trans. & CIEDE2000 & Efficient & Invisible & 100.0\% & Perceptual color distance \\
ColorFool \cite{shamsabadi2020colorfool} & 2020  & U     & $\square$,$\blacksquare$ & Color Trans. & $-$   & Costly & Marked & 95.90\% & Adversarial colorization \\
\bottomrule
\end{tabular}%

  }
 
  \begin{tablenotes}
    \footnotesize
    \item[*] This column is the adversarial knowledge of different attacks. $\square$: white-box. $\blacksquare$: black-box. ${\color{gray} \blacksquare}$: gray-box.
    \item[**] We only count the best result of the most difficult attack reported in the papers. E.g. for works that have both white-box and black-box attacks, we only report the latter, for it is more difficult. For works with both untargeted and targeted attacks, we report the latter for the same reason.
  \end{tablenotes}
  \end{threeparttable}
\end{table}%

    \subsection{White-box adversarial attack for 2D deep learning models}
        Table \ref{tab: summary of 2D digital AE} summarizes recent year's 2D adversarial attacks for the white-box model. We divide them into the following categories according to their specific attack methodologies.
         \begin{itemize}
            \item Optimization-based attacks. This kind of attack usually describe finding AEs as objective optimization problems and solve these problems by existing or self-defined objective optimization method, include \textit{L-BFGS} \cite{szegedy2013intriguing}, \textit{UAP} \cite{moosavi2017universal}, \textit{C\&W} \cite{carlini2017towards}, \textit{EAD} \cite{chen2018ead} and \textit{OptMargin} \cite{he2018decision}, etc.

            \item Fast-gradient-based attacks. Instead of reaching a local minimum through optimizing an objective, this kind of attack finds adversarial examples through direct and explicit gradient computation. Therefore, they can usually find an adversarial example very quickly, although the perturbation may not be the optimum. Moreover, because these attacks lack clear objective functions, they are usually untargeted.
            
            \item GAN-based attacks. Instead of optimizing the noises through gradient descent, this kind of attack generates the perturbation through the generative adversarial network (GAN) and optimizes the variable in the latent space. For example, unrestricted AE \cite{song2018constructing} can generate semantic adversarial examples from drafts.
            
            \item Spatial-transformation-based attacks. Unlike additive noises, this attack utilizes global or local spatial transformations to generate adversarial examples. The former includes Manifool \cite{kanbak2018geometric} etc. and the latter includes stAdv \cite{xiao2018spatially} etc. This kind of attack usually uses semantic loss rather than p-norm-based distance metrics.
            
            \item Colorization-transformation-based attacks. Some works also transform images into different color spaces and manipulate the color space instead because human beings prefer to classify objects according to their shapes rather than colors. These attacks include ColorFool \cite{shamsabadi2020colorfool}, Chroma-shift \cite{aydin2021imperceptible} and SemanticAdv \cite{hosseini2018semantic}, etc.            
        \end{itemize}
        Some attacks also combine several different methods. For example, Zhao et al. \cite{zhao2019perturbations} combined transformation and pixel-wise perturbations to enhance the attack strength. However, most attacks adopt one of the above methods as their main methodology. Therefore, we will introduce these attacks according to their primary methodology.

        \subsubsection{Fast-gradient-based attacks}      
        
        The greatest difference between optimization-based and fast-gradient-based attacks is that the latter can quickly find adversarial examples in one or several iterations. However, they usually cannot reach a local optimum and need larger perturbation to mislead the models. Because of their high efficiency in finding adversarial examples, they can better combine with the adversarial training process to train robust neural networks.
        
        \paragraph{Single-step fast gradient methods} 
        In 2015, Goodfellow \textit{et al.} \cite{goodfellow2014explaining} conjectured that the AE exists because of the linear representation of features in high-level space and proposed gradient-based one-step attack method named \textit{fast gradient sign method} (\textbf{FGSM}), which estimates adversarial examples by $x' = x + \epsilon sign(\bigtriangledown_x \mathcal{L}_{\mathcal{F}}(x,y)).
        $ Moreover, they embedded FGSM into \textit{adversarial training} to mine hard examples and train robust models. The loss function of FGSM-based adversarial training is 
        \begin{equation}
            \widetilde{\mathcal{L}}_{\mathcal{F}}= \lambda \mathcal{L}_{\mathcal{F}}(x,y)  + (1-\lambda)\mathcal{L}_{\mathcal{F}}(x+\epsilon sign(\bigtriangledown_x \mathcal{L}_{\mathcal{F}}(x,y)),y).
            \label{Eq-advtraining}
        \end{equation}
        The loss function has two parts. The first part is the same as the normal training procedure, while the second part considers the negative influence of adversarial examples.

        \paragraph{Multi-step fast gradient methods} 
        Kurakin \textit{et al.} \cite{kurakin2018adversarial} extended the FGSM method to an iterative method (\textbf{BIM}). At the end of each step, they clipped the image into [0,1]. They also proposed \textbf{LLCM}, which uses the least likely label as the target label. In addition, they tried to attack the camera through printed images. 
        but their attacks suffered low ASRs in the physical setting, for they lacked consideration of fabrication distortion. 
        
        Moosavi \textit{et al.} \cite{moosavi2016deepfool} proposed \textbf{Deepfool} attack, which iteratively estimates the distance from the normal samples to the classifying hyperplane. Because of the nonlinearity of the classification boundary, they linearize the hyperplane at each iteration. If the model $\mathcal{F}(x)$ is a binary classifier, at each iteration, $x$ is updated by $ x^{t+1}  \gets x^t -\frac{\mathcal{F}(x^t)}{\left \| {\bigtriangledown \mathcal{F}(x^t)} \right\|_2^2}	\bigtriangledown \mathcal{F}(x^t)$. If the model $\mathcal{F}$ is a multiclass differentiable classifier, they use a polyhedron $\widetilde{P}_i$ to approximate the label space. Deepfool is also untargeted but can achieve smaller perturbations than the FGSM attack.
        
        Madry \textit{et. al.} \cite{madry2017towards} proposed projection gradient descent (\textbf{PGD}), which projects $x_t$ onto the neighborhood of the input at the end of each iteration through $x^{t+1}=\textit{proj}_{x+\mathcal{S}}(x^t + \epsilon \, sign(\bigtriangledown_x L(\theta,x,y)))$. Osadchy \textit{et al.} \cite{Osadchy2017} also proposed a similar method called I-FGSM and apply it to the image CAPTCHAs to distinguish between computers and humans.    
        
        Tramer \textit{et al.} \cite{tramer2017ensemble} found that the steep curvature artifacts near the inputs may reduce the attack strength of single-step attack. Therefore, they proposed \textbf{R-FGSM}, which introduces random noises into the input images. The update formulation is $x^{adv}=x'+(\epsilon-\alpha)\cdot sign(\bigtriangledown_{x'}\mathcal{L}(x',y_{true}))$, where $x'=x+\alpha \cdot sign(\mathcal{N}(\textbf{0}^m,\textbf{I}^m))$, $\mathcal{N}(\textbf{0}^m,\textbf{I}^m)$ is the standard normal distribution. R-FGSM attack can be regarded as a single-step variant of the PGD attack.    
        
        \textbf{MI-FSGM} \cite{dong2018boosting} introduced a momentum term into the iterative FGSM and achieved better performance than FGSM and BIM in the untargeted attack. It firstly updated the momentum by $g_{t+1} = \mu \cdot g_t + \frac{\mathlarger{\bigtriangledown}_x\mathcal{L}(x_t',y)}{\left\|\mathlarger{\bigtriangledown}_x\mathcal{L}(x_t',y)\right\|_1}$, then updated the adversarial example $x_{t}'$ by $x_{t+1}' = x_t' + \alpha \cdot sign(g_{t+1})$.
        By generating a velocity vector in the direction of gradient descent, MI-FGSM alleviates the problem of unstableness after multi-step iterations and enhances the transferability of AEs.

        \textbf{M-DI2-FGSM} \cite{xie2019improving} integrated random transformations into the MI-FGSM to improve the adversarial example's transferability. Random resize and padding are used as the main transforms, and the transferability is increased by about 20\% on the ImageNet dataset. However, this method may cause the loss to decline unstably. Dong \textit{et al.} \cite{dong2019evading} also proposed a similar attack named \textbf{TI-BIM} based on BIM and transformation. Because the gradient of transformed images equals transforming the gradient of original examples, they multiply the gradient of original images with a uniform, linear or Gaussian kernel at each iteration to represent different translations.
        
        Rony \textit{et al.} \cite{rony2019decoupling} proposed Decoupling Direction and Norm attack (\textbf{DDN}) based on PGD. At each iteration, the perturbation $\delta_k$ is projected to an $\epsilon_k$-ball around $x$ through $x_k' \leftarrow x+\epsilon_k\frac{\delta_k}{\left\|\delta_k\right\|_2}$. Different from PGD that sets $\epsilon_k$ as a constant, they increase the $\epsilon_k$ at each iteration if $x_k'$ is still not adversarial and reduce the $\epsilon_k$ on the other hand. Their method achieved  competitive results compared with state-of-art $L_2$-norm attacks and needed fewer iterations. 

\begin{figure}[t]
            \centering
            \includegraphics[scale=0.5]{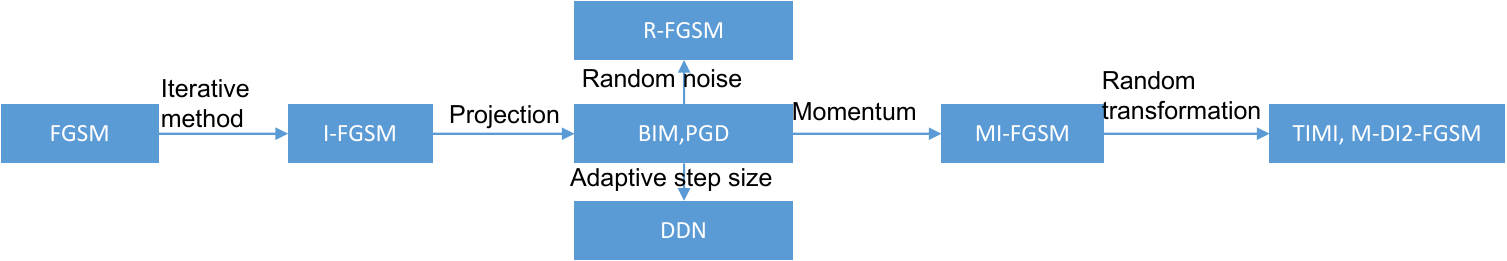}         
            \caption{The relationships between different fast-gradient-based 2D adversarial attacks}  
            \label{fig-FGSM}
            \Description{The relationships between different fast-gradient-based attacks}
            \end{figure}

        The relationships between different fast-gradient-based adversarial attacks are shown in Figure.\ref{fig-FGSM}.
       
        \subsubsection{Optimization-based attacks}
        Optimization-based attacks usually model the adversarial as optimization objectives and then leverage existing optimization methods or design custom optimization methods to solve these optimization problems. Moreover, most optimization-based attacks are based on $L_p$-norm distance. Different distance metrics can generate different kinds of perturbations. $L_2$ and $L_\infty$ distances usually generate uniform and dense perturbation, while $L_0$ and $L_1$ distances tend to generate sparse perturbation. Moreover, Besides image-independent attacks, universal perturbation has also attracted much research interest. Therefore, according to the different kinds of perturbations, we divide optimization-based attacks into dense, sparse, and universal perturbations. 

        \paragraph{Dense perturbation.} When the concept of adversarial example is first proposed, it is defined as minimizing the total perturbation in $L_2$ or $L_\infty$ distance because they are easier to compute gradient than other distances. These metrics lead to uniform perturbation on the whole image. The representative attacks include L-BFGS \cite{szegedy2013intriguing}, C\&W \cite{carlini2017towards}, etc.
        
        The concept of adversarial example was proposed by Szegedy \textit{et al.} \cite{szegedy2013intriguing} in 2013. They formulated the problem of finding adversarial examples as a box-constrained optimization problem under $l_2$ distance and utilized \textbf{L-BFGS} algorithm to find an approximate solution. In detail, for an input image $x \in \mathbb{R}^m$, they minimize $c\left \| {\delta} \right\|_2 + \mathcal{L}_{\mathcal{F}}(x+\delta, y'), \text{s.t. } x+\delta \in [0,1]^m$, in which $\mathcal{L}_{\mathcal{F}}(\cdot,\cdot)$ is a cross-entropy function.  In addition, they also analyzed the upper Lipschitz constant of the model and proposed that the regularization of the model parameters may avoid the existence of adversarial examples. However, experiments show that L-BFGS has the weakness of inefficiency.
        

        Defensive distillation \cite{papernot2016distillation} improved the model robustness by training models on \textit{soft training labels} to reduce the network's Jacobian matrix, making the model less sensitive to the input and avoiding overfitting against any samples. \textbf{C\&W} \cite{carlini2017towards} designed an alternative loss to break defensive distillation, that is
         $
         max(max({Z_2(x')_{i:i\neq y'}})-Z_2(x')_{y'}, -\kappa),
         \label{CWloss}
         $
         where $\kappa$ is a parameter to control the strength of AE. 
         Moreover, some former attacks, like BIM \cite{kurakin2018adversarial}, directly clip $x'$ into $[0,1]^m$, which may result in zero gradients. C\&W changed the optimized variable from $\delta$ to $w \coloneqq arctanh(\delta)$.
        Because $tanh$ is a bounded and differential function, C\&W can remove the box constraint and use \textit{Adam} optimizer. 

        Gradient mask \cite{papernot2017practical} prevented adversarial attack through shattered, randomized, or vanishing/exploding gradients. Athalye \textit{et al.} \cite{athalye2018obfuscated} claimed that gradient masks also cannot provide enough robustness. They proposed \textbf{BPDA, EOT}, and \textbf{Reparameterization} attacks to break these defenses, respectively. Take BPDA as an example. Gradient shattering introduces a non-differentiable layer $f^i(x)$ to prevent the adversary from getting the gradient. BPDA attack adopts a differentiable layer $g(x)$ to substitute $f^i(x)$ in the back-propagation process.

        Adversarial training \cite{szegedy2013intriguing} is also a popular defensive method. It uses adversarial examples as training data to improve model robustness (See Eq.\ref{Eq-advtraining}). Adversarial training through PGD \cite{madry2017towards} has been proven effective in defending most first-order attacks, like FGSM and DeepFool. However, Zhang \textit{et al.} \cite{zhang2018limitations} found that adversarial trained networks may have \textbf{blind spots} or unusual examples. They first use kNN to measure the difference from the test example to the training dataset and adopt simple methods like scaling and shifting all pixels to find the blind-spot example of the model. Then, they attack these infrequent samples by C\&W attack. 
        
        To improve the transferability of adversarial perturbation, \textbf{TAP} attack \cite{zhou2018transferable} increases the feature distance of all latent layers between the normal and adversarial image. In addition, they also found that applying a low-pass filter on the perturbation can also improve transferability. Observing that optimizing on an independent example cannot be globally optimal, \textbf{DAA} \cite{zheng2019distributionally} searches an adversarial data distribution that satisfies $L_\infty$ constraint while maximizing the generalization error by Wasserstein gradient flows. DAA can reduce the accuracy of CIFAR to 44.71\% for adversarial-trained networks. To make the perturbation more indistinguishable, \textbf{sC\&W} \cite{zhang2020smooth} adds Laplacian smoothing constraint on the original C\&W attack. This constraint makes the perturbation locally smooth on the even areas and dissonant on the high variance area. Like the sC\&W attack, \textbf{SSAH} \cite{luo2022frequency} composes a semantic similarity loss and a low-frequency restraint. The former minimizes the cosine similarity of the features between the original and target image. The latter restrains the perturbation into the high-frequency regions.

        Semantic segmentation and object-detecting tasks are also significant in computer vision. Xie \textit{et al.} \cite{xie2017adversarial} proposed dense adversary generation (\textbf{DAG}) attack against multi-object detection task. Observing that this task usually needs to classify plural targets in an image, DAG attacks all targets simultaneously using the iterative gradient descent. DAG can also transfer among different models, which is probably because it accumulates perturbations from multiple targets.

        \paragraph{Sparse perturbation.} In some scenarios, the number of pixels being modified is more crucial than the overall magnitude of the perturbation. Therefore, some researchers proposed generating sparse perturbations through $l_0$ or $l_1$ distance, equivalent to modifying as few pixels as possible. However, specific techniques are needed because these distances are non-differentiable. For example, \textbf{JSMA} \cite{papernot2016limitations} selects the most meaningful pixels by comparing the gradient of $Z(x)$ and adds them to the set of modified pixels one by one until the attack succeeds or a certain threshold is reached. \textbf{EDA} attack \cite{chen2018ead} formulates the problem as an Elastic-net regularized optimization problem that combines the $l_1$ and $l_2$ distance. Because $l_1$ distance is non-differentiable, EDA applies the FISTA algorithm to solve this optimization function. The $l_1$-DeepFool \cite{moosavi2016deepfool} can generate sparse perturbations efficiently through linearizing the boundary, but it is greatly affected by the clipping function. To solve this problem, \textbf{SparseFool} \cite{modas2019sparsefool}  projects $x'$ on one component of the normal vector at each iteration. If the projection cannot generate an adversarial example, this direction will be ignored in the next iteration. SparseFool iteratively estimates the minimum perturbation to the boundary until the predicted label is changed. \textbf{RobustAdv}  \cite{luo2018towards} generates imperceptible and robust adversarial examples through perception distance, which is defined as $D(X',X)=\sum_{i=1}^{N}\delta_i\cdot Sen(x_i)$, where $N$ is the total pixels being modified, $\delta_i$ is the perturbation of the pixel $x_i$, and $Sen(x_i)$ is the perturbation sensitivity of $x_i$. Because humans are more sensitive to the perturbation in smoothing regions, RobustAdv defines $Sen(x_i)$ as the reciprocal of the standard deviation around $x_i$. \textbf{GreedyFool} \cite{dong2020greedyfool} uses a two-step greedy-based optimization to generate sparse perturbation. Firstly, the candidate manipulation locations are selected according to their gradient and their distortion map generated by a GAN. Secondly, unnecessary points are discarded to improve the sparsity. 
        There are also some black-box attacks that can generate sparse perturbation, such as OnePixel \cite{su2019one} and CornerSearch \cite{croce2019sparse} and GeoDA\cite{rahmati2020geoda}. We will discuss them in the next subsection.
        
        \paragraph{Universal perturbation.} Rather than optimizing the perturbation on a specific image, universal perturbation produces image-agnostic perturbations through optimizing perturbation over the whole or part of the training dataset.
        
        In 2017, Moosavi-Dezfooli \textit{et al.} \cite{moosavi2017universal} firstly proposed the concept of universal adversarial perturbations (\textbf{UAP}). They successfully attacked most images in the validation set by aggregating the minimum perturbations on the training images. Experiments show that UAP attack can not only attack unseen images but also transfer between different models with a success rate of 40\textasciitilde60\%. Later, in 2018, Khrulkov \textit{et al.} \cite{khrulkov2018art} proposed \textbf{SV-UAP} attack, which approximates universal perturbation problem as a (p, q)-singular problem. Suppose $J_{i}(x)$ is the Jacobian matrix of $i^{th}$-to-last layer. For a small vector $\delta$, $Z_i(x+\delta)-Z_i(x) \approx J_i(x) \delta$. Therefore, the problem of finding a universal perturbation can be formulated as
        $
        \max_{\delta} \; \sum_{x_j \in X_b} \left\| J_i(x_j) \delta \right\|_q^q, \;s.t.\; \left\|\delta \right\|_p = C
        $
        where $X_b$ is a subset of the training dataset. Then, they use the stochastic power method to solve this (p,q)-singular vector problem and achieved a fooling rate of 60\% on 50000 images by using only 64 images for optimization. Mopuri \textit{et al.} \cite{mopuri2018generalizable} proposed a generalizable data-free UAP attack (\textbf{GD-UAP}) in which the adversary does not need to access the exact training data. The idea is to find an image-agnostic perturbation that triggers more additional model activation. The propagation effect of the neural network will eventually lead to misclassification. They use the knowledge about the distribution of the training dataset to improve the fooling rate.
        In 2020, Zhang \textit{et al.} \cite{zhang2020understanding} firstly proposed a targeted UAP attack named \textbf{DF-UAP}. They analyzed the similarity between the predicted logits of clean images and UAP using the Pearson correlation coefficient and found that UAP has a dominant role in prediction compared with clean images. Based on this discovery, they randomly sample the proxy dataset at each iteration to update the perturbation. Inspired by the success of steganography of neural networks, In 2021, Zhang \textit{et al.} \cite{zhang2021universal} proposed \textbf{HP-UAP}, which adds a high-pass filter in the iteration to make the UAP more invisible to human eyes with only a small decrease of attack success rate.
        

        
        \subsubsection{GAN-based attacks}
        The generative adversarial network (GAN) was proposed in 2014 by Goodfellow and quickly became a powerful tool for learning the distribution of training datasets. The first related work, adversarial transformation network (\textbf{ATN}) \cite{ baluja2018learning}, is reported in 2017. Its loss function is $ \min_{\theta} \sum_{x \in \mathcal{X}}\beta L_2(\mathcal{G}(x),x) + L_2(f(\mathcal{G} (x)),r(f(x),t))$, where $r$ is a reranking function to maintain the logits' ranking order except the target class. 
        Song \textit{et al.} \cite{song2018constructing} proposed to use GAN to generate \textbf{unrestricted AEs} from noises $z$ rather than adding perturbation. The loss function includes three parts. The first is to make the victim model predict $\mathcal{G}(z,y_{s})$ as the target class. The second is to limit the search region of $z$, and the third is to make the auxiliary classifier predict $\mathcal{G}(z,y_{s})$ as the source class $y_s$. They also engage humans to evaluate the fidelity of AEs. Moreover, they proposed a noise-augmented version to improve the ASR. 
        Konda \textit{et al.} \cite{mopuri2018nag} proposed \textbf{NAG-UAP}, which models the distribution of UAP by a UAP generator. The loss function of the GAN network includes two parts: a fooling loss and a diversity loss. The former's goal is to minimize the classifier's output on the ground truth label, while the latter's goal is to produce diversified UAPs by maximizing the gap of the hidden layer's outputs on different UAPs. 
        Poursaeed \textit{et al.} \cite{poursaeed2018generative} proposed \textbf{GAP} attack that generates both universal and image-independent noises for targeted and untargeted attacks. For universal attacks, they use image-to-image translation networks to generate universal perturbations from random noises and crop the perturbations to have a fixed norm. Next, the perturbation is added to the normal images and sent to the victim model to calculate the fooling loss. For image-independent attacks, the perturbations are generated from the original image instead.        
        He \textit{et al.} \cite{he2022transferable} utilized a generative model to improve the transferability of sparse perturbation. They first decoupled perturbation into a magnitude component and a location component. Then, they add a sparse loss onto the location component. Because the location component is a binary operator, they proposed a binary quantization operator to train the generator.  

        The emergence of conditional neural networks makes directly editing artificially designed properties possible (e.g. with and without eyeglasses). Joshi \textit{et al.} \cite{joshi2019semantic} used a conditional generative model to generate semantic adversarial examples. Their attack optimized over the attributes space of the conditional generative model to manipulate semantic features like wearing glasses or not and different skin colors to fool the deep learning models. Qiu \textit{et al.} \cite{qiu2020semanticadv} also proposed a similar method. But their method can choose arbitrary targets to attack. In addition, they manipulated the interpolated feature space instead of the attributes manifold.

        \subsubsection{Spatial-transformation-based attacks} Although deep neural networks like CNN are designed to be invariant and robust to transformations like translation and rotation, researchers have shown this is not always the case. 


            \begin{figure}[t]
            \centering
            \subfigure[ManiFool]{
            \includegraphics[scale=0.23]{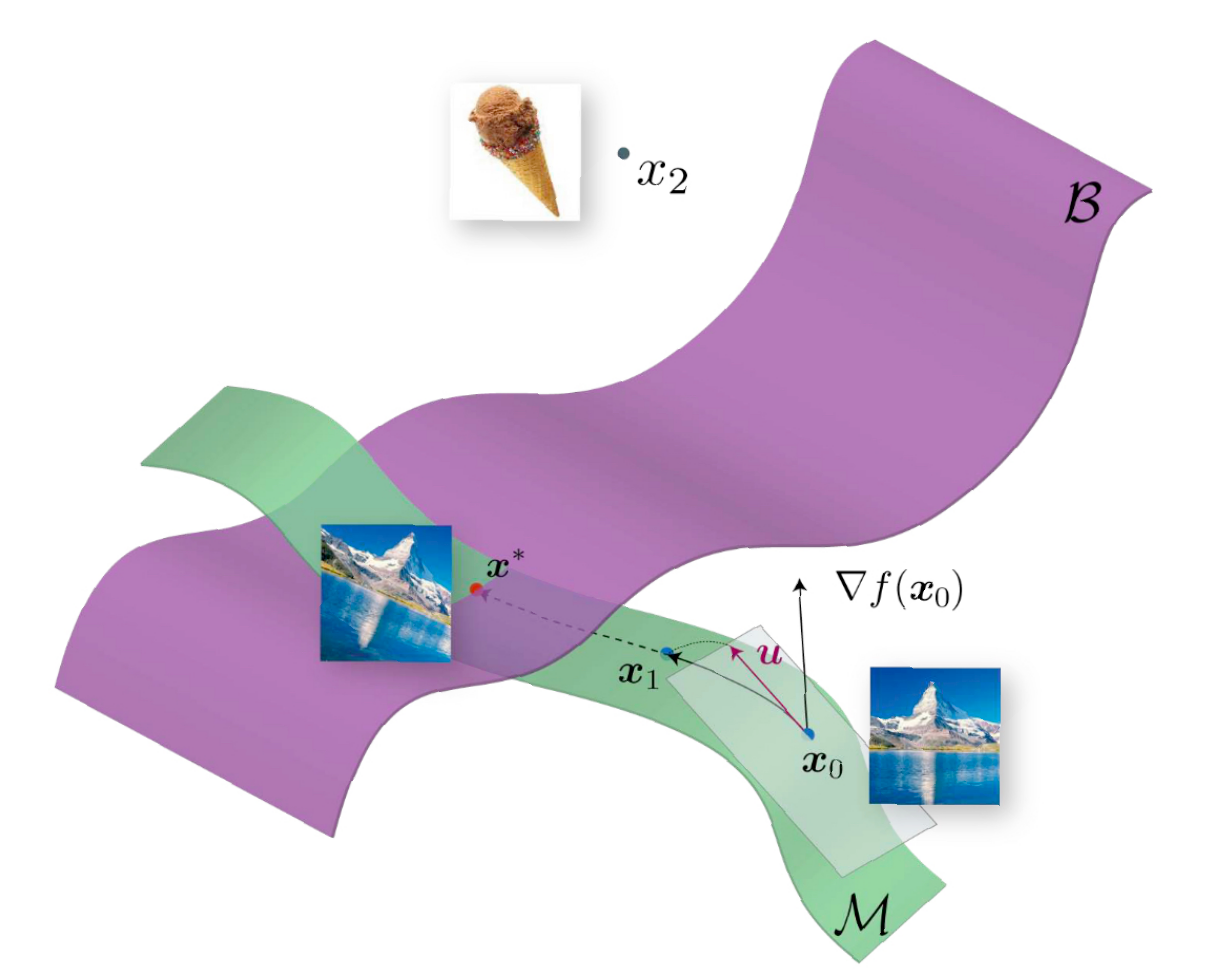}}            
            \subfigure[stAdv]{
            \includegraphics[scale=0.47]{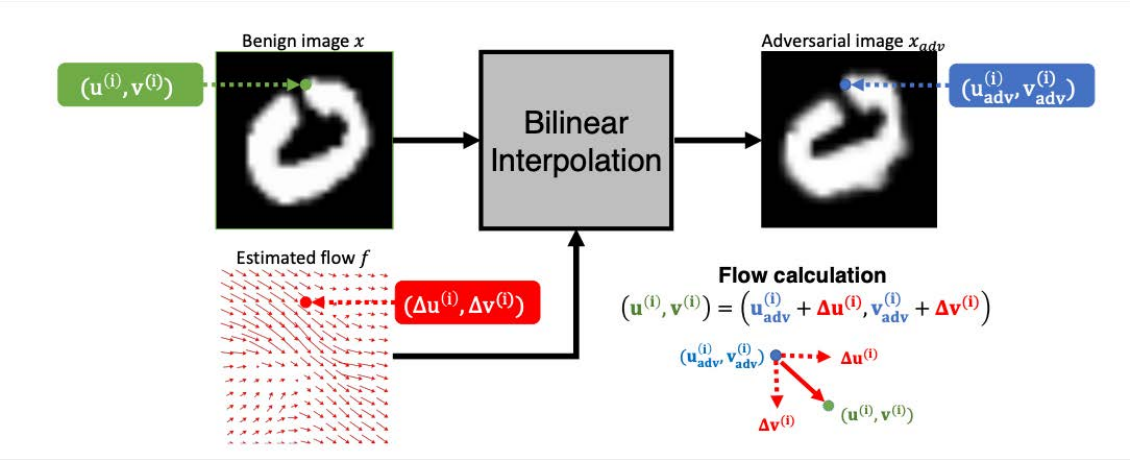}}
            \caption{Comparing globally and locally transformation attacks. (a) ManiFool attack \cite{kanbak2018geometric} optimizes rotation and transition parameters on the manifolds. (b) stAdv attack \cite{xiao2018spatially} optimizes the flow field on pixel coordinate space. The pixel values are viewed as a function of the pixel coordinates. The flow field is obtained by backpropagation from the adversarial loss function to pixel coordinates.}  \label{fig-stAdv}
            \Description{The structure of this paper.}
            \end{figure}
            
        \paragraph{Globally spatial-transformation-based attacks}
        The first work of transformation-based attacks, \textbf{Manitest} \cite{fawzi2015manitest}, was proposed in 2015. It casts the problem of computing the gap between converted images as the geodesics along the transformation manifold, which is defined as the minimum length of the curve on the manifold. Manitest utilizes fast marching to find adversarial transformations, which can assure minimal transformation on the searched grids. However, it suffers a low efficiency when the transformation variety increases. \textbf{SimpleTrans} \cite{engstrom2018rotation} also uses simple spatial transformations to fool networks. It contains a white-box algorithm based on gradient descent and a black-box algorithm based on grid search. In 2018, Kanbak \textit{et al.} \cite{kanbak2018geometric} proposed \textbf{Manifool} to analyze the transformation robustness of deep learning models, as shown in Figure.\ref{fig-stAdv} (a). They also used the geodesic distance to measure the two transformations' distance. The Manifool consists of two steps: determining the changing direction of the image and mapping this change onto the manifold. Although their method is not as accurate as Manitest's, they can achieve high efficiency even with operating similarity and affine transformations simultaneously. 

        \paragraph{Locally spatial-transformation-based attacks}
        As shown in Figure.\ref{fig-stAdv} (b), the \textbf{stAdv} attack \cite{xiao2018spatially} defined the transformation as a spatial flow field, where each flow vector corresponds to a displacement of the corresponding pixel.  To enhance the local smoothness of the flow field to make the AE more natural, they define a flow cost as the total variation of the flow field. They also visualized the CAM attention and showed that their method can better distract attention than C\&W methods.
        In 2019, Alaifari \textit{et al.} \cite{alaifari2018adef} proposed \textbf{ADef} attack inspired by Deepfool. They iteratively deform the original image through gradient descent to move the image over the boundary of the classification. The pixel's value is modeled as a function of its coordinate. Moreover, a two-dimensional Gaussian filter is used to inflict smoothness on the flow field. In 2020,
        \textbf{EdgeFool} \cite{shamsabadi2020edgefool} generates adversarial examples through enhancing image details. Contrary to ColorFool \cite{shamsabadi2020colorfool}, EdgeFool only modifies the $L$ channel and leave $a$ and $b$ channel unmodified. It firstly trains an FCNN network to learn a smoothing image $I_s$ from the input image $I$, and modifies the image details $I_d:=I-I_s$ to generate adversarial examples. 
        Inspired by EdgeFool, \textbf{FilterFool} \cite{shamsabadi2021semantically} viciously modifies a picture by imitating normal filter functions like detail enhancement, gamma correction, and log transformation. Their attack comprises an FCNN network, a traditional filter, and a classifier. The FCNN network is trained to generate perturbation that simultaneously fools the classifier and makes the perturbed image's structure similar to the filtered image. Experiments show that FilterFool can improve the transferability of adversarial examples.
        Wong \textit{et al.} \cite{wong2019wasserstein} generated AEs by Wasserstein distance instead of $l_p$ distance, which equals moving pixel mass in the images.  Their attack follows the PGD attack but projects $x$ onto a Wasserstein ball rather than $l_{\infty}$ ball and uses projected Sinkhorn iteration (\textbf{PSI}) to accelerate the computation. Compared to the $l_p$-based perturbation that indiscriminately affects both foreground and background pixels, Wasserstein perturbation only changes the foreground for images with monochrome backgrounds.

        \subsubsection{Colorization-transformation-based attack}
        Besides manipulating images in the spatial domain, some works transform images into different color spaces and then modify the color space instead, for humans prefer to classify images based on geometry rather than colors. These attacks define the perturbation as "content-preserving perturbation", which is usually measured by perceptual similarity \cite{wang2021demiguise, aydin2021imperceptible} rather than $l_p$ distance.
        
        In 2018, Hosseini \textit{et al.} \cite{hosseini2018semantic} first proposed \textbf{SemanticAdv}. It transforms the original image from RGB to HSV domain and leverages random search to modify the Hue and Saturation components. However, their adversarial examples may look factitious because hue and saturation components are modified at the same time. \textbf{ColorFool} \cite{shamsabadi2020colorfool} expanded SemanticAdv to generate more realistic pictures by conducting content-based segmentation on target images and revising the interesting and unimportant regions, respectively. They modify the $a$ and $b$ channels in the $Lab$ color space and leave the lightness channel $L$ unchanged. 
        In 2019, Laidlaw \textit{et al.} \cite{laidlaw2019functional} proposed \textbf{ReColorAdv} attack, which computes the flow field in RGB or CIELUV color space. They also define a smooth loss to improve imperceptibility. In addition, they prove that combining the color space distortion with additive noises can expand the perturbation space. Zeng \textit{et al.} \cite{zeng2019adversarial} firstly edited the 3D object's physical properties, such as affine transformation, color, and illumination, and then rendered it into an image to fool the classifier. They optimize the physical variables by FGSM for derivable factors or the zeroth-order optimization approach for non-derivable factors. Bhattad \textit{et al.} \cite{bhattad2019unrestricted} generated unrestricted AEs through semantic perturbation like colorization (\textbf{cAdv}) and texture transfer (\textbf{tAdv}). For cAdv attack, a pre-trained colorization network \cite{zhang2016colorful} is used to modify the color. They attack the input hints and masks at the same time to make the color more realistic. For the tAdv attack, they extract the texture from the target image using a VGG19 network and leverage an extra restraint on the cross-layer gram matrices to control the texture transfer's strength. Experiments show that they can generate photorealistic AEs.
        In 2020, Zhao \textit{et al.} \cite{zhao2020towards} proposed \textbf{PerC-C\&W} and \textbf{PerC-AL} attack. Their attacks directly operate in the RGB color space and use CIEDE2000 as the perceptual color distance. PerC-AL improves the optimizing efficiency by alternatively minimizing the adversarial loss and perceptual loss. Experiments show that although the perturbation size is slightly higher than C\&W method in $L_2$ metric, it is still imperceptible for human eyes and achieves better transferability.
        In 2012, inspired by the stAdv attack, Aydin \textit{et al.} \cite{aydin2021imperceptible} proposed a more imperceptible attack called \textbf{Spatial Chroma-shift}. The basic idea is that humans are less sensitive to color changes than brightness changes. Therefore, they only apply the spatial transformation to the colorspace. They transform the color space from RGB to YUV and then only compute the flow field on UV channels. Moreover, they use LPIPS and DISTS to evaluate the perceptual similarity and show that their method outperforms C\&W and stAdv attacks.
       
        \subsection{Black-box adversarial attack for 2D deep learning models}
        All the above attacks generated AEs in the white-box assumption. However, in reality, the adversary often has little prior knowledge of the model internal architecture and parameters. Black-box attacks are proposed to solve this problem. 
        
        \subsubsection{Different black-box scenarios}    
        According to the information the adversary can access, the black-box settings can be classified into the \textbf{query-constrained scenario, score-based scenario, and hard label scenario} \cite{ilyas2018black}. For the first circumstance, the attacker can only query the victim model a limited number of times. For example, some APIs may charge fees if the adversary queries many times. For the second circumstance, confidence scores are available, such as the Google Cloud Vision API. In the last circumstance, only top-1 or top-k predicted labels are output. For example, the Apple Photo app can automatically classify the user's pictures without displaying score information.

        \begin{table}
        \scriptsize
        \centering
          \caption{Summary of main \textbf{black-box} adversarial attacks in 2D CV tasks sorted by the algorithm and published year}
          \label{tab-blackbox}%
          \begin{threeparttable}
          \setlength{\tabcolsep}{3mm}{
            \begin{tabular}{p{4.945em}lp{2em}p{1.945em}p{3.165em}p{2.445em}p{4.445em}lp{8.665em}}
            \toprule
            \multirow{2}[4]{*}{Attacks} & \multicolumn{1}{l}{\multirow{2}[4]{*}{Year}} & \multicolumn{2}{p{8em}}{Threat Model} & \multirow{2}[4]{*}{Algorithm} & \multirow{2}[4]{*}{Distance} & \multicolumn{2}{p{6.725em}}{Performance} & \multirow{2}[4]{*}{Key idea} \\
            \cmidrule{3-4}\cmidrule{7-8}\multicolumn{1}{l}{} &    & Goal & Knowl.\tnote{*} & \multicolumn{1}{l}{} & \multicolumn{1}{l}{} & Efficiency\tnote{**} & \multicolumn{1}{p{2.28em}}{ASR\tnote{***}} & \multicolumn{1}{l}{} \\
            \midrule
            TML \cite{papernot2016transferability} & 2016 & U  & $\blacksquare$ & \multicolumn{1}{l}{Subs. model} & $L_\infty$ & Costly & 96.19\% & Data augment. \\
            \multicolumn{1}{l}{TIMI \cite{dong2019evading}} & 2019 & \multicolumn{1}{l}{U} & $\blacksquare$ & \multicolumn{1}{l}{Subs. model} & \multicolumn{1}{l}{$L_2$} & \multicolumn{1}{l}{Efficient} & 49.00\% & \multicolumn{1}{l}{Tranform-invariant} \\
            \multicolumn{1}{l}{FDA \cite{ganeshan2019fda}} & 2019 & \multicolumn{1}{l}{U} & $\blacksquare$ & \multicolumn{1}{l}{Subs. model} & \multicolumn{1}{l}{$L_\infty$} & \multicolumn{1}{l}{Costly} & 80.20\% & \multicolumn{1}{l}{Feature distortion} \\
            \multicolumn{1}{l}{ILA \cite{huang2019enhancing}} & 2019 & \multicolumn{1}{l}{U} & $\blacksquare$ & \multicolumn{1}{l}{Subs. model} & \multicolumn{1}{l}{$L_\infty$} & \multicolumn{1}{l}{Costly} & 85.80\% & \multicolumn{1}{l}{Feature distortion} \\
            \multicolumn{1}{l}{SIM \cite{lin2019nesterov}} & 2019 & \multicolumn{1}{l}{U} & $\blacksquare$ & \multicolumn{1}{l}{Subs. model} & \multicolumn{1}{l}{$L_\infty$} & Medium & 77.20\% & \multicolumn{1}{l}{Scale-invariant} \\
            \multicolumn{1}{l}{TTTA \cite{li2020towards}} & 2020 & \multicolumn{1}{l}{T} & $\blacksquare$ & \multicolumn{1}{l}{Subs. model} & \multicolumn{1}{l}{Poincare} & \multicolumn{1}{l}{Efficient} & 42.90\% & \multicolumn{1}{l}{Pointcare distance} \\
            \multicolumn{1}{l}{RDI \cite{zou2020improving}} & 2020 & \multicolumn{1}{l}{T\&U} & $\blacksquare$ & \multicolumn{1}{l}{Subs. model} & \multicolumn{1}{l}{$L_\infty$} & \multicolumn{1}{l}{Costly} & 67.80\% & \multicolumn{1}{l}{Multi-scale gradient} \\
            \multicolumn{1}{l}{DI-MI-TI \cite{zhao2021success}} & 2021 & \multicolumn{1}{l}{T} & $\blacksquare$ & \multicolumn{1}{l}{Subs. model} & \multicolumn{1}{l}{$L_\infty$} & \multicolumn{1}{l}{Costly} & 62.20\% & \multicolumn{1}{l}{Enough iteration} \\
            \multicolumn{1}{l}{FIA \cite{wang2021feature}} & 2021 & \multicolumn{1}{l}{U} & $\blacksquare$ & \multicolumn{1}{l}{Subs. model} & \multicolumn{1}{l}{$L_\infty$} & \multicolumn{1}{l}{Costly} & 83.50\% & \multicolumn{1}{l}{Feature importance} \\
            VT\cite{wang2021enhancing} & 2021 & U  & $\blacksquare$ & \multicolumn{1}{l}{Subs. model} & \multicolumn{1}{l}{$L_\infty$} & Medium & 76.50\% & Gradient variance \\
            NAA\cite{zhang2022improving} & 2022 & U  & $\blacksquare$ & \multicolumn{1}{l}{Subs. model} & $L_\infty$ & Costly & 85.0\% & \multicolumn{1}{l}{Feature importance} \\
            ODI\cite{byun2022improving} & 2022 & T  & $\blacksquare$ & \multicolumn{1}{l}{Subs. model} & $L_\infty$ & Efficient & 81.6\% & 3D render \\
            \multicolumn{1}{l}{Img2video \cite{wei2022cross}} & 2022 & \multicolumn{1}{l}{U} & $\blacksquare$ & \multicolumn{1}{l}{Subs. model} & $L_\infty$ & Costly & 77.88\% & \multicolumn{1}{l}{Cross-modality} \\
            GNAE\cite{zhao2018generating} & 2018 & U  & $\blacksquare$ & GAN & $L_2$ & Costly & 78.00\% & Natural AE \\
            AdvGAN\cite{xiao2018generating} & 2018 & T  & $\blacksquare$,$\color{gray}\blacksquare$ & GAN & $L_2$ & Efficient & 92.76\% & Distilled model \\
            ATTA\cite{wu2021improving} & 2021 & U  & $\blacksquare$ & \multicolumn{1}{l}{GAN} & \multicolumn{1}{l}{$L_\infty$} & Costly & 61.80\% & Adv transform \\
            \multicolumn{1}{l}{Boundary Attack \cite{brendel2017decision}} & 2017 & \multicolumn{1}{l}{T\&U} & $\blacksquare$ & \multicolumn{1}{l}{Decision } & \multicolumn{1}{l}{$L_0$} & Costly & $-$ & \multicolumn{1}{l}{Random walk} \\
            \multicolumn{1}{l}{BiasedBA \cite{brunner2019guessing}} & 2019 & \multicolumn{1}{l}{T\&U} & $\blacksquare$ & \multicolumn{1}{l}{Decision } & \multicolumn{1}{l}{$L_2$} & \multicolumn{1}{l}{Efficient} & 85.00\% & \multicolumn{1}{l}{Biased sampling} \\
            OPT\cite{cheng2019query} & 2019 & T\&U & $\blacksquare$ & \multicolumn{1}{l}{Decision } & $L_2$ & Medium & 100.00\% & Binary search \\
            \multicolumn{1}{l}{HopSkipJump \cite{chen2020hopskipjumpattack}} & 2020 & \multicolumn{1}{l}{T\&U} & $\blacksquare$ & \multicolumn{1}{l}{Decision } & \multicolumn{1}{l}{$L_2$,$L_\infty$} & \multicolumn{1}{l}{Efficient} & 60.00\% & \multicolumn{1}{l}{Grandient+BA} \\
            \multicolumn{1}{l}{SignOPT \cite{cheng2020sign}} & 2020 & \multicolumn{1}{l}{T\&U} & $\blacksquare$ & \multicolumn{1}{l}{Decision } & $L_2$ & \multicolumn{1}{l}{Efficient} & 94.00\% & \multicolumn{1}{l}{Estimate gradient sign} \\
            \multicolumn{1}{l}{RayS \cite{chen2020rays}} & 2020 & \multicolumn{1}{l}{U} & $\blacksquare$ & \multicolumn{1}{l}{Decision } & \multicolumn{1}{l}{$L_\infty$} & \multicolumn{1}{l}{Efficient} & 99.80\% & \multicolumn{1}{l}{Early stop} \\
            QAIR\cite{li2021qair} & 2021 & U  & $\blacksquare$ & \multicolumn{1}{l}{Decision } & \multicolumn{1}{l}{$L_\infty$} & Efficient & 98.00\% & Image retrieval \\
            \multicolumn{1}{l}{BayesAttack \cite{shukla2021simple}} & 2021 & \multicolumn{1}{l}{T\&U} & $\blacksquare$ & \multicolumn{1}{l}{Decision } & \multicolumn{1}{l}{$L_2$,$L_\infty$} & Efficient & 67.48\% & \multicolumn{1}{l}{Bayesian optimization} \\
            
            \multicolumn{1}{l}{Surfree \cite{maho2021surfree}} & 2021 & \multicolumn{1}{l}{T\&U} & $\blacksquare$ & \multicolumn{1}{l}{Geometric} & \multicolumn{1}{l}{$L_2$} & Efficient & 90.00\% & \multicolumn{1}{l}{Orthogonal projection} \\
            
            \multicolumn{1}{l}{TangentAttack \cite{ma2021finding}} & 2021 & \multicolumn{1}{l}{T\&U} & $\blacksquare$ & \multicolumn{1}{l}{Geometric} & \multicolumn{1}{l}{$L_2$} & Efficient & - & \multicolumn{1}{l}{Semi-ellipsoid} \\
            
            \multicolumn{1}{l}{TriangletAttack \cite{ma2021finding}} & 2022 & \multicolumn{1}{l}{T\&U} & $\blacksquare$ & \multicolumn{1}{l}{Geometric} & \multicolumn{1}{l}{$L_2$} & Efficient & 44.5\% & \multicolumn{1}{l}{Triangle inequality} \\

            ZOO\cite{chen2017zoo} & 2017 & T  & $\blacksquare$ & Score & $L_2$ & Costly & 97.00\% & Finite difference \\
            LocSearchAdv\cite{narodytska2017simple} & 2017 & T\&U & $\blacksquare$, $\color{gray}\blacksquare$ & Score & $L_0$ & Costly & 70.78\% & Local Search \\
            Autozoom\cite{tu2019autozoom} & 2018 & T  & $\blacksquare$ & Score & $-$ & Efficient & 93.00\% & Autoencoder \\
            PCA\cite{bhagoji2018practical} & 2018 & U  & $\blacksquare$ & Score & $L_\infty$   & Costly & 89.50\% & Principal component \\
            SimBA\cite{guo2019simple} & 2019 & T\&U & $\blacksquare$ & Score & $L_2$ & Medium & 96.50\% & Orthonormal basis \\
            CornerSearch\cite{croce2019sparse} & 2019 & U  & $\blacksquare$ & Score & $L_0$ & Costly & 99.56\% & Sparse noises \\
            OnePixel\cite{su2019one} & 2019 & T\&U & $\blacksquare$, $\color{gray}\blacksquare$ & Score & $L_0$ & Costly & 67.97\% & Differential evolution \\
            PBBA\cite{moon2019parsimonious} & 2019 & T\&U & $\blacksquare$ & Score & \multicolumn{1}{l}{$L_\infty$} & Medium & 99.90\% & Discrete surrogate \\
            \multicolumn{1}{l}{Nattack \cite{li2019nattack}} & 2019 & \multicolumn{1}{l}{U} & $\blacksquare$ & Score & \multicolumn{1}{l}{$L_2$,$L_\infty$} & \multicolumn{1}{l}{Costly} & 100.00\% & \multicolumn{1}{l}{Adversarial distribution} \\
            \multicolumn{1}{l}{Bandis \cite{ilyas2018prior}} & 2019 & \multicolumn{1}{l}{U} & $\blacksquare$ & Score & \multicolumn{1}{l}{$L_2$,$L_\infty$} & Medium & 95.40\% & \multicolumn{1}{l}{Gradient priors} \\
            \multicolumn{1}{l}{SignHunter\cite{al2019there}} & 2020 & \multicolumn{1}{l}{T\&U} & $\blacksquare$ & Score & \multicolumn{1}{l}{$L_\infty$,$L_2$} & \multicolumn{1}{l}{Efficient} & 91.47\% & \multicolumn{1}{l}{Grad. sign estimator} \\
            \multicolumn{1}{l}{Square attack \cite{andriushchenko2020square}} & 2020 & \multicolumn{1}{l}{T\&U} & $\blacksquare$ & Score & \multicolumn{1}{l}{$L_2$,$L_\infty$} & \multicolumn{1}{l}{Efficient} & 99.40\% & \multicolumn{1}{l}{Shrinking squares} \\
            \multicolumn{1}{l}{Sparse-RS \cite{croce2022sparse}} & 2022 & \multicolumn{1}{l}{T\&U} & $\blacksquare$ & Score & $L_\infty$ & Costly & 95.80\% & \multicolumn{1}{l}{Random search} \\
            NES\cite{ilyas2018black} & 2018 & T\&U & $\blacksquare$ & Score/Decision & $L_\infty$   & Costly & 88.20\% & Natural evolution \\
            Subspace\cite{guo2019subspace} & 2019 & U  & $\blacksquare$ & \multicolumn{1}{l}{Subs.+Score} & $L_\infty$   & Costly & 96.60\% & Multi-subs. model \\
            \multicolumn{1}{l}{P-RGF\cite{cheng2019improving}} & 2019 & \multicolumn{1}{l}{U} & $\blacksquare$ & \multicolumn{1}{l}{Subs.+Score} & \multicolumn{1}{l}{$L_2$} & \multicolumn{1}{l}{Medium} & 99.10\% & \multicolumn{1}{l}{Transfer-based priors} \\
            SimBA++\cite{yang2020learning} & 2020 & U  & $\blacksquare$ & Subs.+Score & $L_2$ & Efficient & 99.40\% & Mixed method \\
            \bottomrule
            \end{tabular}%

       }
      \begin{tablenotes}
        \footnotesize
        \item{*} This column contains different attacks' adversarial knowledge. $\square$: white-box. $\blacksquare$: black-box. ${\color{gray} \blacksquare}$: gray-box.
        \item{**} For query-based attacks, we use efficiency to evaluate the queries needed to find the adversarial example.
        \item{***} ASR is the abbr. of attack success rate. As mentioned, we only count the best result of the hardest attack reported in the paper.
      \end{tablenotes}
      \end{threeparttable}
    \end{table}%
    
        \subsubsection{Categories of black-box 2D digital adversarial attacks}
       Different settings result in various methods. This review classifies black-box attacks into three different categories according to their settings and corresponding algorithms. 
       \begin{itemize}
            \item Substitute-model-based attacks. This method utilizes the transferability of adversarial examples, which can convert black-box problems into white-box problems, therefore significantly improving efficiency. However, limited by the transferability of the surrogate model, the attack success rate is usually not very high.
            \item Score-based query attacks. We classify query-based attacks into score-based and decision-based. The continuous confidence score can be accessed for score-based attacks to estimate the decision boundary or gradient.
            \item Decision-based query attacks. These attacks suppose the adversary can only access the predicted labels of the model. They are far more difficult than score-based attacks because less information can be leveraged.
            \item Geometry-based query attacks. These attacks also aim at hard-label settings. However, they use geometric information of the decision boundary rather than estimate the gradients to improve the query efficiency.
            \item Generative-model-based attacks. GAN can also be used for black-box attacks. For example, in AdvGAN \cite{xiao2018generating}, the classifier is displaced by a distilled model, which can be viewed as a transfer-based attack. In GNAE \cite{zhao2018generating}, random search algorithms are operated in the $z$ latent space, which can be regarded as a query-based attack. 
        \end{itemize} 
       Some attacks also combine different methods together to improve the efficiency and attack success rate, such as SimBA++ \cite{yang2020learning}, P-RGF \cite{cheng2019improving}, and Subspace attacks \cite{guo2019subspace}. Table.\ref{tab-blackbox} shows some representative 2D black-box adversarial attacks. 
       
        
        \subsubsection{Substitute-model-based black-box attacks.}
        The substitute-model-based attack is based on the transferability of adversarial examples. Therefore, improving the transferability is crucial for this kind of method. 
        
        Data augmentation and transformation are widely used for improving transferability. \textbf{TML} \cite{papernot2016transferability} used Jacobian-based dataset augmentation to expand the dataset and train the substitute model alternatively,
        which can be formulated as $\mathcal{D}_{\rho+1} = (x+\lambda_\rho \text{sign}(J_{\hat{\mathcal{F}}}(x:x \in \mathcal{D}_\rho))) \cup \mathcal{D}_\rho$, where $\hat{\mathcal{F}}$ is the surrogate model. $\mathcal{D}_{\rho}$ and $\mathcal{D}_{\rho+1}$ is the training dataset before and after $\rho^{th}$ iteration, $\lambda_\rho$ is a periodical step size. 
        In addition, they introduced reservoir sampling into the data augmentation process to reduce the iteration number. \textbf{DI} \cite{xie2019improving} promoted transferability through randomly resizing and padding the dataset. \textbf{TIMI} \cite{dong2019evading} improved the transferability of MI-FGSM by enriching the input varieties, which is equivalent to multiplying the gradient with a smoothing kernel, such as a rotation, transition, or rescale matrix. \textbf{SIM} \cite{lin2019nesterov} used Nesterov accelerated gradient and resized the image into different scales to prevent the local model from overfitting.
        \textbf{RDI} \cite{zou2020improving} improved DI and TIMI through multi-scale gradient and region fitting.
        
        Targeted attacks are more difficult than untargeted attacks for surrogate-model-based methods. Zhao \textit{et al.} \cite{zhao2021success} found that a simple targeted function $Z_{y'}(F(x^{adv}))$ with sufficient iterations can produce SOTA results, where $Z_{y'}$ is the output logit of target label. Byun \textit{et al.} \cite{byun2022improving} improved the transferability of the targeted attack through object-based diverse input (\textbf{ODI}), which utilized a differentiable 3D renderer to render the 2D adversarial examples on 3D objects. Noticing that the gradient of MI-FGSM tends to be dominated by the past gradients after a few updates, Li \textit{et al.} \cite{li2020towards} proposed \textbf{TTTA} that adaptively adjusts the gradient through Pointcare distance. 
        Through Pointcare distance, the gradient increases when and only when the $f(x)$ comes closer to the target class.

        Some works found that manipulating the surrogate model's latent layer features can alleviate overfitting and utilize the shared features, resulting in transferability improvement. Intermediate level attack (\textbf{ILA}) \cite{huang2019enhancing} improved C\&W and FGSM attacks by maximizing the distortion of a pre-specified layer, while Feature disruptive attack (\textbf{FDA}) \cite{ganeshan2019fda} disrupted all latent layer's output. However, these two attacks regard all neurons as equally important. The neurons that negatively influence the ground truth prediction should be amplified rather than suppressed. \textbf{FIA} \cite{wang2021feature} judged the neuron importance by the mean gradient of a set of transformed images. But this method suffers from gradient saturation. To solve this, \textbf{NAA} \cite{zhang2022improving} quantified neuron importance by computing neuron attribution. 
        Besides image classification tasks, \textbf{Img2video} attack \cite{wei2022cross} attacked the black-box video model by generating the AE of each frame on a surrogate model. They maximize the distance of low-level features between normal and adversarial video frames.

        Some attacks combine the surrogate model and other methods to get better performance. \textbf{P-RGF} attack \cite{cheng2019improving} utilized transfer-based priors to get more accurate gradient estimation and save queries.
        They formulated the gradient estimation problem as $\min_{\hat{g}} L(\hat{g})=\mathbb{E}\left\|\bigtriangledown_x f(x)-b\hat{g}\right\|_2^2$, where $b$ is a scaling factor. 
        They utilized the estimated cosine similarity between the actual and transferred gradient to control the transfer strength. Their method can also be incorporated with other priors to save queries. \textbf{Subspace attack} \cite{guo2019subspace} shrank the search space of stochastic vectors by transferring the gradients from a group of substitute networks. \textbf{SimBA++} \cite{yang2020learning} also integrated transferability-based and queries-based model. Moreover, SimBA++ updated the surrogate model based on the query results.

        \subsubsection{Score-based query attacks} The goal of query-based attacks is to craft adversarial examples through fewer queries and achieve a higher attack success rate. Based on whether the continuous confidence score is accessible,query-based methods can be classified into score-based and decision-based query methods.  
        
        \paragraph{Gradient-estimation-based methods}
        Most of the score-based query attacks are based on gradient estimation. The simplest gradient estimation method is the \textit{finite difference}, such as zero-order optimization attack (\textbf{ZOO}) \cite{chen2017zoo}. For ZOO, the gradient is estimated through $\frac{\partial f(x)}{\partial x_i} \approx \frac{f(x+\delta e_i)-f(x-\delta e_i)}{2\delta}$, where $e_i$ is a standard basis, $\delta$ is a  random vector sampled from Gaussian distribution near the original data point. Moreover, they used a resize function and increased dimension gradually to improve efficiency. However, because they need to use all the standard basis vectors in each iteration, the query cost is proportional to the image size, which is computationally expensive. Later, some improvements were proposed to reduce the random vectors to be sampled. \textbf{PCA} attack \cite{bhagoji2018practical} reduced search dimensions by analyzing the principal components of input data. \textbf{Autozoom} \cite{tu2019autozoom} firstly trained an Autoencoder to encode the image of the target class into latent space. Then, it sampled the random vectors from the latent space and mapped them to random perturbations with reduced dimensions to compute the finite difference. Al-Dujaili \textit{et al.} \cite{al2019there} proposed a sign-based gradient estimation algorithm. The basic idea is to estimate the sign of the directional derivative. They proposed a divide and conquer method named \textbf{SignHunter} that leverages the separability of the derivative of the adversarial objective function,  which decreases the query number from $2^n$ to $O(n)$.

        \textit{Heuristic algorithms} can also be used for gradient estimation. Ilylas \textit{et al.} \cite{ilyas2018black} estimated the gradient through a variant of natural evolution strategy (\textbf{NES}). For scoreless settings, they leveraged noise robustness to substitute the confidence score. Specifically, they sampled a set of $\delta_i$ from a normal distribution by antithetic sampling. Then, the expectation of estimated gradient of $F(x')$ approximately equals to $\frac{1}{\sigma n}\sum_{i=1}^{n}\delta_i F(x'+\sigma\delta_i)$. Later, Ilylas \textit{et al.} \cite{ilyas2018prior} formulated the gradient estimation problem as finding a vector to maximize $\mathbb{E}[\hat{g}^Tg^{*}]$ and used the least squares method to solve this problem and proved it is an equivalence of NES. They utilized two kinds of priors to improve efficiency. Firstly, the present gradient highly correlates with the gradient of the last step. Secondly, adjacent pixels often have similar gradients. Therefore, they designed a gradient estimation framework based on \textbf{bandit optimization}, where the action is a priors-based gradient estimation in each round. 

        \paragraph{Greedy-algorithm-based methods}     
         \textbf{SimBA} \cite{guo2019simple} is a simple but powerful attack based on the greedy algorithm. Instead of computing all basis vectors at each iteration (like ZOO attack), it randomly chooses one vector from a predefined orthogonal basis at each iteration. If this vector can reduce the confidence score, it is applied to the image, and the next random vector is sampled. Otherwise, this direction is deserted. They found that the DCT basis is especially efficient. Although this method is simple, it outperforms NES and ZOO in the aspect of query cost. \textbf{SimBA++} \cite{yang2020learning} improved the query efficiency by combining SimBA with the transferability-based attack TIMI \cite{dong2019evading}. Instead of randomly sampling changing direction, they correlate the sampling probability to the surrogate model's gradient. Moreover, they proposed high-order gradient approximation (HOGA) to distill the target model in both forward and backward steps. SimBA++ greatly decreases the query times and achieves a higher attack success rate than SimBA.

         \paragraph{Grid-search-based methods}
        $\mathcal{N}$\textbf{attack} \cite{li2019nattack} tries to find a distribution of AEs in the neighborhood of $x$. But it does not need to access the model parameters. It formulates the optimization problem of finding adversarial distribution around $x$ as $\min_{\theta}J(\theta):= \int{f(x')\pi_S(x'|\theta)dx'}$, where $f(x')$ is an untargeted attack loss function, $\theta =(\mu,\sigma^2)$ are the parameters of Gaussian distribution and $x'=x+\text{proj}_S(1/2(tanh(z)+1))$, where $z \sim \mathcal{N}(z|\theta)$. To solve this problem, it uses grid search to find the best $\sigma$ and updates $\mu$ through NES.        
        \textbf{PBBA} \cite{moon2019parsimonious} searched AEs along the border of $l_\infty$ ball. This simplification makes it possible to refine the query performance by optimizing discrete surrogate problems. It firstly zones the images as squares and operates regional optimization in the grid by choosing $x'$ from $\{x-\epsilon,x+\epsilon\}$. Then, it adjusts the squares and duplicates this procedure until it finds an AE.
        \textbf{Square attack} reduced the search space through diminishing squares. Unlike PBBA \cite{moon2019parsimonious} that used a fixed grid, the locations of squares are optimized. Moreover, they sampled $\delta$ along the boundary of $L_p$-norm ball to improve the query efficiency. Specifically speaking, for $L_\infty$-square attack, $x_i^{\prime}\in \{x_i-\epsilon,x_i+\epsilon\}$. For $L_2$-square attack, because noise values are correlated, they increase the budget of one window and simultaneously decrease the budget of another.

        \paragraph{Dimension-reduction-based methods}        
        Some works reduced the search dimensions by reducing the number of crafted pixels. The first work is \textbf{LocSearchAdv} \cite{narodytska2017simple}, which used local search to estimate implicit gradient. It first randomly picked up some pixels as initial points. At each round, the adjacent pixels of these points are checked and updated. It only needed to manipulate 0.5\% pixels for the untargeted attack on ImageNet. \textbf{CornerSearch} \cite{croce2019sparse} also used local search to find sparse perturbations, but it only added perturbations to pixels with high variation to make them imperceptible. \textbf{OnePixel} attack \cite{su2019one} considered the extreme situation and showed the possibility of fooling the classifier by only revising one pixel. It used differential evolution to optimize this problem. Each candidate solution contains its pixel coordinates and color values. The candidate solutions are updated by random crossover. \textbf{Sparse-RS} \cite{croce2022sparse} utilized random search, which is appropriate for zero-order optimization problems with sparse restraint. It achieved the SOTA attack success rate on multiple attacks, including the $l_0$-norm noises, adversarial patches, and adversarial borders.
        
        \subsubsection{Decision-based query attacks}
        In some scenarios, the adversary can only get the final decision ( hard label) rather than the confidence score (soft label). This is a more challenging case because the gradient value is hard to estimate. 

        \paragraph{Gradient-estimation-based methods}
        Some works proposed to estimate the \textit{gradient directions} instead. The first work is \textbf{Boundary attack} \cite{brendel2017decision}, which first initializes a point outside the ground truth region and then uses a random walk algorithm to make it closer to the boundary at each iteration. Although it can realize almost the same perturbation level compared with white-box attacks, it needs exponential time to find adversarial examples. \textbf{Biased boundary attack} \cite{brunner2019guessing} used bias sampling to re-understand and improve boundary attacks. Three different biases are utilized. The first bias is sampling more vectors in the direction of low-frequency distortion. The second bias is sampling denser data from regions where the adversarial and benign images have larger differences. The last bias is the prior-based gradient from the substitute model. \textbf{HSJA} \cite{chen2020hopskipjumpattack} includes three steps: binary search to find the boundary point, estimation of gradient direction, and geometric progression to update the boundary point. Because the gradient guides the search direction, HSJA significantly decreases query number compared to the Boundary attack.

        \textbf{OPT} \cite{cheng2019query} is another representative hard-label-based attack, which estimated the shortest distance of the benign sample from the boundary by fine-grained search and binary search. They estimated the gradient of the search direction through the randomized gradient-free method and updated the search direction through gradient descent. \textbf{SignOPT} \cite{cheng2020sign} promoted OPT's efficiency by only estimating the sign of gradient over search direction by just one query and averaging the gradient sign over a group of random directions. \textbf{RayS} \cite{chen2020rays} reframed the successive problem of finding the nearest classification hyperplane as a discontinuous problem without gradient estimation, which searches over a group of ray orientations. Moreover, all unnecessary directions are terminated early through a quick check. This greatly promoted search efficiency. \textbf{Bayes attacks} \cite{shukla2021simple} reduced query budgets to 1000 through Bayesian optimization. However, because Bayesian optimization is not suitable for search space with large dimensions, they reduce the search dimension through FFT and use nearest-neighbor upsampling to find adversarial examples. Their method significantly reduced the query times compared with OPT and Sign-OPT attacks. \textbf{QAIR} \cite{li2021qair} considers a more difficult setting where the attacker can only get top-k unlabeled images from the target model in the image retrieval task. They proposed a correlation-based loss by the difference of top-k unlabeled feedback retrieved by benign and adversarial images. Moreover, they improved the efficiency through recursive network theft to get gradient priors.

        \paragraph{Geometry-based methods}
        Geometry attacks also aim at the hard-label scenario. However, they directly use geometric information of the decision boundary rather than estimate the gradients to improve efficiency. \textbf{Surfree} \cite{maho2021surfree} assumes the boundary is a hyperplane and iteratively runs binary searches over orthonormal directions to find the clean image's projection on the boundary. Compared to HSJA \cite{chen2020hopskipjumpattack}, the new direction is randomly sampled from the low-frequency subband generated by DCT rather than using gradient directions, which reduces the queries to a few hundred. \textbf{TangentAttack} \cite{ma2021finding} assumed the boundary is a semi-ellipsoid and adjusted the search direction to follow the optimal tangent line rather than the gradient directions, which need less distortion than HSJA.
        \textbf{TriangleAttack} \cite{wang2022triangle} utilized triangle inequality to search the boundary point and also used DCT for dimension reduction. Less than a thousand quires are required in this attack. \textbf{CGBA} \cite{reza2023cgba} restrict the search on a semicircular path to make sure to find a boundary point regardless of the boundary curvature, which is quite efficient for untargeted attack..

        \subsubsection{Generative-model-based black-box attacks}
        GAN can also be used for black-box attacks. \textbf{GNAE} \cite{zhao2018generating} generates natural AEs by searching for perturbations in the latent space rather than the input space. It first learns a generator $\mathcal{G}_{\theta}$ that maps normally distributed variables $z$ to the input $x$. It then learns an inverter $\mathcal{I}_{\gamma}$ that maps $x$ back to latent space by minimizing the divergence between $z$ and reconstructed $\mathcal{I}_\gamma(x)$. Their objective is 
        $\min_{\tilde{z}}\left\|\tilde{z}-\mathcal{I}_\gamma(x)\right\|_2\;s.t.\;\mathcal{F}(\mathcal{G}_\theta(\tilde{z}))\neq\mathcal{F}(x)$. The feasible $\tilde{z}$ is found through iterative stochastic search and hybrid shrinking search. 
      
        Xiao \textit{et al.} \cite{xiao2018generating} proposed \textbf{AdvGAN} to improve the attack transferability. AdvGAN consists of a perturbation generator, a discriminator, and a classifier. The generator is trained to generate perturbations from the original instance. The discriminator is trained to distinguish AEs from clean instances. The classifier is a distilled model of the target black-box network. They alternately train these modules by first training the GAN and then training the classifier through 
        $
            \min_{f}\mathbb{E}_{x \sim D_{data}}\mathcal{L}(f(x), b(x))+\mathcal{L}(f(x+\mathcal{G}(x)), b(x+\mathcal{G}(x)))
        $
        , where $b(x)$ is the target model, and $f(x)$ is the distilled model. $\mathcal{L}$ is a cross-entropy loss function. Their black-box attack achieved 92.7\% on the MNIST dataset.  
        
        Some previous transfer-based attacks improved transferability through simple transformations, like rotation and translation \cite{dong2019evading}, random resizing and padding \cite{xie2019improving} and multi-scaling \cite{lin2019nesterov}. Wu \textit{et al.} \cite{wu2021improving} proposed \textbf{ATTA} to improve transferability through learning adversarial transformations that can better alleviate the adversarial property. They formulate this problem as a min-max problem. The inner problem is to find $x_{adv}$ that maximizes the classification loss. The outer problem is to learn a transformation through CNN to minimize the classification loss.         
        
   \subsection{The safety impact of physical-realizable 2D adversarial attacks}
   
        \begin{table}
        \scriptsize
        \centering
          \caption{Summary of main works that examine the safety of real-world 2D CV applications through physical adversarial attacks}
          \label{tab-physicalattack}%
          \begin{threeparttable}
          \setlength{\tabcolsep}{3mm}{
                \begin{tabular}{lllp{1.665em}p{2.72em}lllll}
                \toprule
                \multicolumn{1}{l}{\multirow{2}[4]{*}{Attacks}} & \multicolumn{1}{l}{\multirow{2}[4]{*}{Year}} & \multicolumn{3}{p{6.05em}}{Threat Model} & \multicolumn{1}{l}{\multirow{2}[4]{*}{Algorithm}} & \multicolumn{1}{l}{\multirow{2}[4]{*}{Distance}} & \multicolumn{2}{p{3.89em}}{Performance} & \multicolumn{1}{l}{\multirow{2}[4]{*}{Scenario}} \\
                \cmidrule{3-5}\cmidrule{8-9}   &    & \multicolumn{1}{p{1.665em}}{Goal} & Knowl.\tnote{*} & Interface                                                                                          &    &    & \multicolumn{1}{p{2.335em}}{Camouflage} &  Succ. \tnote{**} &  \\
                \midrule
                UAP-patch \cite{liu2020bias} & 2020 & U  & $\square$ & Patch & \multicolumn{1}{p{4em}}{Gradient} & Style loss & \multicolumn{1}{p{2.335em}}{Marked} & 74.1\% & Auto check-Out \\
                Facial Acc. \cite{sharif2016accessorize} & 2016 & \multicolumn{1}{p{1.665em}}{T\&U} & $\square$ & Patch & \multicolumn{1}{p{4em}}{Gradient} & \multicolumn{1}{p{2.78em}}{$L_\infty$} & \multicolumn{1}{p{2.335em}}{Marked} & 80.0\% & \multicolumn{1}{p{5.39em}}{Face rcg.} \\
                \multicolumn{1}{p{4.335em}}{AGN\cite{sharif2019general}} & 2019 & \multicolumn{1}{p{1.665em}}{T\&U} & $\square$ & Patch & GAN & \multicolumn{1}{p{2.78em}}{$-$} & \multicolumn{1}{p{2.335em}}{Slight} & 70.0\% & \multicolumn{1}{p{5.39em}}{Face rcg.} \\
                FRSadv\cite{nguyen2020adversarial} & 2020 & T  & $\square$ & \multicolumn{1}{l}{Light} & Gradient & $L_\infty$ & Marked & 92.0\% & Face rcg. \\
                GenAP \cite{xiao2021improving} & 2021 & \multicolumn{1}{p{1.665em}}{T\&U} & $\blacksquare$ & Patch & GAN & \multicolumn{1}{p{2.78em}}{$L_\infty$} & \multicolumn{1}{p{2.335em}}{Marked} & 65.0\% & \multicolumn{1}{p{5.39em}}{Face rcg.} \\
                AdvMakeUp \cite{yin2021adv} & 2021 & T  & $\square$ & Patch & Gradient & Style, Content & Slight & 22.0\% & \multicolumn{1}{p{5.39em}}{Face rcg.} \\
                AdvPatch \cite{brown2017adversarial} & 2017 & T  & $\square$ & Patch & \multicolumn{1}{p{4em}}{Gradient} & \multicolumn{1}{p{2.78em}}{$L_\infty$} & \multicolumn{1}{p{2.335em}}{Marked} & 93.0\% & \multicolumn{1}{p{5.39em}}{Img clas.} \\
                \multicolumn{1}{p{4.335em}}{EoT\cite{athalye2018synthesizing}} & 2018 & \multicolumn{1}{p{1.665em}}{T} & $\square$ & Patch & \multicolumn{1}{p{4em}}{Gradient} & \multicolumn{1}{p{2.78em}}{$L_2$} & \multicolumn{1}{p{2.335em}}{Slight} & 82.0\% & \multicolumn{1}{p{5.39em}}{Img clas.} \\
                CiPer\cite{agarwal2020noise} & 2020 & U  & $\blacksquare$ & \multicolumn{1}{l}{Sensor} & \multicolumn{1}{p{4em}}{Greedy} & \multicolumn{1}{p{2.78em}}{$L_\infty$} & Marked & 33.0\% & \multicolumn{1}{p{5.39em}}{Img clas.} \\
                Nat-Patch \cite{hu2021naturalistic} & 2021 & U  & $\square$ & Patch & GAN & Smoth loss & Invisible & 48.0\% & Object detect \\
                CAMOU \cite{zhang2018camou} & 2018 & U  & $\blacksquare$ & Patch & Subs. Model & $-$ & Marked & 32.7\% & Self-driving \\
                RP2 \cite{eykholt2018robust} & 2018 & T  & $\square$ & Patch & \multicolumn{1}{p{4em}}{Gradient} & $L_1,L_2$ & Marked & 84.8\% & Self-driving \\
                ShapeShifter \cite{chen2018shapeshifter} & 2018 & T\&U & $\square$ & Patch & \multicolumn{1}{p{4em}}{Gradient} & $L_2$ & Marked & 87.0\% & Self-driving \\
                MeshAdv \cite{xiao2019meshadv} & 2019 & T\&U & $\square$ & \multicolumn{1}{l}{Renderer} & \multicolumn{1}{p{4em}}{Gradient} & $Lp.$ & Slight & 100.0\% & Self-driving \\
                AdvCam\cite{duan2020adversarial} & 2020 & T\&U & $\square$ & Patch & Gradient & Style loss & Slight & 80.0\% & Self-driving \\
                DAS \cite{wang2021dual} & 2021 & U  & $\square$ & Patch & \multicolumn{1}{p{4em}}{Gradient} & $L_2, T.V.$ & Marked & 20.0\% & Self-driving \\
                AMPLE \cite{ji2021poltergeist} & 2021 & T\&U & $\blacksquare$ & \multicolumn{1}{l}{Sensor} & Bayesian & Energy loss & Marked & 87.9\% & Self-driving \\
                DirtyRoad \cite{sato2021dirty} & 2021 & U  & $\square$ & Patch & \multicolumn{1}{p{4em}}{Gradient} & $L_1, L_2, L_\infty$ & Invisible & 97.5\% & Self-driving \\
                OPAD\cite{gnanasambandam2021optical} & 2021 & T  & $\square$ & \multicolumn{1}{l}{Light} & Gradient & $L_2, L_\infty$ & Slight & 48.0\% & Self-driving \\
                DTA \cite{suryanto2022dta} & 2022 & T  & $\square$ & Patch & Gradient & $-$ & Marked & 63.0\% & Self-driving \\
                TrajAdv \cite{zhang2022adversarial} & 2022 & U  & $\square$$\blacksquare$ & Patch & Gradient,PSO & $L_\infty$ & Slight & 62.2\% & Self-driving \\
                FCA\cite{wang2022fca} & 2022 & U  & $\square$ & Patch & Gradient & Smooth loss & Marked & 60.0\% & Self-driving \\
                \bottomrule
                \end{tabular}%

       }
      \begin{tablenotes}
        \footnotesize
        \item{*} This column is the adversarial knowledge of different attacks. $\square$: white-box. $\blacksquare$: black-box. ${\color{gray} \blacksquare}$: gray-box.
        \item{**}  For objector detection, the attack success rate means the average rate of escaping from being detected. As mentioned, we only count the best result of the hardest attack reported in the paper.
      \end{tablenotes}
      \end{threeparttable}
    \end{table}%
       When deep learning models are employed in the real world, ensuring the safety of human life and property is the primary concern. A number of studies have investigated adversarial attacks in safety-critical applications, such as face recognition and self-driving systems. referred to as \textit{physical adversarial attack}. This subsection reviews 2D physical adversarial attacks in safety-critical applications. 
       
       The main challenge of physical adversarial attacks is the ever-changing physical environment, such as the background light noise, varying viewpoints, and different distances. Moreover, because the adversary cannot directly modify the input images at the pixel level, certain mediums are needed to pollute the data obliquely, such as patches \cite{sharif2016accessorize,xiao2021improving}, illumination \cite{gnanasambandam2021optical,Li_2023_physicalworld}, and sensors \cite{sayles2021invisible}. The patch-based attack is the most common physical attack. It uses printed patches to cover the whole or part of the target object to spoof the classifier. Some works also render 2D adversarial images on 3D objects, like turtles \cite{athalye2018synthesizing} and cars \cite{wang2021dual}. This can be seen as a more general form of adversarial patches. 

       \subsubsection{The safety of face recognition system}
        As mentioned before, a major challenge for the 2D physical-world adversarial attack is the changing environment, like varying distances and ambient light. To make the AE more robust to these changes and can fool the classifier consistently, Athalye \textit{et al.} \cite{athalye2018synthesizing} proposed expectation on transformation (\textbf{EoT}) that calculates the expectation of the log-likelihood of the target class on transformed images, which is formulated as 
        \begin{equation}
            x^{adv} = arg\min_{\Delta x}\Sigma^{k}_{i=1}(w_i\mathcal{L}(\mathcal{T}_i(x)+\Delta x,y)),\;s.t.\|\Delta x\|_p\leq\epsilon,
        \end{equation}
        where $\mathcal{T}_i$ is the $i_{th}$ transformation, such as random rotations and transitions, $w_i$ is the weight and subject to $\Sigma_{i=1}^{k}w_i=1$. Besides rotation and transition, they also regarded the 3D rendering as a transformation. They use the $l_2$ norm in the LAB color space as the distance metric because it is closer to human perception. Brown \textit{et al.} \cite{brown2017adversarial} inducted EOT, patch position, and multiple training images into the optimization round to generate targeted, universal, and robust adversarial patterns. However, their method has poor transferability when the pattern size is small. 
       
       In 2016, Sharif \textit{et al.} \cite{sharif2016accessorize} proposed \textbf{facial accessory attack}. The perturbation is limited to the area of facial accessories like eyeglass frames to make the attack more reasonable. They also use EOT to improve the physical attack's robustness. Moreover, they minimize the total variation of the patches to make them more natural and use the NPS score to guarantee the patch is printable. The total loss function is $ arg\min_{\delta} \sum_{x \in X}L_{s}(\mathcal{F}(x+\delta),l) + k_1 TV(\delta) +k_2 NPS(\delta)$, where $L_{s}$ is a Softmax loss on the target label. $TV$ is the total variance loss. $NPS$ is the non-printable score that defined as $NPS(\hat{p}) = \prod_{p \in P}|\hat{p}-p|$, where $P$ is the set of printable colors. In 2019, Sharif \textit{et al.} proposed \textbf{AGN} attack \cite{sharif2019general}, which improved the inconspicuousness of adversarial patches through a generative framework to generate AEs with multiple objectives, such as robustness, imperceptibility, and scalability. Unlike previous work that uses total variation as smoothness loss, they use GAN to ensure the generated patches look like real-world designs. 

      In 2020, \textbf{FRSadv} \cite{nguyen2020adversarial} attacked the face recognition system through adversarial illumination. They project adversarial patterns onto the human face, and the camera captures the images and predicts wrong labels. They also used EOT to improve the robustness. \textbf{Adv-Makeup} \cite{yin2021adv} fooled the FRS through adversarial makeup. Their attack first generates realistic eye shadow through GAN, then blends the generated eye shadow onto the source image through the gradient, content, and style losses. Adv-Makeup attacks an ensemble of models to improve transferability. \textbf{FaceAdv} \cite{shen2021effective} has a sticker generator and a converter. The generator chooses the most vulnerable area to attack, and the converter renders the patches to the face with different angles and sizes to improve physical robustness.  
      
       In 2021, Xiao \textit{et al.} \cite{xiao2021improving} proposed the \textbf{GenAP} attack, which regularizes the patches on the latent space of GAN to make the adversarial patches more natural and transferable. They first trained a StyleGAN on normal face datasets, then used it to generate AEs and crop them into the patch area. Instead of optimizing the image directly, they optimized the latent variable $w$ on the $\mathcal{W}+$ space of StyleGAN to minimize the feature distance between $x^{adv}$ and the target $x_t$. Experiments show that their methods can improve the adversarial pattern's transferability. However, their method needs a pre-trained generator trained on the victim examples. Moreover, their methods are difficult to converge, which may be because the cropping function unexpectedly changes the AEs.

         \subsubsection{The safety of self-driving system} 
          \begin{figure}[t]
        \centering  
        \includegraphics[scale=0.4]{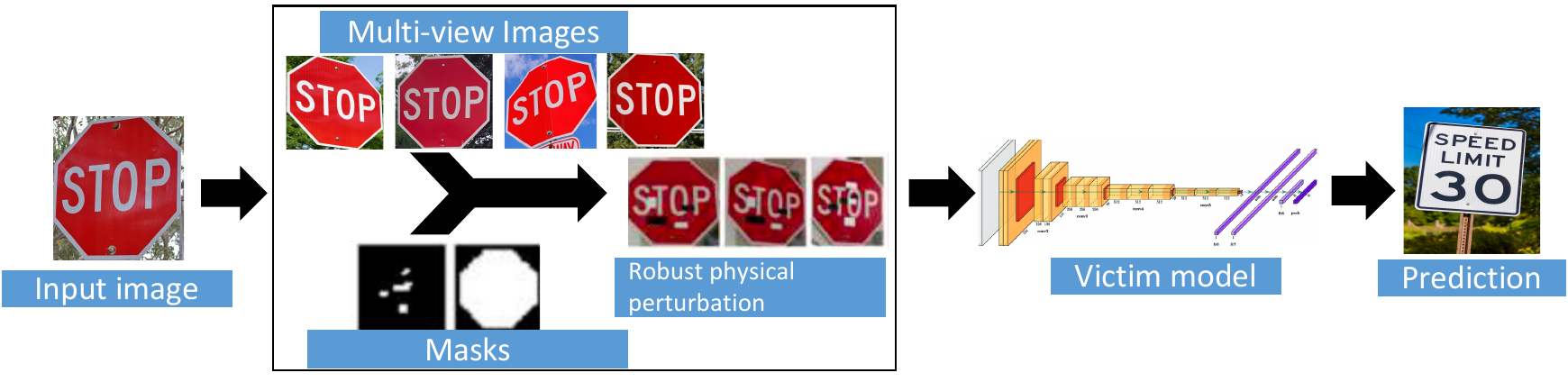}
        \caption{The general procedure of generating robust physical perturbation through expectation over transformations \cite{eykholt2018robust, athalye2018synthesizing, brown2017adversarial,sharif2016accessorize}. Images are sampled from different distances and view angles to improve the real-world attack's robustness against position changes.}
        \label{fig: RP2}    
        \end{figure} 
          A relatively simple setting for adversarial attacks in self-driving is to hide planar objects from the detector or classifier, like traffic signs and roads. 
          Chen \textit{et al.} \cite{chen2018shapeshifter} proposed \textbf{ShapeShifter} to attack the Fast R-CNN detector. They applied EoT to the objection detector field to improve the attack robustness. Rather than using the adversarial patch, they modify the background of the stop sign directly. 
          Eykholt \textit{et al.} \cite{eykholt2018robust} proposed robust physical perturbation (\textbf{RP2}) to make deep learning models misclassify the road sign. They improved EoT by sampling from a set of both simulated and physical distortions. The general procedure is shown in Figure.\ref{fig: RP2}. They first operate $L_1$ optimization to find approximate positions of the adversarial patches and then operate $L_2$ optimization to modify their RGB values. 
          \textbf{AdvCam} \cite{duan2020adversarial} generated adversarial traffic signs with natural styles through neural style transfer. The loss function consists of an adversarial loss, a style loss, a content loss, and a smoothness loss. The style loss and content loss are calculated through the features extracted from the shallow and deep layers of the feature extraction neural network. 
          Sato \textit{et al.} \cite{sato2021dirty} attacked the automated lane centering system through \textbf{dirty road patches}, which may result in the car running out of the street. Their objective function is to make the car drive out of the lane in the shortest time. They achieved 97.5\% ASR and only needed 0.903 sec for each attack. 
          \textbf{TrajAdv} \cite{zhang2022adversarial} evaluated the robustness of the trajectory prediction of the self-driving car through single-frame or multi-frame adversarial perturbation. They considered the acceleration and speed limitation to make the adversarial perturbation more natural.
          \textbf{OPAD} \cite{gnanasambandam2021optical} cheated the traffic sign classifier by optical adversarial perturbation. To overcome the nonlinear effect of the projector, they estimated the radiometric and spectral response to rectify the distortion.
        
         A more complex scenario is to hide nonplanar objects like cars from the detector. Because the 3D renderer is not differential, and the vehicle detector is usually a black box. \textbf{CAMOU} \cite{zhang2018camou} used a distilled model to learn the behavior of the 3D renderer and detector jointly. When camouflage is changed, the original distilled model may fail to fit the 3D transformations and vehicle detector. Therefore, they alternatively train the distilled model and optimize the adversarial patches. Some studies utilized a 3D differential renderer to map 3D adversarial mesh or camouflage to 2D images. Xiao \textit{et al.} \cite{xiao2019meshadv} proposed \textbf{MeshAdv} to hide cars from 2D detectors by printable 3D mesh. They leverage a differential renderer \cite{kato2018neural} to map 3D adversarial mesh to 2D images. Laplacian loss is applied on the mesh vertices to improve the smoothness. \textbf{DAS} \cite{wang2021dual} hides the car from the detector by simultaneously distracting the model and human attention. They optimize the model attention distraction loss function by minimizing the salient region and decreasing the salient region's pixel values. To make the image unnoticeable to humans, they initial the adversarial camouflage as a natural image and then constrain the perturbation in its edge area. \textbf{AMPLE} \cite{ji2021poltergeist} can hide or create a car through a sensor-injection attack by rotating and vibrating the camera to blur the image. It uses Bayesian methods to optimize the rotation and vibration parameters. To cheat the detector in multiple viewpoints, \textbf{FCA} \cite{wang2022fca} colored the adversarial camouflage onto the car's whole surface through a differential renderer. The loss function includes three parts. The first part is to cut down the IoU between the original and predicted boxes. The second part is to decrease the objectness confidence. The third part is to lower the predicted logits of the target class. Suryato \textit{et al.} \cite{suryanto2022dta} found that the differential neural renderer that previous works used fails to perform diverse physical world conversions because of an absence of domination of environment variances. Therefore, they proposed \textbf{DTA} attack to hide the car from the detector, which utilized a differential transformation network to get photorealistic images. 

         Besides face recognition and self-driving scenario, the safety of other scenarios against adversarial attacks also are investigated. Liu \textit{et al.} \cite{liu2020bias} evaluated the safety of the automatic check-out system through universal adversarial patches. They leveraged perceptual and semantic biases of models to improve the generalization ability. Hu \textit{et al.} \cite{hu2021naturalistic} leveraged StyleGAN and BigGAN to generate adversarial patches with natural-looking content to evade from person detector. They minimized the objectness score and classification confidence of the target class simultaneously. Agarwal \textit{et al.} \cite{agarwal2020noise} leveraged the noise produced by the environment and imaging process to reduce the classification accuracy of black-box models. The noises are extracted through Gaussian and Laplacian filters.

\section{Adversarial attacks for 3D deep learning models}
\label{3D attack}

    \subsection{The difference between 3D and 2D adversarial attacks}
As a result of affordable 3D data acquisition devices and the rich information provided by the geometry feature, 3D data are widely used in many safe-critical tasks. As a result, their security also attracts more and more attention.
Compared with 2D adversarial attacks, 3D adversarial attacks have some significant differences:
\begin{itemize}
    \item Compared with images, point clouds contain point-wise coordinates with unordered and irregular sampling values. Some 2D attacks that utilize the image structure cannot be directly applied to these kinds of data, such as the boundary attack, which estimates the boundary point by the weighted average of pixel value between source and target images. Interpolation between point clouds directly will deform the 3D shapes.
    \item 3D adversarial attacks have a larger disturbance space and degree of freedom than 2D adversarial attacks. In addition to modifying the value, the adversary can add and delete points.   
    \item Unlike 2D adversarial attacks, which use all pixels for classification, point clouds do not use all points. Due to the large number of points, the point cloud is usually sampled before the classification. For example, PointNet sampled 1024 points for 3D object classification on the ModelNet40 dataset.
    \item Most 2D adversarial attacks are based on pixel-wise $L_p$ distance. However, point-wise attacks can be easily defended by outlier point removal, and $L_p$ distance is unsuitable for disordered data, like the point cloud. 
    \end{itemize}
Therefore, previous cannot be directly used for 3D adversarial attacks. To solve these problems, 3D adversarial attacks are proposed for 3D data specifically. We will introduce these attacks according to their algorithms and scenarios.

    \subsection{Catalog of 3D adversarial attacks}
   We summarize the recent works of 3D digital and physical adversarial attacks in Table.\ref{tab-3D}. According to the attack algorithms used for generating adversarial examples, we classify the 3D adversarial attacks into the following categories, 
    \begin{itemize}
        \item Gradient-based optimization methods. For most 3D adversarial attacks, gradient-based methods are used. Some of them are based on $C\&W$ loss, such as \cite{xiao2018generating} and \cite{tsai2020robust}, while others are based on fast-gradient methods, such as \cite{liu2019extending}. Some works relax $L_0$-norm attack to differential versions, such as \cite{zheng2019pointcloud}.
        \item Generative-model-based methods. Some works, like AdvPC \cite{hamdi2020advpc} and LG-GAN \cite{zhou2020lg}, utilize a generative model to generate adversarial point clouds. These attacks can improve the transferability of 3D adversarial examples.       
        \item Heuristic-algorithm-based methods. In black-box settings, the gradient of the 3D model is unavailable. Hence, some black-box attacks generate adversarial examples using heuristic algorithms like evolution algorithms. 
        \item 3D-transformation-based methods. Rather than modifying the point cloud, some attacks found that 3D models are also vulnerable to rigid-body transformations in the 3D Euclidean space, like rotations and translations.
        \end{itemize}
    Recently, some works also proposed 3D physical-realizable adversarial examples by 3D printing \cite{tsai2020robust} or laser emitter \cite{cao2019adversarial2}. In addition, some studies \cite{wen2020geometry,huang2022shape} classified the 3D digital adversarial attacks into point dropping, point adding, and point shifting attacks according to different perturbation types. However, some 3D attacks are not based on the point cloud, such as 3D mesh attack \cite{cao2019adversarial} and volumetric network attack \cite{wicker2019robustness}. Therefore, we classify 3D attacks by the algorithms rather than the perturbation types.   
    
    \subsection{3D adversarial attacks in the digital world}
    
        \subsubsection{Gradient-based method}    
        Some attacks are based on 2D adversarial attacks but with new loss functions. The first work of 3D adversarial attack, \textbf{3DAdv} \cite{xiao2018generating}, is based on C\&W. It includes the independent point and adversarial cluster attacks. The first attack uses Hausdor and Chamfer distance to measure the maximum and average distance between the original and the adversarial point cloud. Chamfer distance is defined as $\mathcal{D}_C(x,x') = \frac{1}{\left\| x' \right\|} \sum_{p' \in x'}\left(\min_{p \in x}\left\|p'-p\right\|_2^2\right)$. The second attack used the Chamfer and the farthest pair-wise distance to generate an adversarial cluster. They achieved a 100\% ASR with an acceptable noise budget. However, they only evaluated the 3D-Adv attack on the PointNet network, which was proven less adversarial robust than other 3D models, like PointNet++. Liu  \textit{et.al.} \cite{liu2019extending} extended PGD to \textbf{3D-PGD}. They projected the perturbation to the tangent plane of the adversarial point cloud to reduce the outlier points. This is equivalent to changing the distribution of the surface points. However, Tsai \textit{et.al.} stated that this attack could be easily defended by a resampling method \cite{tsai2020robust}.  
        
         Some attacks discard important points or generate points with specific shapes. Zheng \textit{et al.} \cite{zheng2019pointcloud} evaluated each point's importance through \textbf{point discarding}. Because point discarding is non-differential, they relax it by moving points towards the point cloud's interior. Then, the critical score is evaluated by the directional derivative in the spherical coordinates. \textbf{ShapeAdv} \cite{liu2020adversarial} includes three kinds of perturbations: uniform distribution perturbation, adversarial sticks, and adversarial sinks. The first one generates evenly distributed perturbation through resampling during the optimization. The second one adds auxiliary features like adversarial sticks. However, because the stick's angle is difficult to optimize, they use a projection and clip function to approximate it. The last one pulls points into the inside of the point cloud. \textbf{MinimalAdv} \cite{kim2021minimal} fools the classifier by manipulating fewest points. They formulated the problem as a $L_0$ and $L_2$ optimization problem and relaxed it to $L_1$ problem. However, their method drops significantly after the resampling process because only a few points are adversarial, and most points (>95\%) are benign.
        
        Some attacks are proposed to improve the smoothness of the AEs. Tsai \textit{et.al.} \cite{tsai2020robust} used Chamfer and \textbf{kNN} distance to generate a geometry-aware AE. The kNN loss can restrict the distance between neighboring points in the point cloud and, hence, can significantly reduce outlier points. They successfully achieved targeted attacks on PointNet++ in the physical world by 3D-printed objects. However, their method results in rough surfaces. \bm{$GeoA^3$} \cite{wen2020geometry} improves the fidelity of AE by combining Chamfer distance and the consistency of local curvatures between $\mathcal{P}$ and $\mathcal{P}'$, which is measured through the direction of normal vectors. Moreover, to make $\mathcal{P}'$ more robust to resampling, they proposed $GeoA_{+}^3$ that contains a uniformity loss to promote the regularity of surface points. Then, they proposed iterative normal projection to optimize this objective function. Huang \textit{et al.} \cite{huang2022shape} proposed \textbf{shape-invariant attack} to improve the AE's surface smoothness. They shift the point along the tangent plane of the surface so that the perturbations are more imperceptible than the previous methods. Moreover, they combined the surrogate model and query-based methods to improve the black-box attack's efficiency. They first calculated a sensitivity map on the surrogate model according to the maximum gradient magnitude. Then, they shifted points according to the sensitivity ranking. This query method can reduce the query times by about 20\% compared with SimBA \cite{guo2019simple}, and SimBA++ \cite{yang2020learning} attacks.  
    
        Recently, statistical outlier removal (\textbf{SOR}) has been proposed to defend against adversarial point clouds. To break this countermeasure, \textbf{JGBA} \cite{ma2020efficient} embeds SOR into the attack algorithm. But because SOR is not differentiable, they replace it with a relaxation function using first-order approximation. Then, they optimize the original and SOR-filtered point cloud simultaneously. This attack successfully defeated SOR and SOR-based DUP-Net defenses.
    
        Some attacks aim to improve the transferability of 3D adversarial examples. \textbf{ITA} \cite{liu2022imperceptible} limits the perturbation along the direction of the normal vector and improves the transferability through adversarial transformations. The adversarial transformation composes a simple two-layer neural network and is learned through an adversarial training procedure. \textbf{AOF} \cite{liu2022boosting} improves the transferability by boosting the classification loss of low-frequency component, which is separated from the original PC by orthogonal decomposition of the graph Laplacian matrix.
             
         \begin{table}
            \scriptsize
            \centering
              \caption{Summary of main digital and physical adversarial attacks in 3D CV tasks sorted by the algorithm and published year}
              \label{tab-3D}%
              \begin{threeparttable}
              \setlength{\tabcolsep}{3mm}{    
                \begin{tabular}{llllllllll}
                \toprule
                \multicolumn{1}{l}{\multirow{2}[4]{*}{Attacks}} & \multicolumn{1}{l}{\multirow{2}[4]{*}{Year}} & \multicolumn{3}{p{4.885em}}{Threat Model} & \multicolumn{1}{l}{\multirow{2}[4]{*}{Method}} & \multicolumn{1}{l}{\multirow{2}[4]{*}{Distance\tnote{**}}} & \multicolumn{2}{p{4.61em}}{Performance} & \multicolumn{1}{l}{\multirow{2}[4]{*}{Key idea}} \\
                \cmidrule{3-5}\cmidrule{8-9}   &    & \multicolumn{1}{p{1.165em}}{Goal} & \multicolumn{1}{p{1.61em}}{Knowl\tnote{*}} & \multicolumn{1}{p{2.11em}}{Interface                                                                                         } &    &    & \multicolumn{1}{p{2.945em}}{Efficiency} & \multicolumn{1}{p{1.665em}}{ Succ. } &  \\
                \midrule
                3DAdv  \cite{xiao2018generating} & 2019 & T  & $\square$ & Digital & Gradient & $L_2$, $Cf.$ & Efficient & 99\% & Chamfer loss \\
                3D-PGD\cite{liu2019extending} & 2019 & U  & $\square$ & Digital & Gradient & $L_2$ & Efficient & 88.10\% & Gradient projection \\
                Saliency\cite{zheng2019pointcloud} & 2019 & U  & $\square$ & Digital & Gradient & $L_0$ & Efficient & 40\% & Saliency map \\
                LidarAdv \cite{cao2019adversarial} & 2019 & T\&U & $\square$   & Physical & Gradient & $L_2, Lp.$ & Costly & 71\% & Proxy function \\
                Adv-Lidar \cite{cao2019adversarial2} & 2019 & T\&U & $\square$   & Physical & Gradient & $L_0$ & Costly & 75\% & Global sampling \\
                KNN\cite{tsai2020robust} & 2020 & U\&T & $\square$ & Physical & Gradient & $Cf.,kNN$ & Efficient & 94.69\% & 3D printable \\
                Rooftop\cite{tu2020physically} & 2020 & U  & \multicolumn{1}{p{1.61em}}{$\square$    $\color{black}\blacksquare$} & Physical & Gradient & $Lp.$ & Costly & 80\% & Adversarial mesh \\
                $GeoA^3$, $GeoA_{+}^{3}$\cite{wen2020geometry} & 2020 & T  & $\square$ & Digital & Gradient & $Cf., Hd., LC$ & Efficient & 100\% & Local curvatures \\
                \multicolumn{1}{p{3.89em}}{ShapeAdv\cite{liu2020adversarial}} & 2020 & U  & $\square$ & Digital & Gradient & $L_2$ & Costly & 95\% & Shape attack \\
                JGBA \cite{ma2020efficient} & 2020 & T\&U & $\square$ $\color{gray}\blacksquare$ & Digital & Gradient & $L_2$ & Efficient & 98.90\% & Break SOR \\
                CTRI\cite{zhao2020isometry} & 2020 & T  & $\square$ & Digital & Gradient & Spectral & Efficient & 98\% & Restricted isometry  \\
                MinimalAdv\cite{kim2021minimal} & 2021 & U  & $\square$ & Digital & Gradient & $L_2,L_0,Cf.,Hd.$ & Costly & 89\% & Minimal perturbation \\
                ShapeInv\cite{huang2022shape} & 2022 & U  & \multicolumn{1}{p{1.61em}}{$\square$    $\color{black}\blacksquare$} & Digital & Gradient & \multicolumn{1}{p{1.555em}}{$L_\infty$} & Efficient & 100.00\% & Sensitivity map \\
                AOF \cite{liu2022boosting} & 2022 & U  & $\square$ & Physical & Gradient & $L_\infty$ & Costly & 99.76\% & Low frequency \\
                AdvPC \cite{hamdi2020advpc} & 2020 & U  & \multicolumn{1}{p{1.61em}}{$\square$    $\color{black}\blacksquare$} & Digital & GAN & \multicolumn{1}{p{1.555em}}{$L_\infty$} & Costly & 64.40\% & GAN \\
                LG-GAN \cite{zhou2020lg} & 2020 & T  & $\square$ & Digital & GAN & $L_2$ & Efficient & 97\% & Generative model \\
                ITA \cite{liu2022imperceptible} & 2022 & T\&U & \multicolumn{1}{p{1.61em}}{$\square$    $\color{black}\blacksquare$} & Digital & GAN & $L_2, Cf.$ & Efficient & 29.89\% & Adv. transform \\
                EvolutionAdv\cite{cao2019adversarial} & 2019 & U  & $\color{black}\blacksquare$ & Physical & Evolution & $-$ & Costly & 62\% & Evolution \\
                Camdar-adv \cite{chen2021camdar} & 2021 & T  & $\color{black}\blacksquare$ & Physical & Evolution & $Lp.$ & Costly & 99\% & Multi-modality \\
                ISO \cite{wicker2019robustness} & 2019 & U  & \multicolumn{1}{p{1.61em}}{$\square$    $\color{black}\blacksquare$} & Digital & Greedy & $L_0$ & Costly & 100\% & Critical point set \\
                OcclusionPoint\cite{sun2020towards} & 2020 & U  & $\color{black}\blacksquare$ & Physical & Greedy & $L_0$ & Efficient & 80\% & Lidar occlusion \\
                TSI\cite{zhao2020isometry} & 2020 & U  & $\color{black}\blacksquare$ & Digital & Random & $-$ & Efficient & 95\% & Thompson sampling \\
                \bottomrule
                \end{tabular}%

              }
          \begin{tablenotes}
            \footnotesize
            \item{*} This column is the adversarial knowledge of different attacks. $\square$: white-box. $\blacksquare$: black-box. ${\color{gray} \blacksquare}$: gray-box.
            \item{**} Cf.: Chamfer distance. Hd: Hausdorff distance. Lp: Laplacian distance. LC: Local curvature distance. 
          \end{tablenotes}
          \end{threeparttable}
        \end{table}
            
    \subsubsection{Generative-model-based methods} 
        Hamdi \textit{et al.} \cite{hamdi2020advpc} proposed \textbf{AdvPC} attack, which used GAN model to produce an adversarial point cloud. The loss of AdvPC has two parts. The first one is the pre-trained classifier's loss. The second is the auto-encoder loss to ensure that the reconstructed point cloud can still fool the classifier. Experiments show that AdvPC attacks can improve the transferability of adversarial point clouds. Moreover, AdvPC outperforms 3D-adv and KNN attacks on several different defense methods. However, they only evaluated untargeted attacks. 
    
        \textbf{LG-GAN} \cite{zhou2020lg} generated targeted adversarial examples through the multi-branch generative network. It first learns the multi-layer features of the input 3D data through the GAN and then unitizes a class encoder to mix the label information into the multi-layer features. At last, the 3D data are rebuilt from the features. The GAN includes a classification loss, a $L_2$-norm reconstruction loss, and a graph adversarial loss.

    \subsubsection{Heuristic-algorithm-based methods}
        In the black-box scenario, the gradient is hard to obtain. Therefore, black-box attacks are often based on heuristic algorithms. Cao \textit{et al.} \cite{cao2019adversarial} proposed an evolution-based black-box algorithm \textbf{EvolutionAdv}. They set the population as mesh vertices of the object and the adaption function as $-L(f(x^{adv}))$ and randomly add some novel individuals at each period with perturbations sampling from a normal distribution.  Wicker \textit{et al.} \cite{wicker2019robustness} proposed an Iterative Salience Occlusion attack (\textbf{ISO}) to break PointNet and volumetric networks through a greedy algorithm. They first identify the vital point set through queries. Then, they drop the most critical points iteratively until the classification result is false.         
        
    \subsubsection{3D-transformation-based methods}
    \textbf{TSI} \cite{zhao2020isometry} found that 3D models are vulnerable to affine transformations like rotations and transitions. Therefore, they proposed to attack the 3D model by isometric transformation. A random algorithm based on Thompson sampling is proposed to optimize the rotation angles. Compared with random sampling, Thompson sampling is more productive and can unitize the prior knowledge. \textbf{CTRI} \cite{zhao2020isometry} searched an adversarial isometric transformation (such as rotation) by minimizing the spectral-norm loss, which is equivalent to finding a smallest $\delta$ such that $(1-\delta)\left\|x\right\|^2 \leq \left\|Ax\right\|^2 \leq (1+\delta)\left\|x\right\|^2$, where $A$ is the rotation matrix and $x$ is the point cloud.

    \subsection{The safety impact of physical-realizable 3D adversarial attacks}
        In recent years, 3D physical adversarial attacks have been proposed to cheat Lidar-based detectors through printable mesh \cite{tu2020physically} or laser emitter \cite{cao2019adversarial2}. These attacks pose severe threats to real-world applications. We classify these attacks into simulation-based and physical-system-based 3D attacks based on their implementation methods. 
        \subsubsection{Simulation-based 3D attacks}
        Some of them are implemented in the simulation setting through a Lidar renderer. For example, Cao \textit{et al.} \cite{cao2019adversarial} proposed an optimization-based algorithm \textbf{LidarAdv} for hiding 3D mesh from Lidar detection. They first soften the preprocess function through the differential proxy function. Then, they generated adversarial mesh through $L_2$ loss. However, they only evaluated their attack on the Apollo platform in the simulation setting rather than the newest detection models. Tu \textit{et al.} \cite{tu2020physically} proposed \textbf{Rooftop} attack to hide the self-driving car from Lidar by placing an adversarial mesh on the vehicle. They optimize the initial mesh through a fooling loss and a Laplacian loss to improve its smoothness. Their vehicle-agnostic perturbation can achieve an 80\% occlusion rate. But they also only simulated their attack with CAD models and a Lidar renderer.
        \subsubsection{Physical-system-based 3D attacks}        
        Later, some attacks are proposed to attack real-world Lidar sensors. Cao \textit{et al.} \cite{cao2019adversarial2} proposed \textbf{Adv-Lidar} that can mislead a real-world self-driving system to detect nonexistent obstacles. Rather than modifying 3D mesh, they manipulate the 3D point cloud by laser emitter instead. They first analyzed a real-world self-driving system and found that previous attacks cannot create unreal barriers because of the preprocessing steps. So, they considered the preprocessing steps in the optimization. Then, they added adversarial points into the pristine point cloud in the limited angle and distance ranges. In addition, they used global sampling to avoid being caught in a local minimum. Their method successfully cheated a real-world self-driving system. Chen \textit{et al.} \cite{chen2021camdar} proposed a \textbf{multi-modality attack} that attacks Lidar and the camera simultaneously through the 3D adversarial mesh and 2D adversarial patch. They used the evolution algorithm to find 2D perturbations and used Tu \textit{et al.}'s method \cite{tu2020physically} to find 3D adversarial mesh. Sun \textit{et al.} \cite{sun2020towards} proposed a black-box attack against the Lidar detector in the self-driving setting. They noticed that the detect models are vulnerable to distant and occluded vehicle attacks, in which they can fool the detector by just a few points. Their attacks can cheat BEV-based, voxel-based, and point-based 3D object detectors.

\section{Future directions and challenges}
\label{Sec_challenge}

\subsection{Improving the transferability of adversarial examples}
A real-world feasible adversarial attack should be able to attack unseen models. Therefore, the transferability of adversarial samples has attracted more and more attention. There are several promising directions to improve the transferability of adversarial examples, including
\begin{itemize}
 \item \textit{Random/adversarial transformations.}
   These methods assume that adversarial examples that can survive random or adversarial transformations can better transfer between different models. The existing methods include isometry transformation \cite{dong2019evading}, random resizing \cite{xie2019improving}, multi-scale images \cite{lin2019nesterov}, or learnable adversarial transformation \cite{wu2021improving}. However, most of these transformations are very simple. It is still unknown why such vanilla transformations can improve the transferability and whether better transformations exist.
    
 \item \textit{Generating perturbation by generative model.} Some works \cite{xiao2018generating, xiao2021improving} proposed optimizing the latent feature can improve transferability. However, in our experiment, we find that searching adversarial perturbation on the latent space may cause the optimization to be unstable or non-converging. So, there is still room for improving the generative model's architecture and loss function.
    
 \item \textit{Manipulating the latent layer features.}  Some works \cite{ganeshan2019fda,huang2019enhancing, wang2021feature, zhang2022improving} maximize the middle layer features' difference between the pristine and adversarial images to improve the transferability. But Zhang \textit{et al.} \cite{zhang2022improving} have shown that these methods also have great promotion space.
 
 \item \textit{Ensemble-based approaches.}
    Some works also improve transferability by attacking an ensemble of models \cite{dong2018boosting,yin2021adv}. However, training multiple substitute models is computationally expensive. How to train ensemble models efficiently and how to design or select substitute models are still open problems.
\end{itemize}
Moreover, at present, most of these methods improve transferability empirically. Few of them proved a lower bound of the transferability or explained the mechanism behind the transferability. One possible way of understanding transferability is through the similarity of decision boundaries. For example, 
Liu \textit{et al.} \cite{liu2016delving}, and Tram\`er \textit{et al.} \cite{tramer2017space} analyzed the decision boundary of different models and contributed the transferability to their similar boundaries. However, this theory cannot explain the asymmetry of transferability \cite{wu2018understanding}. 
Ilyas \textit{et al.} \cite{ilyas2019adversarial} suggests adversarial perturbation is non-robust features rather than noise, and different models may learn the same non-robust features. Therefore, the AEs can transfer between them.  However, they only prove their hypothesis on a simple dataset with two classes and cannot explain why sometimes the two models predict the same AE as different false labels. Finding more reasonable theoretical interpretations of transferability is still an urgent task for robust deep learning.

\subsection{Towards semantic perturbation}
Imperceptible perturbation by $L_p$-norm distance is vulnerable to physical variables, like distance and background light. Therefore, some works proposed to fool the classifier while maintaining semantic consistency for human beings. At present, the main directions and challenges of semantic perturbation include:
\begin{itemize}
    \item \textit{Color-space transformations.} Some studies claimed that the DNN is biased toward texture, while human beings like to classify objects according to their shapes \cite{geirhos2018imagenet}. Therefore, some works operated color distortion in different color spaces to generate semantic-preserving perturbation. However, how to better measure the perceptual distance to be consistent with human perception is still an open problem.
    \item \textit{Global/Local spatial transformations.} Global and local spatial transformations can also generate adversarial examples while reserving semantics. However, because it's hard to compute gradients regarding the transformation parameters, there is still a demand to find an efficient way to optimize the transformation parameters. 
    \item \textit{Manipulating image content attributes.} Some attacks utilized the conditional GAN to modify the image content attributes. However, better disentangling the latent features and controlling the attributes still need research.
    \item \textit{Attacks based on diffusion model.} Recently, the diffusion model \cite{ho2020denoising} has been proposed to produce high-quality images, which has outperformed GAN on some tasks. A few works have used them for generating unconstrained semantic AEs \cite{chen2023content}. But these works usually estimate the gradients roughly. A more precise method to calculate the gradients, like the stochastic differential equation, may improve the performance.
\end{itemize}
In addition, although many works explore the semantic-reserving perturbations, few of them tried to apply themselves to the physical world to bypass the camera system. There still is a question of whether these semantic perturbations can behave consistently in the physical setting. 

\subsection{Making adversarial examples physically achievable}
Generating physical-realizable AEs that are robust enough to physical condition changes
and hard to notice simultaneously is still an open question. Physical attacks are more difficult than digital attacks because of the changing physical variables like view angles and distance. Some work \cite{athalye2018synthesizing,brown2017adversarial} proposed to improve the attack robustness through expectation over transformation (EoT). However, EoT significantly improves the perturbation size. One possible solution to this trade-off is using semantic perturbation, like colorization-based attacks. The multi-objective optimization strategy \cite{jia2022adv} is also a possible way to balance stealthiness and attack strength.

Moreover, because implementing physical experiments needs lots of expenditure, some researchers simulated the physical adversarial attack through a differential 3D renderer. However, current rendering techniques are susceptible to geometric and lighting transformations that distort the synthesized image. The gap between realistic photos and rendered images can affect the attack success rate or even make the adversarial attack ineffective. There is still a need to find better renderer or simulation methods or construct real-world datasets for evaluation. 

In addition, present physical AEs often attack the target deep learning model directly. However, off-the-shelf applications often contain complete pipelines, including data acquisition and preprocessing, which may unexpectedly influence the effect of AEs. For example, the image filter and the point cloud sampling process may influence the ASRs. There is still promotion space for present physical attacks to compromise these off-the-shelf applications.

\subsection{Designing efficient 3D black-box adversarial attacks}
Only a few works \cite{cao2019adversarial,sun2020towards,wicker2019robustness} explored the 3D black-box adversarial attacks, while most used very primitive or heuristic algorithms like random sampling, greedy search, and evolution algorithm. Many techniques like priors-transferring and boundary estimation have been exploited for 2D black-box attacks. Migrating these skills to 3D data may foster more efficient algorithms with fewer queries. For example, recently, 3DHacker \cite{tao20233dhacker} used boundary estimation to improve the query efficiency of hard-label attacks.

\subsection{Evaluating the robustness of newly proposed models} Novel 2D and 3D computer vision models' robustness, e.g., Transformer \cite{vaswani2017attention} and Point Transformer \cite{zhao2021point} , still need to be studies. Additionally, likelihood-based generative models have emerged, like the diffusion model \cite{ho2020denoising}, which can produce high-quality images. Some pilot studies have already evaluated their robustness \cite{zhuang2023pilot}. But these works are very rudimentary, and there is a lot of room for improvement. Moreover, the diffusion model is also helpful to improve the performance of generative-model-based attacks, such as unrestricted semantic attacks. However, how to backward its gradients to improve attack efficiency and how to disentangle its latent noises to control image attributes better still needs further research.

\subsection{Evaluating the safety of novel CV applications}
An increasing number of novel computer vision applications have incorporated deep learning, such as medical image processing, rain and fog removal, and pedestrian detection. The security requirements and potential attack surfaces of these emerging tasks vary a lot. Consequently, designing adversarial attacks for these novel tasks necessitates special considerations. For example, Schmalfuss et al. proposed an adversarial weather attack for motion estimation \cite{schmalfuss2023distracting} that simulates weather effects by utilizing adversarially optimized particles.

\subsection{Breaking newly proposed defenses}
Many adversarial attacks are proposed specifically targeting certain defense mechanisms, such as the C\&W attack to break the defensive distillation, the BPDA attack to break gradient shattering, and the BlindSpot attack to break the adversarial training. With the development of this race of adversarial machine learning, an increasing number of novel defenses are proposed, such as provable defenses and adversarial purification \cite{nie2022diffusion}. It is still unknown whether these defense mechanisms can be defeated by stronger adversarial attacks.
\section{Conclusion}

In this survey, we comprehensively review the recent progress of adversarial attacks that damage the robustness and safety of deep learning models in computer vision tasks. In contrast to previous works, we summarize the adversarial attacks for 3D computer vision for the first time. Moreover, we extend the connotation of adversarial examples to imperceptible and semantic perturbations. For semantic perturbations, we systematically summarize the latest methods, including local and global spatial transformation, color space distortion, and attribution modification. What's more, we investigate the physically realizable adversarial attacks towards safety-critical missions like self-driving vehicles and face recognition. In the end, we give some understanding of improving the transferability of AEs, making the AEs physically achievable, boosting the 3D black-box attack's efficiency, evaluating the robustness of emerging models, and designing adversarial attacks for novel applications and defenses.

\label{Sec_conclusion}



\bibliographystyle{ACM-Reference-Format}
\bibliography{reference}


\begin{thebibliography}{178}


\ifx \showCODEN    \undefined \def \showCODEN     #1{\unskip}     \fi
\ifx \showDOI      \undefined \def \showDOI       #1{#1}\fi
\ifx \showISBNx    \undefined \def \showISBNx     #1{\unskip}     \fi
\ifx \showISBNxiii \undefined \def \showISBNxiii  #1{\unskip}     \fi
\ifx \showISSN     \undefined \def \showISSN      #1{\unskip}     \fi
\ifx \showLCCN     \undefined \def \showLCCN      #1{\unskip}     \fi
\ifx \shownote     \undefined \def \shownote      #1{#1}          \fi
\ifx \showarticletitle \undefined \def \showarticletitle #1{#1}   \fi
\ifx \showURL      \undefined \def \showURL       {\relax}        \fi
\providecommand\bibfield[2]{#2}
\providecommand\bibinfo[2]{#2}
\providecommand\natexlab[1]{#1}
\providecommand\showeprint[2][]{arXiv:#2}

\bibitem[Agarwal et~al\mbox{.}(2020)]%
        {agarwal2020noise}
\bibfield{author}{\bibinfo{person}{Akshay Agarwal}, \bibinfo{person}{Mayank
  Vatsa}, \bibinfo{person}{Richa Singh}, {and} \bibinfo{person}{Nalini~K
  Ratha}.} \bibinfo{year}{2020}\natexlab{}.
\newblock \showarticletitle{Noise is inside me! generating adversarial
  perturbations with noise derived from natural filters}. In
  \bibinfo{booktitle}{\emph{Proceedings of the IEEE/CVF Conference on Computer
  Vision and Pattern Recognition Workshops}}. \bibinfo{pages}{774--775}.
\newblock


\bibitem[Al-Dujaili and O'Reilly(2019)]%
        {al2019there}
\bibfield{author}{\bibinfo{person}{Abdullah Al-Dujaili} {and}
  \bibinfo{person}{Una-May O'Reilly}.} \bibinfo{year}{2019}\natexlab{}.
\newblock \showarticletitle{There are no bit parts for sign bits in black-box
  attacks}.
\newblock \bibinfo{journal}{\emph{arXiv preprint arXiv:1902.06894}}
  (\bibinfo{year}{2019}).
\newblock


\bibitem[Alaifari et~al\mbox{.}(2018)]%
        {alaifari2018adef}
\bibfield{author}{\bibinfo{person}{Rima Alaifari}, \bibinfo{person}{Giovanni~S
  Alberti}, {and} \bibinfo{person}{Tandri Gauksson}.}
  \bibinfo{year}{2018}\natexlab{}.
\newblock \showarticletitle{ADef: an Iterative Algorithm to Construct
  Adversarial Deformations}. In \bibinfo{booktitle}{\emph{International
  Conference on Learning Representations}}.
\newblock


\bibitem[Andriushchenko et~al\mbox{.}(2020)]%
        {andriushchenko2020square}
\bibfield{author}{\bibinfo{person}{Maksym Andriushchenko},
  \bibinfo{person}{Francesco Croce}, \bibinfo{person}{Nicolas Flammarion},
  {and} \bibinfo{person}{Matthias Hein}.} \bibinfo{year}{2020}\natexlab{}.
\newblock \showarticletitle{Square attack: a query-efficient black-box
  adversarial attack via random search}. In \bibinfo{booktitle}{\emph{European
  Conference on Computer Vision}}. Springer, \bibinfo{pages}{484--501}.
\newblock


\bibitem[Athalye et~al\mbox{.}(2018a)]%
        {athalye2018obfuscated}
\bibfield{author}{\bibinfo{person}{Anish Athalye}, \bibinfo{person}{Nicholas
  Carlini}, {and} \bibinfo{person}{David Wagner}.}
  \bibinfo{year}{2018}\natexlab{a}.
\newblock \showarticletitle{Obfuscated gradients give a false sense of
  security: Circumventing defenses to adversarial examples}. In
  \bibinfo{booktitle}{\emph{International conference on machine learning}}.
  PMLR, \bibinfo{pages}{274--283}.
\newblock


\bibitem[Athalye et~al\mbox{.}(2018b)]%
        {athalye2018synthesizing}
\bibfield{author}{\bibinfo{person}{Anish Athalye}, \bibinfo{person}{Logan
  Engstrom}, \bibinfo{person}{Andrew Ilyas}, {and} \bibinfo{person}{Kevin
  Kwok}.} \bibinfo{year}{2018}\natexlab{b}.
\newblock \showarticletitle{Synthesizing robust adversarial examples}. In
  \bibinfo{booktitle}{\emph{International conference on machine learning}}.
  PMLR, \bibinfo{pages}{284--293}.
\newblock


\bibitem[Aydin et~al\mbox{.}(2021)]%
        {aydin2021imperceptible}
\bibfield{author}{\bibinfo{person}{Ayberk Aydin}, \bibinfo{person}{Deniz Sen},
  \bibinfo{person}{Berat~Tuna Karli}, \bibinfo{person}{Oguz Hanoglu}, {and}
  \bibinfo{person}{Alptekin Temizel}.} \bibinfo{year}{2021}\natexlab{}.
\newblock \showarticletitle{Imperceptible Adversarial Examples by Spatial
  Chroma-Shift}. In \bibinfo{booktitle}{\emph{Proceedings of the 1st
  International Workshop on Adversarial Learning for Multimedia}}.
  \bibinfo{pages}{8--14}.
\newblock


\bibitem[Baluja and Fischer(2018)]%
        {baluja2018learning}
\bibfield{author}{\bibinfo{person}{Shumeet Baluja} {and} \bibinfo{person}{Ian
  Fischer}.} \bibinfo{year}{2018}\natexlab{}.
\newblock \showarticletitle{Learning to attack: Adversarial transformation
  networks}. In \bibinfo{booktitle}{\emph{Proceedings of the AAAI Conference on
  Artificial Intelligence}}, Vol.~\bibinfo{volume}{32}.
\newblock


\bibitem[Bhagoji et~al\mbox{.}(2018)]%
        {bhagoji2018practical}
\bibfield{author}{\bibinfo{person}{Arjun~Nitin Bhagoji},
  \bibinfo{person}{Warren He}, \bibinfo{person}{Bo Li}, {and}
  \bibinfo{person}{Dawn Song}.} \bibinfo{year}{2018}\natexlab{}.
\newblock \showarticletitle{Practical black-box attacks on deep neural networks
  using efficient query mechanisms}. In \bibinfo{booktitle}{\emph{Proceedings
  of the European Conference on Computer Vision (ECCV)}}.
  \bibinfo{pages}{154--169}.
\newblock


\bibitem[Bhattad et~al\mbox{.}(2019)]%
        {bhattad2019unrestricted}
\bibfield{author}{\bibinfo{person}{Anand Bhattad}, \bibinfo{person}{Min~Jin
  Chong}, \bibinfo{person}{Kaizhao Liang}, \bibinfo{person}{Bo Li}, {and}
  \bibinfo{person}{DA Forsyth}.} \bibinfo{year}{2019}\natexlab{}.
\newblock \showarticletitle{Unrestricted Adversarial Examples via Semantic
  Manipulation}. In \bibinfo{booktitle}{\emph{International Conference on
  Learning Representations}}.
\newblock


\bibitem[Brendel et~al\mbox{.}(2017)]%
        {brendel2017decision}
\bibfield{author}{\bibinfo{person}{Wieland Brendel}, \bibinfo{person}{Jonas
  Rauber}, {and} \bibinfo{person}{Matthias Bethge}.}
  \bibinfo{year}{2017}\natexlab{}.
\newblock \showarticletitle{Decision-based adversarial attacks: Reliable
  attacks against black-box machine learning models}.
\newblock \bibinfo{journal}{\emph{arXiv preprint arXiv:1712.04248}}
  (\bibinfo{year}{2017}).
\newblock


\bibitem[Brown et~al\mbox{.}(2017)]%
        {brown2017adversarial}
\bibfield{author}{\bibinfo{person}{Tom~B Brown}, \bibinfo{person}{Dandelion
  Man{\'e}}, \bibinfo{person}{Aurko Roy}, \bibinfo{person}{Mart{\'\i}n Abadi},
  {and} \bibinfo{person}{Justin Gilmer}.} \bibinfo{year}{2017}\natexlab{}.
\newblock \showarticletitle{Adversarial patch}.
\newblock \bibinfo{journal}{\emph{arXiv preprint arXiv:1712.09665}}
  (\bibinfo{year}{2017}).
\newblock


\bibitem[Brunner et~al\mbox{.}(2019)]%
        {brunner2019guessing}
\bibfield{author}{\bibinfo{person}{Thomas Brunner}, \bibinfo{person}{Frederik
  Diehl}, \bibinfo{person}{Michael~Truong Le}, {and} \bibinfo{person}{Alois
  Knoll}.} \bibinfo{year}{2019}\natexlab{}.
\newblock \showarticletitle{Guessing smart: Biased sampling for efficient
  black-box adversarial attacks}. In \bibinfo{booktitle}{\emph{Proceedings of
  the IEEE/CVF International Conference on Computer Vision}}.
  \bibinfo{pages}{4958--4966}.
\newblock


\bibitem[Byun et~al\mbox{.}(2022)]%
        {byun2022improving}
\bibfield{author}{\bibinfo{person}{Junyoung Byun}, \bibinfo{person}{Seungju
  Cho}, \bibinfo{person}{Myung-Joon Kwon}, \bibinfo{person}{Hee-Seon Kim},
  {and} \bibinfo{person}{Changick Kim}.} \bibinfo{year}{2022}\natexlab{}.
\newblock \showarticletitle{Improving the Transferability of Targeted
  Adversarial Examples through Object-Based Diverse Input}. In
  \bibinfo{booktitle}{\emph{Proceedings of the IEEE/CVF Conference on Computer
  Vision and Pattern Recognition}}. \bibinfo{pages}{15244--15253}.
\newblock


\bibitem[Cao et~al\mbox{.}(2019a)]%
        {cao2019adversarial2}
\bibfield{author}{\bibinfo{person}{Yulong Cao}, \bibinfo{person}{Chaowei Xiao},
  \bibinfo{person}{Benjamin Cyr}, \bibinfo{person}{Yimeng Zhou},
  \bibinfo{person}{Won Park}, \bibinfo{person}{Sara Rampazzi},
  \bibinfo{person}{Qi~Alfred Chen}, \bibinfo{person}{Kevin Fu}, {and}
  \bibinfo{person}{Z~Morley Mao}.} \bibinfo{year}{2019}\natexlab{a}.
\newblock \showarticletitle{Adversarial sensor attack on lidar-based perception
  in autonomous driving}. In \bibinfo{booktitle}{\emph{Proceedings of the 2019
  ACM SIGSAC conference on computer and communications security}}.
  \bibinfo{pages}{2267--2281}.
\newblock


\bibitem[Cao et~al\mbox{.}(2019b)]%
        {cao2019adversarial}
\bibfield{author}{\bibinfo{person}{Yulong Cao}, \bibinfo{person}{Chaowei Xiao},
  \bibinfo{person}{Dawei Yang}, \bibinfo{person}{Jing Fang},
  \bibinfo{person}{Ruigang Yang}, \bibinfo{person}{Mingyan Liu}, {and}
  \bibinfo{person}{Bo Li}.} \bibinfo{year}{2019}\natexlab{b}.
\newblock \showarticletitle{Adversarial objects against lidar-based autonomous
  driving systems}.
\newblock \bibinfo{journal}{\emph{arXiv preprint arXiv:1907.05418}}
  (\bibinfo{year}{2019}).
\newblock


\bibitem[Carlini and Wagner(2017)]%
        {carlini2017towards}
\bibfield{author}{\bibinfo{person}{Nicholas Carlini} {and}
  \bibinfo{person}{David Wagner}.} \bibinfo{year}{2017}\natexlab{}.
\newblock \showarticletitle{Towards evaluating the robustness of neural
  networks}. In \bibinfo{booktitle}{\emph{2017 ieee symposium on security and
  privacy (sp)}}. Ieee, \bibinfo{pages}{39--57}.
\newblock


\bibitem[Chen and Huang(2021)]%
        {chen2021camdar}
\bibfield{author}{\bibinfo{person}{Chang Chen} {and} \bibinfo{person}{Teng
  Huang}.} \bibinfo{year}{2021}\natexlab{}.
\newblock \showarticletitle{Camdar-adv: generating adversarial patches on 3D
  object}.
\newblock \bibinfo{journal}{\emph{International Journal of Intelligent
  Systems}} \bibinfo{volume}{36}, \bibinfo{number}{3} (\bibinfo{year}{2021}),
  \bibinfo{pages}{1441--1453}.
\newblock


\bibitem[Chen and Gu(2020)]%
        {chen2020rays}
\bibfield{author}{\bibinfo{person}{Jinghui Chen} {and}
  \bibinfo{person}{Quanquan Gu}.} \bibinfo{year}{2020}\natexlab{}.
\newblock \showarticletitle{Rays: A ray searching method for hard-label
  adversarial attack}. In \bibinfo{booktitle}{\emph{Proceedings of the 26th ACM
  SIGKDD International Conference on Knowledge Discovery \& Data Mining}}.
  \bibinfo{pages}{1739--1747}.
\newblock


\bibitem[Chen et~al\mbox{.}(2020)]%
        {chen2020hopskipjumpattack}
\bibfield{author}{\bibinfo{person}{Jianbo Chen}, \bibinfo{person}{Michael~I
  Jordan}, {and} \bibinfo{person}{Martin~J Wainwright}.}
  \bibinfo{year}{2020}\natexlab{}.
\newblock \showarticletitle{Hopskipjumpattack: A query-efficient decision-based
  attack}. In \bibinfo{booktitle}{\emph{2020 ieee symposium on security and
  privacy (sp)}}. IEEE, \bibinfo{pages}{1277--1294}.
\newblock


\bibitem[Chen et~al\mbox{.}(2018b)]%
        {chen2018ead}
\bibfield{author}{\bibinfo{person}{Pin-Yu Chen}, \bibinfo{person}{Yash Sharma},
  \bibinfo{person}{Huan Zhang}, \bibinfo{person}{Jinfeng Yi}, {and}
  \bibinfo{person}{Cho-Jui Hsieh}.} \bibinfo{year}{2018}\natexlab{b}.
\newblock \showarticletitle{Ead: elastic-net attacks to deep neural networks
  via adversarial examples}. In \bibinfo{booktitle}{\emph{Proceedings of the
  AAAI conference on artificial intelligence}}, Vol.~\bibinfo{volume}{32}.
\newblock


\bibitem[Chen et~al\mbox{.}(2017)]%
        {chen2017zoo}
\bibfield{author}{\bibinfo{person}{Pin-Yu Chen}, \bibinfo{person}{Huan Zhang},
  \bibinfo{person}{Yash Sharma}, \bibinfo{person}{Jinfeng Yi}, {and}
  \bibinfo{person}{Cho-Jui Hsieh}.} \bibinfo{year}{2017}\natexlab{}.
\newblock \showarticletitle{Zoo: Zeroth order optimization based black-box
  attacks to deep neural networks without training substitute models}. In
  \bibinfo{booktitle}{\emph{Proceedings of the 10th ACM workshop on artificial
  intelligence and security}}. \bibinfo{pages}{15--26}.
\newblock


\bibitem[Chen et~al\mbox{.}(2018a)]%
        {chen2018shapeshifter}
\bibfield{author}{\bibinfo{person}{Shang-Tse Chen}, \bibinfo{person}{Cory
  Cornelius}, \bibinfo{person}{Jason Martin}, {and} \bibinfo{person}{Duen
  Horng~Polo Chau}.} \bibinfo{year}{2018}\natexlab{a}.
\newblock \showarticletitle{Shapeshifter: Robust physical adversarial attack on
  faster r-cnn object detector}. In \bibinfo{booktitle}{\emph{Joint European
  Conference on Machine Learning and Knowledge Discovery in Databases}}.
  Springer, \bibinfo{pages}{52--68}.
\newblock


\bibitem[Chen et~al\mbox{.}(2023)]%
        {chen2023content}
\bibfield{author}{\bibinfo{person}{Zhaoyu Chen}, \bibinfo{person}{Bo Li},
  \bibinfo{person}{Shuang Wu}, \bibinfo{person}{Kaixun Jiang},
  \bibinfo{person}{Shouhong Ding}, {and} \bibinfo{person}{Wenqiang Zhang}.}
  \bibinfo{year}{2023}\natexlab{}.
\newblock \showarticletitle{Content-based Unrestricted Adversarial Attack}.
\newblock \bibinfo{journal}{\emph{arXiv preprint arXiv:2305.10665}}
  (\bibinfo{year}{2023}).
\newblock


\bibitem[Cheng et~al\mbox{.}(2020)]%
        {cheng2020sign}
\bibfield{author}{\bibinfo{person}{Minhao Cheng}, \bibinfo{person}{Simranjit
  Singh}, \bibinfo{person}{Patrick~H Chen}, \bibinfo{person}{Pin-Yu Chen},
  \bibinfo{person}{Sijia Liu}, {and} \bibinfo{person}{Cho-Jui Hsieh}.}
  \bibinfo{year}{2020}\natexlab{}.
\newblock \showarticletitle{Sign-OPT: A Query-Efficient Hard-label Adversarial
  Attack}. In \bibinfo{booktitle}{\emph{International Conference on Learning
  Representations}}.
\newblock


\bibitem[Cheng et~al\mbox{.}(2019b)]%
        {cheng2019query}
\bibfield{author}{\bibinfo{person}{Minhao Cheng}, \bibinfo{person}{Huan Zhang},
  \bibinfo{person}{Cho-Jui Hsieh}, \bibinfo{person}{Thong Le},
  \bibinfo{person}{Pin-Yu Chen}, {and} \bibinfo{person}{Jinfeng Yi}.}
  \bibinfo{year}{2019}\natexlab{b}.
\newblock \showarticletitle{Query-efficient hard-label black-box attack: An
  optimization-based approach}. In \bibinfo{booktitle}{\emph{International
  Conference on Learning Representations}}. International Conference on
  Learning Representations, ICLR.
\newblock


\bibitem[Cheng et~al\mbox{.}(2019a)]%
        {cheng2019improving}
\bibfield{author}{\bibinfo{person}{Shuyu Cheng}, \bibinfo{person}{Yinpeng
  Dong}, \bibinfo{person}{Tianyu Pang}, \bibinfo{person}{Hang Su}, {and}
  \bibinfo{person}{Jun Zhu}.} \bibinfo{year}{2019}\natexlab{a}.
\newblock \showarticletitle{Improving black-box adversarial attacks with a
  transfer-based prior}.
\newblock \bibinfo{journal}{\emph{Advances in neural information processing
  systems}}  \bibinfo{volume}{32} (\bibinfo{year}{2019}).
\newblock


\bibitem[Croce et~al\mbox{.}(2022)]%
        {croce2022sparse}
\bibfield{author}{\bibinfo{person}{Francesco Croce}, \bibinfo{person}{Maksym
  Andriushchenko}, \bibinfo{person}{Naman~D Singh}, \bibinfo{person}{Nicolas
  Flammarion}, {and} \bibinfo{person}{Matthias Hein}.}
  \bibinfo{year}{2022}\natexlab{}.
\newblock \showarticletitle{Sparse-rs: a versatile framework for
  query-efficient sparse black-box adversarial attacks}. In
  \bibinfo{booktitle}{\emph{Proceedings of the AAAI Conference on Artificial
  Intelligence}}, Vol.~\bibinfo{volume}{36}. \bibinfo{pages}{6437--6445}.
\newblock


\bibitem[Croce and Hein(2019)]%
        {croce2019sparse}
\bibfield{author}{\bibinfo{person}{Francesco Croce} {and}
  \bibinfo{person}{Matthias Hein}.} \bibinfo{year}{2019}\natexlab{}.
\newblock \showarticletitle{Sparse and Imperceivable Adversarial Attacks}. In
  \bibinfo{booktitle}{\emph{2019 IEEE/CVF International Conference on Computer
  Vision (ICCV)}}. \bibinfo{pages}{4723--4731}.
\newblock
\urldef\tempurl%
\url{https://doi.org/10.1109/ICCV.2019.00482}
\showDOI{\tempurl}


\bibitem[Dong et~al\mbox{.}(2020)]%
        {dong2020greedyfool}
\bibfield{author}{\bibinfo{person}{Xiaoyi Dong}, \bibinfo{person}{Dongdong
  Chen}, \bibinfo{person}{Jianmin Bao}, \bibinfo{person}{Chuan Qin},
  \bibinfo{person}{Lu Yuan}, \bibinfo{person}{Weiming Zhang},
  \bibinfo{person}{Nenghai Yu}, {and} \bibinfo{person}{Dong Chen}.}
  \bibinfo{year}{2020}\natexlab{}.
\newblock \showarticletitle{GreedyFool: Distortion-aware sparse adversarial
  attack}.
\newblock \bibinfo{journal}{\emph{Advances in Neural Information Processing
  Systems}}  \bibinfo{volume}{33} (\bibinfo{year}{2020}),
  \bibinfo{pages}{11226--11236}.
\newblock


\bibitem[Dong et~al\mbox{.}(2018)]%
        {dong2018boosting}
\bibfield{author}{\bibinfo{person}{Yinpeng Dong}, \bibinfo{person}{Fangzhou
  Liao}, \bibinfo{person}{Tianyu Pang}, \bibinfo{person}{Hang Su},
  \bibinfo{person}{Jun Zhu}, \bibinfo{person}{Xiaolin Hu}, {and}
  \bibinfo{person}{Jianguo Li}.} \bibinfo{year}{2018}\natexlab{}.
\newblock \showarticletitle{Boosting adversarial attacks with momentum}. In
  \bibinfo{booktitle}{\emph{Proceedings of the IEEE conference on computer
  vision and pattern recognition}}. \bibinfo{pages}{9185--9193}.
\newblock


\bibitem[Dong et~al\mbox{.}(2019)]%
        {dong2019evading}
\bibfield{author}{\bibinfo{person}{Yinpeng Dong}, \bibinfo{person}{Tianyu
  Pang}, \bibinfo{person}{Hang Su}, {and} \bibinfo{person}{Jun Zhu}.}
  \bibinfo{year}{2019}\natexlab{}.
\newblock \showarticletitle{Evading defenses to transferable adversarial
  examples by translation-invariant attacks}. In
  \bibinfo{booktitle}{\emph{Proceedings of the IEEE/CVF Conference on Computer
  Vision and Pattern Recognition}}. \bibinfo{pages}{4312--4321}.
\newblock


\bibitem[Duan et~al\mbox{.}(2020)]%
        {duan2020adversarial}
\bibfield{author}{\bibinfo{person}{Ranjie Duan}, \bibinfo{person}{Xingjun Ma},
  \bibinfo{person}{Yisen Wang}, \bibinfo{person}{James Bailey},
  \bibinfo{person}{A.~K. Qin}, {and} \bibinfo{person}{Yun Yang}.}
  \bibinfo{year}{2020}\natexlab{}.
\newblock \showarticletitle{Adversarial Camouflage: Hiding Physical-World
  Attacks With Natural Styles}. In \bibinfo{booktitle}{\emph{Proceedings of the
  IEEE/CVF Conference on Computer Vision and Pattern Recognition (CVPR)}}.
\newblock


\bibitem[Engstrom et~al\mbox{.}(2018)]%
        {engstrom2018rotation}
\bibfield{author}{\bibinfo{person}{Logan Engstrom}, \bibinfo{person}{Brandon
  Tran}, \bibinfo{person}{Dimitris Tsipras}, \bibinfo{person}{Ludwig Schmidt},
  {and} \bibinfo{person}{Aleksander Madry}.} \bibinfo{year}{2018}\natexlab{}.
\newblock \showarticletitle{A rotation and a translation suffice: Fooling cnns
  with simple transformations}.
\newblock  (\bibinfo{year}{2018}).
\newblock


\bibitem[Eykholt et~al\mbox{.}(2018)]%
        {eykholt2018robust}
\bibfield{author}{\bibinfo{person}{Kevin Eykholt}, \bibinfo{person}{Ivan
  Evtimov}, \bibinfo{person}{Earlence Fernandes}, \bibinfo{person}{Bo Li},
  \bibinfo{person}{Amir Rahmati}, \bibinfo{person}{Chaowei Xiao},
  \bibinfo{person}{Atul Prakash}, \bibinfo{person}{Tadayoshi Kohno}, {and}
  \bibinfo{person}{Dawn Song}.} \bibinfo{year}{2018}\natexlab{}.
\newblock \showarticletitle{Robust physical-world attacks on deep learning
  visual classification}. In \bibinfo{booktitle}{\emph{Proceedings of the IEEE
  conference on computer vision and pattern recognition}}.
  \bibinfo{pages}{1625--1634}.
\newblock


\bibitem[Fawzi and Frossard(2015)]%
        {fawzi2015manitest}
\bibfield{author}{\bibinfo{person}{Alhussein Fawzi} {and}
  \bibinfo{person}{Pascal Frossard}.} \bibinfo{year}{2015}\natexlab{}.
\newblock \showarticletitle{Manitest: Are classifiers really invariant?}. In
  \bibinfo{booktitle}{\emph{British Machine Vision Conference (BMVC)}}.
\newblock


\bibitem[Ganeshan et~al\mbox{.}(2019)]%
        {ganeshan2019fda}
\bibfield{author}{\bibinfo{person}{Aditya Ganeshan}, \bibinfo{person}{Vivek
  BS}, {and} \bibinfo{person}{R~Venkatesh Babu}.}
  \bibinfo{year}{2019}\natexlab{}.
\newblock \showarticletitle{Fda: Feature disruptive attack}. In
  \bibinfo{booktitle}{\emph{Proceedings of the IEEE/CVF International
  Conference on Computer Vision}}. \bibinfo{pages}{8069--8079}.
\newblock


\bibitem[Geirhos et~al\mbox{.}(2018)]%
        {geirhos2018imagenet}
\bibfield{author}{\bibinfo{person}{Robert Geirhos}, \bibinfo{person}{Patricia
  Rubisch}, \bibinfo{person}{Claudio Michaelis}, \bibinfo{person}{Matthias
  Bethge}, \bibinfo{person}{Felix~A Wichmann}, {and} \bibinfo{person}{Wieland
  Brendel}.} \bibinfo{year}{2018}\natexlab{}.
\newblock \showarticletitle{ImageNet-trained CNNs are biased towards texture;
  increasing shape bias improves accuracy and robustness}.
\newblock \bibinfo{journal}{\emph{arXiv preprint arXiv:1811.12231}}
  (\bibinfo{year}{2018}).
\newblock


\bibitem[Gilmer et~al\mbox{.}(2018)]%
        {gilmer2018motivating}
\bibfield{author}{\bibinfo{person}{Justin Gilmer}, \bibinfo{person}{Ryan~P
  Adams}, \bibinfo{person}{Ian Goodfellow}, \bibinfo{person}{David Andersen},
  {and} \bibinfo{person}{George~E Dahl}.} \bibinfo{year}{2018}\natexlab{}.
\newblock \showarticletitle{Motivating the rules of the game for adversarial
  example research}.
\newblock \bibinfo{journal}{\emph{arXiv preprint arXiv:1807.06732}}
  (\bibinfo{year}{2018}).
\newblock


\bibitem[Gnanasambandam et~al\mbox{.}(2021)]%
        {gnanasambandam2021optical}
\bibfield{author}{\bibinfo{person}{Abhiram Gnanasambandam},
  \bibinfo{person}{Alex~M Sherman}, {and} \bibinfo{person}{Stanley~H Chan}.}
  \bibinfo{year}{2021}\natexlab{}.
\newblock \showarticletitle{Optical adversarial attack}. In
  \bibinfo{booktitle}{\emph{Proceedings of the IEEE/CVF International
  Conference on Computer Vision}}. \bibinfo{pages}{92--101}.
\newblock


\bibitem[Goodfellow et~al\mbox{.}(2014)]%
        {goodfellow2014explaining}
\bibfield{author}{\bibinfo{person}{Ian~J Goodfellow}, \bibinfo{person}{Jonathon
  Shlens}, {and} \bibinfo{person}{Christian Szegedy}.}
  \bibinfo{year}{2014}\natexlab{}.
\newblock \showarticletitle{Explaining and harnessing adversarial examples}.
\newblock \bibinfo{journal}{\emph{arXiv preprint arXiv:1412.6572}}
  (\bibinfo{year}{2014}).
\newblock


\bibitem[Guo et~al\mbox{.}(2019a)]%
        {guo2019simple}
\bibfield{author}{\bibinfo{person}{Chuan Guo}, \bibinfo{person}{Jacob Gardner},
  \bibinfo{person}{Yurong You}, \bibinfo{person}{Andrew~Gordon Wilson}, {and}
  \bibinfo{person}{Kilian Weinberger}.} \bibinfo{year}{2019}\natexlab{a}.
\newblock \showarticletitle{Simple black-box adversarial attacks}. In
  \bibinfo{booktitle}{\emph{International Conference on Machine Learning}}.
  PMLR, \bibinfo{pages}{2484--2493}.
\newblock


\bibitem[Guo et~al\mbox{.}(2021)]%
        {guo2021pct}
\bibfield{author}{\bibinfo{person}{Meng-Hao Guo}, \bibinfo{person}{Jun-Xiong
  Cai}, \bibinfo{person}{Zheng-Ning Liu}, \bibinfo{person}{Tai-Jiang Mu},
  \bibinfo{person}{Ralph~R Martin}, {and} \bibinfo{person}{Shi-Min Hu}.}
  \bibinfo{year}{2021}\natexlab{}.
\newblock \showarticletitle{Pct: Point cloud transformer}.
\newblock \bibinfo{journal}{\emph{Computational Visual Media}}
  \bibinfo{volume}{7}, \bibinfo{number}{2} (\bibinfo{year}{2021}),
  \bibinfo{pages}{187--199}.
\newblock


\bibitem[Guo et~al\mbox{.}(2019b)]%
        {guo2019subspace}
\bibfield{author}{\bibinfo{person}{Yiwen Guo}, \bibinfo{person}{Ziang Yan},
  {and} \bibinfo{person}{Changshui Zhang}.} \bibinfo{year}{2019}\natexlab{b}.
\newblock \showarticletitle{Subspace attack: Exploiting promising subspaces for
  query-efficient black-box attacks}.
\newblock \bibinfo{journal}{\emph{Advances in Neural Information Processing
  Systems}}  \bibinfo{volume}{32} (\bibinfo{year}{2019}).
\newblock


\bibitem[Hamdi et~al\mbox{.}(2020)]%
        {hamdi2020advpc}
\bibfield{author}{\bibinfo{person}{Abdullah Hamdi}, \bibinfo{person}{Sara
  Rojas}, \bibinfo{person}{Ali Thabet}, {and} \bibinfo{person}{Bernard
  Ghanem}.} \bibinfo{year}{2020}\natexlab{}.
\newblock \showarticletitle{Advpc: Transferable adversarial perturbations on 3d
  point clouds}. In \bibinfo{booktitle}{\emph{European Conference on Computer
  Vision}}. Springer, \bibinfo{pages}{241--257}.
\newblock


\bibitem[He et~al\mbox{.}(2018)]%
        {he2018decision}
\bibfield{author}{\bibinfo{person}{Warren He}, \bibinfo{person}{Bo Li}, {and}
  \bibinfo{person}{Dawn Song}.} \bibinfo{year}{2018}\natexlab{}.
\newblock \showarticletitle{Decision boundary analysis of adversarial
  examples}. In \bibinfo{booktitle}{\emph{International Conference on Learning
  Representations}}.
\newblock


\bibitem[He et~al\mbox{.}(2022a)]%
        {He2022TowardsSecurity}
\bibfield{author}{\bibinfo{person}{Yingzhe He}, \bibinfo{person}{Guozhu Meng},
  \bibinfo{person}{Kai Chen}, \bibinfo{person}{Xingbo Hu}, {and}
  \bibinfo{person}{Jinwen He}.} \bibinfo{year}{2022}\natexlab{a}.
\newblock \showarticletitle{Towards Security Threats of Deep Learning Systems:
  A Survey}.
\newblock \bibinfo{journal}{\emph{IEEE Transactions on Software Engineering}}
  \bibinfo{volume}{48}, \bibinfo{number}{5} (\bibinfo{year}{2022}),
  \bibinfo{pages}{1743--1770}.
\newblock


\bibitem[He et~al\mbox{.}(2022b)]%
        {he2022transferable}
\bibfield{author}{\bibinfo{person}{Ziwen He}, \bibinfo{person}{Wei Wang},
  \bibinfo{person}{Jing Dong}, {and} \bibinfo{person}{Tieniu Tan}.}
  \bibinfo{year}{2022}\natexlab{b}.
\newblock \showarticletitle{Transferable Sparse Adversarial Attack}. In
  \bibinfo{booktitle}{\emph{Proceedings of the IEEE/CVF Conference on Computer
  Vision and Pattern Recognition}}. \bibinfo{pages}{14963--14972}.
\newblock


\bibitem[Ho et~al\mbox{.}(2020)]%
        {ho2020denoising}
\bibfield{author}{\bibinfo{person}{Jonathan Ho}, \bibinfo{person}{Ajay Jain},
  {and} \bibinfo{person}{Pieter Abbeel}.} \bibinfo{year}{2020}\natexlab{}.
\newblock \showarticletitle{Denoising diffusion probabilistic models}.
\newblock \bibinfo{journal}{\emph{Advances in neural information processing
  systems}}  \bibinfo{volume}{33} (\bibinfo{year}{2020}),
  \bibinfo{pages}{6840--6851}.
\newblock


\bibitem[Hosseini and Poovendran(2018)]%
        {hosseini2018semantic}
\bibfield{author}{\bibinfo{person}{Hossein Hosseini} {and}
  \bibinfo{person}{Radha Poovendran}.} \bibinfo{year}{2018}\natexlab{}.
\newblock \showarticletitle{Semantic adversarial examples}. In
  \bibinfo{booktitle}{\emph{Proceedings of the IEEE Conference on Computer
  Vision and Pattern Recognition Workshops}}. \bibinfo{pages}{1614--1619}.
\newblock


\bibitem[Hu et~al\mbox{.}(2021)]%
        {hu2021naturalistic}
\bibfield{author}{\bibinfo{person}{Yu-Chih-Tuan Hu}, \bibinfo{person}{Bo-Han
  Kung}, \bibinfo{person}{Daniel~Stanley Tan}, \bibinfo{person}{Jun-Cheng
  Chen}, \bibinfo{person}{Kai-Lung Hua}, {and} \bibinfo{person}{Wen-Huang
  Cheng}.} \bibinfo{year}{2021}\natexlab{}.
\newblock \showarticletitle{Naturalistic physical adversarial patch for object
  detectors}. In \bibinfo{booktitle}{\emph{Proceedings of the IEEE/CVF
  International Conference on Computer Vision}}. \bibinfo{pages}{7848--7857}.
\newblock


\bibitem[Hua et~al\mbox{.}(2018)]%
        {ha2018pointwise}
\bibfield{author}{\bibinfo{person}{Binh-Son Hua}, \bibinfo{person}{Minh-Khoi
  Tran}, {and} \bibinfo{person}{Sai-Kit Yeung}.}
  \bibinfo{year}{2018}\natexlab{}.
\newblock \showarticletitle{Pointwise convolutional neural networks}. In
  \bibinfo{booktitle}{\emph{Proceedings of the IEEE conference on computer
  vision and pattern recognition}}. \bibinfo{pages}{984--993}.
\newblock


\bibitem[Huang et~al\mbox{.}(2022)]%
        {huang2022shape}
\bibfield{author}{\bibinfo{person}{Qidong Huang}, \bibinfo{person}{Xiaoyi
  Dong}, \bibinfo{person}{Dongdong Chen}, \bibinfo{person}{Hang Zhou},
  \bibinfo{person}{Weiming Zhang}, {and} \bibinfo{person}{Nenghai Yu}.}
  \bibinfo{year}{2022}\natexlab{}.
\newblock \showarticletitle{Shape-invariant 3D Adversarial Point Clouds}. In
  \bibinfo{booktitle}{\emph{Proceedings of the IEEE/CVF Conference on Computer
  Vision and Pattern Recognition}}. \bibinfo{pages}{15335--15344}.
\newblock


\bibitem[Huang et~al\mbox{.}(2019)]%
        {huang2019enhancing}
\bibfield{author}{\bibinfo{person}{Qian Huang}, \bibinfo{person}{Isay Katsman},
  \bibinfo{person}{Horace He}, \bibinfo{person}{Zeqi Gu},
  \bibinfo{person}{Serge Belongie}, {and} \bibinfo{person}{Ser-Nam Lim}.}
  \bibinfo{year}{2019}\natexlab{}.
\newblock \showarticletitle{Enhancing adversarial example transferability with
  an intermediate level attack}. In \bibinfo{booktitle}{\emph{Proceedings of
  the IEEE/CVF international conference on computer vision}}.
  \bibinfo{pages}{4733--4742}.
\newblock


\bibitem[Ilyas et~al\mbox{.}(2018b)]%
        {ilyas2018black}
\bibfield{author}{\bibinfo{person}{Andrew Ilyas}, \bibinfo{person}{Logan
  Engstrom}, \bibinfo{person}{Anish Athalye}, {and} \bibinfo{person}{Jessy
  Lin}.} \bibinfo{year}{2018}\natexlab{b}.
\newblock \showarticletitle{Black-box adversarial attacks with limited queries
  and information}. In \bibinfo{booktitle}{\emph{International Conference on
  Machine Learning}}. PMLR, \bibinfo{pages}{2137--2146}.
\newblock


\bibitem[Ilyas et~al\mbox{.}(2018a)]%
        {ilyas2018prior}
\bibfield{author}{\bibinfo{person}{Andrew Ilyas}, \bibinfo{person}{Logan
  Engstrom}, {and} \bibinfo{person}{Aleksander Madry}.}
  \bibinfo{year}{2018}\natexlab{a}.
\newblock \showarticletitle{Prior Convictions: Black-box Adversarial Attacks
  with Bandits and Priors}. In \bibinfo{booktitle}{\emph{International
  Conference on Learning Representations}}.
\newblock


\bibitem[Ilyas et~al\mbox{.}(2019)]%
        {ilyas2019adversarial}
\bibfield{author}{\bibinfo{person}{Andrew Ilyas}, \bibinfo{person}{Shibani
  Santurkar}, \bibinfo{person}{Dimitris Tsipras}, \bibinfo{person}{Logan
  Engstrom}, \bibinfo{person}{Brandon Tran}, {and} \bibinfo{person}{Aleksander
  Madry}.} \bibinfo{year}{2019}\natexlab{}.
\newblock \showarticletitle{Adversarial examples are not bugs, they are
  features}.
\newblock \bibinfo{journal}{\emph{Advances in neural information processing
  systems}}  \bibinfo{volume}{32} (\bibinfo{year}{2019}).
\newblock


\bibitem[Ji et~al\mbox{.}(2021)]%
        {ji2021poltergeist}
\bibfield{author}{\bibinfo{person}{Xiaoyu Ji}, \bibinfo{person}{Yushi Cheng},
  \bibinfo{person}{Yuepeng Zhang}, \bibinfo{person}{Kai Wang},
  \bibinfo{person}{Chen Yan}, \bibinfo{person}{Wenyuan Xu}, {and}
  \bibinfo{person}{Kevin Fu}.} \bibinfo{year}{2021}\natexlab{}.
\newblock \showarticletitle{Poltergeist: Acoustic adversarial machine learning
  against cameras and computer vision}. In \bibinfo{booktitle}{\emph{2021 IEEE
  Symposium on Security and Privacy (SP)}}. IEEE, \bibinfo{pages}{160--175}.
\newblock


\bibitem[Jia et~al\mbox{.}(2022)]%
        {jia2022adv}
\bibfield{author}{\bibinfo{person}{Shuai Jia}, \bibinfo{person}{Bangjie Yin},
  \bibinfo{person}{Taiping Yao}, \bibinfo{person}{Shouhong Ding},
  \bibinfo{person}{Chunhua Shen}, \bibinfo{person}{Xiaokang Yang}, {and}
  \bibinfo{person}{Chao Ma}.} \bibinfo{year}{2022}\natexlab{}.
\newblock \showarticletitle{Adv-attribute: Inconspicuous and transferable
  adversarial attack on face recognition}.
\newblock \bibinfo{journal}{\emph{Advances in Neural Information Processing
  Systems}}  \bibinfo{volume}{35} (\bibinfo{year}{2022}),
  \bibinfo{pages}{34136--34147}.
\newblock


\bibitem[Joshi et~al\mbox{.}(2019)]%
        {joshi2019semantic}
\bibfield{author}{\bibinfo{person}{Ameya Joshi}, \bibinfo{person}{Amitangshu
  Mukherjee}, \bibinfo{person}{Soumik Sarkar}, {and} \bibinfo{person}{Chinmay
  Hegde}.} \bibinfo{year}{2019}\natexlab{}.
\newblock \showarticletitle{Semantic adversarial attacks: Parametric
  transformations that fool deep classifiers}. In
  \bibinfo{booktitle}{\emph{Proceedings of the IEEE/CVF international
  conference on computer vision}}. \bibinfo{pages}{4773--4783}.
\newblock


\bibitem[Kanbak et~al\mbox{.}(2018)]%
        {kanbak2018geometric}
\bibfield{author}{\bibinfo{person}{Can Kanbak}, \bibinfo{person}{Seyed-Mohsen
  Moosavi-Dezfooli}, {and} \bibinfo{person}{Pascal Frossard}.}
  \bibinfo{year}{2018}\natexlab{}.
\newblock \showarticletitle{Geometric robustness of deep networks: analysis and
  improvement}. In \bibinfo{booktitle}{\emph{Proceedings of the IEEE Conference
  on Computer Vision and Pattern Recognition}}. \bibinfo{pages}{4441--4449}.
\newblock


\bibitem[Kato et~al\mbox{.}(2018)]%
        {kato2018neural}
\bibfield{author}{\bibinfo{person}{Hiroharu Kato}, \bibinfo{person}{Yoshitaka
  Ushiku}, {and} \bibinfo{person}{Tatsuya Harada}.}
  \bibinfo{year}{2018}\natexlab{}.
\newblock \showarticletitle{Neural 3d mesh renderer}. In
  \bibinfo{booktitle}{\emph{Proceedings of the IEEE conference on computer
  vision and pattern recognition}}. \bibinfo{pages}{3907--3916}.
\newblock


\bibitem[Khrulkov and Oseledets(2018)]%
        {khrulkov2018art}
\bibfield{author}{\bibinfo{person}{Valentin Khrulkov} {and}
  \bibinfo{person}{Ivan Oseledets}.} \bibinfo{year}{2018}\natexlab{}.
\newblock \showarticletitle{Art of singular vectors and universal adversarial
  perturbations}. In \bibinfo{booktitle}{\emph{Proceedings of the IEEE
  Conference on Computer Vision and Pattern Recognition}}.
  \bibinfo{pages}{8562--8570}.
\newblock


\bibitem[Kim et~al\mbox{.}(2021)]%
        {kim2021minimal}
\bibfield{author}{\bibinfo{person}{Jaeyeon Kim}, \bibinfo{person}{Binh-Son
  Hua}, \bibinfo{person}{Thanh Nguyen}, {and} \bibinfo{person}{Sai-Kit Yeung}.}
  \bibinfo{year}{2021}\natexlab{}.
\newblock \showarticletitle{Minimal adversarial examples for deep learning on
  3d point clouds}. In \bibinfo{booktitle}{\emph{Proceedings of the IEEE/CVF
  International Conference on Computer Vision}}. \bibinfo{pages}{7797--7806}.
\newblock


\bibitem[Kurakin et~al\mbox{.}(2018)]%
        {kurakin2018adversarial}
\bibfield{author}{\bibinfo{person}{Alexey Kurakin}, \bibinfo{person}{Ian~J
  Goodfellow}, {and} \bibinfo{person}{Samy Bengio}.}
  \bibinfo{year}{2018}\natexlab{}.
\newblock \showarticletitle{Adversarial examples in the physical world}.
\newblock In \bibinfo{booktitle}{\emph{Artificial intelligence safety and
  security}}. \bibinfo{publisher}{Chapman and Hall/CRC},
  \bibinfo{pages}{99--112}.
\newblock


\bibitem[Kurita et~al\mbox{.}(2020)]%
        {kurita-etal-2020-weight}
\bibfield{author}{\bibinfo{person}{Keita Kurita}, \bibinfo{person}{Paul
  Michel}, {and} \bibinfo{person}{Graham Neubig}.}
  \bibinfo{year}{2020}\natexlab{}.
\newblock \showarticletitle{Weight Poisoning Attacks on Pretrained Models}. In
  \bibinfo{booktitle}{\emph{Proceedings of the 58th Annual Meeting of the
  Association for Computational Linguistics}}. \bibinfo{publisher}{Association
  for Computational Linguistics}, \bibinfo{pages}{2793--2806}.
\newblock


\bibitem[Laidlaw and Feizi(2019)]%
        {laidlaw2019functional}
\bibfield{author}{\bibinfo{person}{Cassidy Laidlaw} {and}
  \bibinfo{person}{Soheil Feizi}.} \bibinfo{year}{2019}\natexlab{}.
\newblock \showarticletitle{Functional adversarial attacks}.
\newblock \bibinfo{journal}{\emph{Advances in neural information processing
  systems}}  \bibinfo{volume}{32} (\bibinfo{year}{2019}).
\newblock


\bibitem[Li et~al\mbox{.}(2020)]%
        {li2020towards}
\bibfield{author}{\bibinfo{person}{Maosen Li}, \bibinfo{person}{Cheng Deng},
  \bibinfo{person}{Tengjiao Li}, \bibinfo{person}{Junchi Yan},
  \bibinfo{person}{Xinbo Gao}, {and} \bibinfo{person}{Heng Huang}.}
  \bibinfo{year}{2020}\natexlab{}.
\newblock \showarticletitle{Towards transferable targeted attack}. In
  \bibinfo{booktitle}{\emph{Proceedings of the IEEE/CVF conference on computer
  vision and pattern recognition}}. \bibinfo{pages}{641--649}.
\newblock


\bibitem[Li et~al\mbox{.}(2021)]%
        {li2021qair}
\bibfield{author}{\bibinfo{person}{Xiaodan Li}, \bibinfo{person}{Jinfeng Li},
  \bibinfo{person}{Yuefeng Chen}, \bibinfo{person}{Shaokai Ye},
  \bibinfo{person}{Yuan He}, \bibinfo{person}{Shuhui Wang},
  \bibinfo{person}{Hang Su}, {and} \bibinfo{person}{Hui Xue}.}
  \bibinfo{year}{2021}\natexlab{}.
\newblock \showarticletitle{Qair: Practical query-efficient black-box attacks
  for image retrieval}. In \bibinfo{booktitle}{\emph{Proceedings of the
  IEEE/CVF Conference on Computer Vision and Pattern Recognition}}.
  \bibinfo{pages}{3330--3339}.
\newblock


\bibitem[Li et~al\mbox{.}(2019)]%
        {li2019nattack}
\bibfield{author}{\bibinfo{person}{Yandong Li}, \bibinfo{person}{Lijun Li},
  \bibinfo{person}{Liqiang Wang}, \bibinfo{person}{Tong Zhang}, {and}
  \bibinfo{person}{Boqing Gong}.} \bibinfo{year}{2019}\natexlab{}.
\newblock \showarticletitle{Nattack: Learning the distributions of adversarial
  examples for an improved black-box attack on deep neural networks}. In
  \bibinfo{booktitle}{\emph{International Conference on Machine Learning}}.
  PMLR, \bibinfo{pages}{3866--3876}.
\newblock


\bibitem[Li et~al\mbox{.}(2023)]%
        {Li_2023_physicalworld}
\bibfield{author}{\bibinfo{person}{Yanjie Li}, \bibinfo{person}{Yiquan Li},
  \bibinfo{person}{Xuelong Dai}, \bibinfo{person}{Songtao Guo}, {and}
  \bibinfo{person}{Bin Xiao}.} \bibinfo{year}{2023}\natexlab{}.
\newblock \showarticletitle{Physical-World Optical Adversarial Attacks on 3D
  Face Recognition}. In \bibinfo{booktitle}{\emph{Proceedings of the IEEE/CVF
  Conference on Computer Vision and Pattern Recognition (CVPR)}}.
  \bibinfo{pages}{24699--24708}.
\newblock


\bibitem[Lin et~al\mbox{.}(2019)]%
        {lin2019nesterov}
\bibfield{author}{\bibinfo{person}{Jiadong Lin}, \bibinfo{person}{Chuanbiao
  Song}, \bibinfo{person}{Kun He}, \bibinfo{person}{Liwei Wang}, {and}
  \bibinfo{person}{John~E Hopcroft}.} \bibinfo{year}{2019}\natexlab{}.
\newblock \showarticletitle{Nesterov Accelerated Gradient and Scale Invariance
  for Adversarial Attacks}. In \bibinfo{booktitle}{\emph{International
  Conference on Learning Representations}}.
\newblock


\bibitem[Liu et~al\mbox{.}(2020a)]%
        {liu2020bias}
\bibfield{author}{\bibinfo{person}{Aishan Liu}, \bibinfo{person}{Jiakai Wang},
  \bibinfo{person}{Xianglong Liu}, \bibinfo{person}{Bowen Cao},
  \bibinfo{person}{Chongzhi Zhang}, {and} \bibinfo{person}{Hang Yu}.}
  \bibinfo{year}{2020}\natexlab{a}.
\newblock \showarticletitle{Bias-based universal adversarial patch attack for
  automatic check-out}. In \bibinfo{booktitle}{\emph{European conference on
  computer vision}}. Springer, \bibinfo{pages}{395--410}.
\newblock


\bibitem[Liu et~al\mbox{.}(2022)]%
        {liu2022boosting}
\bibfield{author}{\bibinfo{person}{Binbin Liu}, \bibinfo{person}{Jinlai Zhang},
  {and} \bibinfo{person}{Jihong Zhu}.} \bibinfo{year}{2022}\natexlab{}.
\newblock \showarticletitle{Boosting 3D Adversarial Attacks with Attacking On
  Frequency}.
\newblock \bibinfo{journal}{\emph{IEEE Access}}  \bibinfo{volume}{10}
  (\bibinfo{year}{2022}), \bibinfo{pages}{50974--50984}.
\newblock


\bibitem[Liu and Hu(2022)]%
        {liu2022imperceptible}
\bibfield{author}{\bibinfo{person}{Daizong Liu} {and} \bibinfo{person}{Wei
  Hu}.} \bibinfo{year}{2022}\natexlab{}.
\newblock \showarticletitle{Imperceptible Transfer Attack and Defense on 3D
  Point Cloud Classification}.
\newblock \bibinfo{journal}{\emph{IEEE Transactions on Pattern Analysis and
  Machine Intelligence}} (\bibinfo{year}{2022}), \bibinfo{pages}{1--18}.
\newblock


\bibitem[Liu et~al\mbox{.}(2019)]%
        {liu2019extending}
\bibfield{author}{\bibinfo{person}{Daniel Liu}, \bibinfo{person}{Ronald Yu},
  {and} \bibinfo{person}{Hao Su}.} \bibinfo{year}{2019}\natexlab{}.
\newblock \showarticletitle{Extending adversarial attacks and defenses to deep
  3d point cloud classifiers}. In \bibinfo{booktitle}{\emph{2019 IEEE
  International Conference on Image Processing (ICIP)}}. IEEE,
  \bibinfo{pages}{2279--2283}.
\newblock


\bibitem[Liu et~al\mbox{.}(2020c)]%
        {liu2020adversarial}
\bibfield{author}{\bibinfo{person}{Daniel Liu}, \bibinfo{person}{Ronald Yu},
  {and} \bibinfo{person}{Hao Su}.} \bibinfo{year}{2020}\natexlab{c}.
\newblock \showarticletitle{Adversarial shape perturbations on 3d point
  clouds}. In \bibinfo{booktitle}{\emph{European Conference on Computer
  Vision}}. Springer, \bibinfo{pages}{88--104}.
\newblock


\bibitem[Liu et~al\mbox{.}(2018)]%
        {liu2018survey}
\bibfield{author}{\bibinfo{person}{Qiang Liu}, \bibinfo{person}{Pan Li},
  \bibinfo{person}{Wentao Zhao}, \bibinfo{person}{Wei Cai},
  \bibinfo{person}{Shui Yu}, {and} \bibinfo{person}{Victor~CM Leung}.}
  \bibinfo{year}{2018}\natexlab{}.
\newblock \showarticletitle{A survey on security threats and defensive
  techniques of machine learning: A data driven view}.
\newblock \bibinfo{journal}{\emph{IEEE access}}  \bibinfo{volume}{6}
  (\bibinfo{year}{2018}), \bibinfo{pages}{12103--12117}.
\newblock


\bibitem[Liu et~al\mbox{.}(2020b)]%
        {liu2020privacy}
\bibfield{author}{\bibinfo{person}{Ximeng Liu}, \bibinfo{person}{Lehui Xie},
  \bibinfo{person}{Yaopeng Wang}, \bibinfo{person}{Jian Zou},
  \bibinfo{person}{Jinbo Xiong}, \bibinfo{person}{Zuobin Ying}, {and}
  \bibinfo{person}{Athanasios~V Vasilakos}.} \bibinfo{year}{2020}\natexlab{b}.
\newblock \showarticletitle{Privacy and security issues in deep learning: A
  survey}.
\newblock \bibinfo{journal}{\emph{IEEE Access}}  \bibinfo{volume}{9}
  (\bibinfo{year}{2020}), \bibinfo{pages}{4566--4593}.
\newblock


\bibitem[Liu et~al\mbox{.}(2016)]%
        {liu2016delving}
\bibfield{author}{\bibinfo{person}{Yanpei Liu}, \bibinfo{person}{Xinyun Chen},
  \bibinfo{person}{Chang Liu}, {and} \bibinfo{person}{Dawn Song}.}
  \bibinfo{year}{2016}\natexlab{}.
\newblock \showarticletitle{Delving into Transferable Adversarial Examples and
  Black-box Attacks}. In \bibinfo{booktitle}{\emph{International Conference on
  Learning Representations}}.
\newblock


\bibitem[Luo et~al\mbox{.}(2018)]%
        {luo2018towards}
\bibfield{author}{\bibinfo{person}{Bo Luo}, \bibinfo{person}{Yannan Liu},
  \bibinfo{person}{Lingxiao Wei}, {and} \bibinfo{person}{Qiang Xu}.}
  \bibinfo{year}{2018}\natexlab{}.
\newblock \showarticletitle{Towards imperceptible and robust adversarial
  example attacks against neural networks}. In
  \bibinfo{booktitle}{\emph{Proceedings of the AAAI Conference on Artificial
  Intelligence}}, Vol.~\bibinfo{volume}{32}.
\newblock


\bibitem[Luo et~al\mbox{.}(2022)]%
        {luo2022frequency}
\bibfield{author}{\bibinfo{person}{Cheng Luo}, \bibinfo{person}{Qinliang Lin},
  \bibinfo{person}{Weicheng Xie}, \bibinfo{person}{Bizhu Wu},
  \bibinfo{person}{Jinheng Xie}, {and} \bibinfo{person}{Linlin Shen}.}
  \bibinfo{year}{2022}\natexlab{}.
\newblock \showarticletitle{Frequency-driven Imperceptible Adversarial Attack
  on Semantic Similarity}. In \bibinfo{booktitle}{\emph{Proceedings of the
  IEEE/CVF Conference on Computer Vision and Pattern Recognition}}.
  \bibinfo{pages}{15315--15324}.
\newblock


\bibitem[Ma et~al\mbox{.}(2021)]%
        {ma2021finding}
\bibfield{author}{\bibinfo{person}{Chen Ma}, \bibinfo{person}{Xiangyu Guo},
  \bibinfo{person}{Li Chen}, \bibinfo{person}{Jun-Hai Yong}, {and}
  \bibinfo{person}{Yisen Wang}.} \bibinfo{year}{2021}\natexlab{}.
\newblock \showarticletitle{Finding optimal tangent points for reducing
  distortions of hard-label attacks}.
\newblock \bibinfo{journal}{\emph{Advances in Neural Information Processing
  Systems}}  \bibinfo{volume}{34} (\bibinfo{year}{2021}),
  \bibinfo{pages}{19288--19300}.
\newblock


\bibitem[Ma et~al\mbox{.}(2020)]%
        {ma2020efficient}
\bibfield{author}{\bibinfo{person}{Chengcheng Ma}, \bibinfo{person}{Weiliang
  Meng}, \bibinfo{person}{Baoyuan Wu}, \bibinfo{person}{Shibiao Xu}, {and}
  \bibinfo{person}{Xiaopeng Zhang}.} \bibinfo{year}{2020}\natexlab{}.
\newblock \showarticletitle{Efficient joint gradient based attack against sor
  defense for 3d point cloud classification}. In
  \bibinfo{booktitle}{\emph{Proceedings of the 28th ACM International
  Conference on Multimedia}}. \bibinfo{pages}{1819--1827}.
\newblock


\bibitem[Machado et~al\mbox{.}(2021)]%
        {machado2021adversarial}
\bibfield{author}{\bibinfo{person}{Gabriel~Resende Machado},
  \bibinfo{person}{Eug{\^e}nio Silva}, {and} \bibinfo{person}{Ronaldo~Ribeiro
  Goldschmidt}.} \bibinfo{year}{2021}\natexlab{}.
\newblock \showarticletitle{Adversarial machine learning in image
  classification: a survey toward the defender’s perspective}.
\newblock \bibinfo{journal}{\emph{ACM Computing Surveys (CSUR)}}
  \bibinfo{volume}{55}, \bibinfo{number}{1} (\bibinfo{year}{2021}),
  \bibinfo{pages}{1--38}.
\newblock


\bibitem[Madry et~al\mbox{.}(2017)]%
        {madry2017towards}
\bibfield{author}{\bibinfo{person}{Aleksander Madry},
  \bibinfo{person}{Aleksandar Makelov}, \bibinfo{person}{Ludwig Schmidt},
  \bibinfo{person}{Dimitris Tsipras}, {and} \bibinfo{person}{Adrian Vladu}.}
  \bibinfo{year}{2017}\natexlab{}.
\newblock \showarticletitle{Towards deep learning models resistant to
  adversarial attacks}.
\newblock \bibinfo{journal}{\emph{arXiv preprint arXiv:1706.06083}}
  (\bibinfo{year}{2017}).
\newblock


\bibitem[Maho et~al\mbox{.}(2021)]%
        {maho2021surfree}
\bibfield{author}{\bibinfo{person}{Thibault Maho}, \bibinfo{person}{Teddy
  Furon}, {and} \bibinfo{person}{Erwan Le~Merrer}.}
  \bibinfo{year}{2021}\natexlab{}.
\newblock \showarticletitle{Surfree: a fast surrogate-free black-box attack}.
  In \bibinfo{booktitle}{\emph{Proceedings of the IEEE/CVF Conference on
  Computer Vision and Pattern Recognition}}. \bibinfo{pages}{10430--10439}.
\newblock


\bibitem[Miller et~al\mbox{.}(2020)]%
        {miller2020adversarial}
\bibfield{author}{\bibinfo{person}{David~J Miller}, \bibinfo{person}{Zhen
  Xiang}, {and} \bibinfo{person}{George Kesidis}.}
  \bibinfo{year}{2020}\natexlab{}.
\newblock \showarticletitle{Adversarial learning targeting deep neural network
  classification: A comprehensive review of defenses against attacks}.
\newblock \bibinfo{journal}{\emph{Proc. IEEE}} \bibinfo{volume}{108},
  \bibinfo{number}{3} (\bibinfo{year}{2020}), \bibinfo{pages}{402--433}.
\newblock


\bibitem[Modas et~al\mbox{.}(2019)]%
        {modas2019sparsefool}
\bibfield{author}{\bibinfo{person}{Apostolos Modas},
  \bibinfo{person}{Seyed-Mohsen Moosavi-Dezfooli}, {and}
  \bibinfo{person}{Pascal Frossard}.} \bibinfo{year}{2019}\natexlab{}.
\newblock \showarticletitle{Sparsefool: a few pixels make a big difference}. In
  \bibinfo{booktitle}{\emph{Proceedings of the IEEE/CVF conference on computer
  vision and pattern recognition}}. \bibinfo{pages}{9087--9096}.
\newblock


\bibitem[Moon et~al\mbox{.}(2019)]%
        {moon2019parsimonious}
\bibfield{author}{\bibinfo{person}{Seungyong Moon}, \bibinfo{person}{Gaon An},
  {and} \bibinfo{person}{Hyun~Oh Song}.} \bibinfo{year}{2019}\natexlab{}.
\newblock \showarticletitle{Parsimonious black-box adversarial attacks via
  efficient combinatorial optimization}. In
  \bibinfo{booktitle}{\emph{International Conference on Machine Learning}}.
  PMLR, \bibinfo{pages}{4636--4645}.
\newblock


\bibitem[Moosavi-Dezfooli et~al\mbox{.}(2017)]%
        {moosavi2017universal}
\bibfield{author}{\bibinfo{person}{Seyed-Mohsen Moosavi-Dezfooli},
  \bibinfo{person}{Alhussein Fawzi}, \bibinfo{person}{Omar Fawzi}, {and}
  \bibinfo{person}{Pascal Frossard}.} \bibinfo{year}{2017}\natexlab{}.
\newblock \showarticletitle{Universal adversarial perturbations}. In
  \bibinfo{booktitle}{\emph{Proceedings of the IEEE conference on computer
  vision and pattern recognition}}. \bibinfo{pages}{1765--1773}.
\newblock


\bibitem[Moosavi-Dezfooli et~al\mbox{.}(2016)]%
        {moosavi2016deepfool}
\bibfield{author}{\bibinfo{person}{Seyed-Mohsen Moosavi-Dezfooli},
  \bibinfo{person}{Alhussein Fawzi}, {and} \bibinfo{person}{Pascal Frossard}.}
  \bibinfo{year}{2016}\natexlab{}.
\newblock \showarticletitle{Deepfool: a simple and accurate method to fool deep
  neural networks}. In \bibinfo{booktitle}{\emph{Proceedings of the IEEE
  conference on computer vision and pattern recognition}}.
  \bibinfo{pages}{2574--2582}.
\newblock


\bibitem[Mopuri et~al\mbox{.}(2018a)]%
        {mopuri2018generalizable}
\bibfield{author}{\bibinfo{person}{Konda~Reddy Mopuri}, \bibinfo{person}{Aditya
  Ganeshan}, {and} \bibinfo{person}{R~Venkatesh Babu}.}
  \bibinfo{year}{2018}\natexlab{a}.
\newblock \showarticletitle{Generalizable data-free objective for crafting
  universal adversarial perturbations}.
\newblock \bibinfo{journal}{\emph{IEEE transactions on pattern analysis and
  machine intelligence}} \bibinfo{volume}{41}, \bibinfo{number}{10}
  (\bibinfo{year}{2018}), \bibinfo{pages}{2452--2465}.
\newblock


\bibitem[Mopuri et~al\mbox{.}(2018b)]%
        {mopuri2018nag}
\bibfield{author}{\bibinfo{person}{Konda~Reddy Mopuri},
  \bibinfo{person}{Utkarsh Ojha}, \bibinfo{person}{Utsav Garg}, {and}
  \bibinfo{person}{R~Venkatesh Babu}.} \bibinfo{year}{2018}\natexlab{b}.
\newblock \showarticletitle{Nag: Network for adversary generation}. In
  \bibinfo{booktitle}{\emph{Proceedings of the IEEE Conference on Computer
  Vision and Pattern Recognition}}. \bibinfo{pages}{742--751}.
\newblock


\bibitem[Narodytska and Kasiviswanathan(2017)]%
        {narodytska2017simple}
\bibfield{author}{\bibinfo{person}{Nina Narodytska} {and}
  \bibinfo{person}{Shiva~Prasad Kasiviswanathan}.}
  \bibinfo{year}{2017}\natexlab{}.
\newblock \showarticletitle{Simple Black-Box Adversarial Attacks on Deep Neural
  Networks.}. In \bibinfo{booktitle}{\emph{CVPR Workshops}},
  Vol.~\bibinfo{volume}{2}. \bibinfo{pages}{2}.
\newblock


\bibitem[Nguyen et~al\mbox{.}(2020)]%
        {nguyen2020adversarial}
\bibfield{author}{\bibinfo{person}{Dinh-Luan Nguyen},
  \bibinfo{person}{Sunpreet~S Arora}, \bibinfo{person}{Yuhang Wu}, {and}
  \bibinfo{person}{Hao Yang}.} \bibinfo{year}{2020}\natexlab{}.
\newblock \showarticletitle{Adversarial light projection attacks on face
  recognition systems: A feasibility study}. In
  \bibinfo{booktitle}{\emph{Proceedings of the IEEE/CVF conference on computer
  vision and pattern recognition workshops}}. \bibinfo{pages}{814--815}.
\newblock


\bibitem[Nie et~al\mbox{.}(2022)]%
        {nie2022diffusion}
\bibfield{author}{\bibinfo{person}{Weili Nie}, \bibinfo{person}{Brandon Guo},
  \bibinfo{person}{Yujia Huang}, \bibinfo{person}{Chaowei Xiao},
  \bibinfo{person}{Arash Vahdat}, {and} \bibinfo{person}{Anima Anandkumar}.}
  \bibinfo{year}{2022}\natexlab{}.
\newblock \showarticletitle{Diffusion models for adversarial purification}.
\newblock \bibinfo{journal}{\emph{arXiv preprint arXiv:2205.07460}}
  (\bibinfo{year}{2022}).
\newblock


\bibitem[Osadchy et~al\mbox{.}(2017)]%
        {Osadchy2017}
\bibfield{author}{\bibinfo{person}{Margarita Osadchy}, \bibinfo{person}{Julio
  Hernandez-Castro}, \bibinfo{person}{Stuart Gibson}, \bibinfo{person}{Orr
  Dunkelman}, {and} \bibinfo{person}{Daniel Pérez-Cabo}.}
  \bibinfo{year}{2017}\natexlab{}.
\newblock \showarticletitle{No Bot Expects the DeepCAPTCHA! Introducing
  Immutable Adversarial Examples, With Applications to CAPTCHA Generation}.
\newblock \bibinfo{journal}{\emph{IEEE Transactions on Information Forensics
  and Security}} \bibinfo{volume}{12}, \bibinfo{number}{11}
  (\bibinfo{year}{2017}), \bibinfo{pages}{2640--2653}.
\newblock


\bibitem[Papernot et~al\mbox{.}(2016a)]%
        {papernot2016transferability}
\bibfield{author}{\bibinfo{person}{Nicolas Papernot}, \bibinfo{person}{Patrick
  McDaniel}, {and} \bibinfo{person}{Ian Goodfellow}.}
  \bibinfo{year}{2016}\natexlab{a}.
\newblock \showarticletitle{Transferability in machine learning: from phenomena
  to black-box attacks using adversarial samples}.
\newblock \bibinfo{journal}{\emph{arXiv preprint arXiv:1605.07277}}
  (\bibinfo{year}{2016}).
\newblock


\bibitem[Papernot et~al\mbox{.}(2017)]%
        {papernot2017practical}
\bibfield{author}{\bibinfo{person}{Nicolas Papernot}, \bibinfo{person}{Patrick
  McDaniel}, \bibinfo{person}{Ian Goodfellow}, \bibinfo{person}{Somesh Jha},
  \bibinfo{person}{Z~Berkay Celik}, {and} \bibinfo{person}{Ananthram Swami}.}
  \bibinfo{year}{2017}\natexlab{}.
\newblock \showarticletitle{Practical black-box attacks against machine
  learning}. In \bibinfo{booktitle}{\emph{Proceedings of the 2017 ACM on Asia
  conference on computer and communications security}}.
  \bibinfo{pages}{506--519}.
\newblock


\bibitem[Papernot et~al\mbox{.}(2016b)]%
        {papernot2016limitations}
\bibfield{author}{\bibinfo{person}{Nicolas Papernot}, \bibinfo{person}{Patrick
  McDaniel}, \bibinfo{person}{Somesh Jha}, \bibinfo{person}{Matt Fredrikson},
  \bibinfo{person}{Z~Berkay Celik}, {and} \bibinfo{person}{Ananthram Swami}.}
  \bibinfo{year}{2016}\natexlab{b}.
\newblock \showarticletitle{The limitations of deep learning in adversarial
  settings}. In \bibinfo{booktitle}{\emph{2016 IEEE European symposium on
  security and privacy (EuroS\&P)}}. IEEE, \bibinfo{pages}{372--387}.
\newblock


\bibitem[Papernot et~al\mbox{.}(2018)]%
        {papernot2018sok}
\bibfield{author}{\bibinfo{person}{Nicolas Papernot}, \bibinfo{person}{Patrick
  McDaniel}, \bibinfo{person}{Arunesh Sinha}, {and} \bibinfo{person}{Michael~P
  Wellman}.} \bibinfo{year}{2018}\natexlab{}.
\newblock \showarticletitle{Sok: Security and privacy in machine learning}. In
  \bibinfo{booktitle}{\emph{2018 IEEE European Symposium on Security and
  Privacy (EuroS\&P)}}. IEEE, \bibinfo{pages}{399--414}.
\newblock


\bibitem[Papernot et~al\mbox{.}(2016c)]%
        {papernot2016distillation}
\bibfield{author}{\bibinfo{person}{Nicolas Papernot}, \bibinfo{person}{Patrick
  McDaniel}, \bibinfo{person}{Xi Wu}, \bibinfo{person}{Somesh Jha}, {and}
  \bibinfo{person}{Ananthram Swami}.} \bibinfo{year}{2016}\natexlab{c}.
\newblock \showarticletitle{Distillation as a defense to adversarial
  perturbations against deep neural networks}. In
  \bibinfo{booktitle}{\emph{2016 IEEE symposium on security and privacy (SP)}}.
  IEEE, \bibinfo{pages}{582--597}.
\newblock


\bibitem[Poursaeed et~al\mbox{.}(2018)]%
        {poursaeed2018generative}
\bibfield{author}{\bibinfo{person}{Omid Poursaeed}, \bibinfo{person}{Isay
  Katsman}, \bibinfo{person}{Bicheng Gao}, {and} \bibinfo{person}{Serge
  Belongie}.} \bibinfo{year}{2018}\natexlab{}.
\newblock \showarticletitle{Generative adversarial perturbations}. In
  \bibinfo{booktitle}{\emph{Proceedings of the IEEE Conference on Computer
  Vision and Pattern Recognition}}. \bibinfo{pages}{4422--4431}.
\newblock


\bibitem[Qi et~al\mbox{.}(2017)]%
        {qi2017pointnet}
\bibfield{author}{\bibinfo{person}{Charles~R Qi}, \bibinfo{person}{Hao Su},
  \bibinfo{person}{Kaichun Mo}, {and} \bibinfo{person}{Leonidas~J Guibas}.}
  \bibinfo{year}{2017}\natexlab{}.
\newblock \showarticletitle{Pointnet: Deep learning on point sets for 3d
  classification and segmentation}. In \bibinfo{booktitle}{\emph{Proceedings of
  the IEEE conference on computer vision and pattern recognition}}.
  \bibinfo{pages}{652--660}.
\newblock


\bibitem[Qiu et~al\mbox{.}(2020)]%
        {qiu2020semanticadv}
\bibfield{author}{\bibinfo{person}{Haonan Qiu}, \bibinfo{person}{Chaowei Xiao},
  \bibinfo{person}{Lei Yang}, \bibinfo{person}{Xinchen Yan},
  \bibinfo{person}{Honglak Lee}, {and} \bibinfo{person}{Bo Li}.}
  \bibinfo{year}{2020}\natexlab{}.
\newblock \showarticletitle{Semanticadv: Generating adversarial examples via
  attribute-conditioned image editing}. In \bibinfo{booktitle}{\emph{European
  Conference on Computer Vision}}. Springer, \bibinfo{pages}{19--37}.
\newblock


\bibitem[Rahmati et~al\mbox{.}(2020)]%
        {rahmati2020geoda}
\bibfield{author}{\bibinfo{person}{Ali Rahmati}, \bibinfo{person}{Seyed-Mohsen
  Moosavi-Dezfooli}, \bibinfo{person}{Pascal Frossard}, {and}
  \bibinfo{person}{Huaiyu Dai}.} \bibinfo{year}{2020}\natexlab{}.
\newblock \showarticletitle{Geoda: a geometric framework for black-box
  adversarial attacks}. In \bibinfo{booktitle}{\emph{Proceedings of the
  IEEE/CVF conference on computer vision and pattern recognition}}.
  \bibinfo{pages}{8446--8455}.
\newblock


\bibitem[Rakin et~al\mbox{.}(2021)]%
        {rakin2021t}
\bibfield{author}{\bibinfo{person}{Adnan~Siraj Rakin}, \bibinfo{person}{Zhezhi
  He}, \bibinfo{person}{Jingtao Li}, \bibinfo{person}{Fan Yao},
  \bibinfo{person}{Chaitali Chakrabarti}, {and} \bibinfo{person}{Deliang Fan}.}
  \bibinfo{year}{2021}\natexlab{}.
\newblock \showarticletitle{T-bfa: Targeted bit-flip adversarial weight
  attack}.
\newblock \bibinfo{journal}{\emph{IEEE Transactions on Pattern Analysis and
  Machine Intelligence}} (\bibinfo{year}{2021}).
\newblock


\bibitem[Reza et~al\mbox{.}(2023)]%
        {reza2023cgba}
\bibfield{author}{\bibinfo{person}{Md~Farhamdur Reza}, \bibinfo{person}{Ali
  Rahmati}, \bibinfo{person}{Tianfu Wu}, {and} \bibinfo{person}{Huaiyu Dai}.}
  \bibinfo{year}{2023}\natexlab{}.
\newblock \showarticletitle{CGBA: Curvature-aware Geometric Black-box Attack}.
\newblock \bibinfo{journal}{\emph{arXiv preprint arXiv:2308.03163}}
  (\bibinfo{year}{2023}).
\newblock


\bibitem[Rony et~al\mbox{.}(2019)]%
        {rony2019decoupling}
\bibfield{author}{\bibinfo{person}{J{\'e}r{\^o}me Rony},
  \bibinfo{person}{Luiz~G Hafemann}, \bibinfo{person}{Luiz~S Oliveira},
  \bibinfo{person}{Ismail~Ben Ayed}, \bibinfo{person}{Robert Sabourin}, {and}
  \bibinfo{person}{Eric Granger}.} \bibinfo{year}{2019}\natexlab{}.
\newblock \showarticletitle{Decoupling direction and norm for efficient
  gradient-based l2 adversarial attacks and defenses}. In
  \bibinfo{booktitle}{\emph{Proceedings of the IEEE/CVF Conference on Computer
  Vision and Pattern Recognition}}. \bibinfo{pages}{4322--4330}.
\newblock


\bibitem[Salem et~al\mbox{.}(2020)]%
        {salem2020updates}
\bibfield{author}{\bibinfo{person}{Ahmed Salem}, \bibinfo{person}{Apratim
  Bhattacharya}, \bibinfo{person}{Michael Backes}, \bibinfo{person}{Mario
  Fritz}, {and} \bibinfo{person}{Yang Zhang}.} \bibinfo{year}{2020}\natexlab{}.
\newblock \showarticletitle{$\{$Updates-Leak$\}$: Data Set Inference and
  Reconstruction Attacks in Online Learning}. In \bibinfo{booktitle}{\emph{29th
  USENIX security symposium (USENIX Security 20)}}.
  \bibinfo{pages}{1291--1308}.
\newblock


\bibitem[Sato et~al\mbox{.}(2021)]%
        {sato2021dirty}
\bibfield{author}{\bibinfo{person}{Takami Sato}, \bibinfo{person}{Junjie Shen},
  \bibinfo{person}{Ningfei Wang}, \bibinfo{person}{Yunhan Jia},
  \bibinfo{person}{Xue Lin}, {and} \bibinfo{person}{Qi~Alfred Chen}.}
  \bibinfo{year}{2021}\natexlab{}.
\newblock \showarticletitle{Dirty road can attack: Security of deep learning
  based automated lane centering under $\{$Physical-World$\}$ attack}. In
  \bibinfo{booktitle}{\emph{30th USENIX Security Symposium (USENIX Security
  21)}}. \bibinfo{pages}{3309--3326}.
\newblock


\bibitem[Sayles et~al\mbox{.}(2021)]%
        {sayles2021invisible}
\bibfield{author}{\bibinfo{person}{Athena Sayles}, \bibinfo{person}{Ashish
  Hooda}, \bibinfo{person}{Mohit Gupta}, \bibinfo{person}{Rahul Chatterjee},
  {and} \bibinfo{person}{Earlence Fernandes}.} \bibinfo{year}{2021}\natexlab{}.
\newblock \showarticletitle{Invisible perturbations: Physical adversarial
  examples exploiting the rolling shutter effect}. In
  \bibinfo{booktitle}{\emph{Proceedings of the IEEE/CVF Conference on Computer
  Vision and Pattern Recognition}}. \bibinfo{pages}{14666--14675}.
\newblock


\bibitem[Schmalfuss et~al\mbox{.}(2023)]%
        {schmalfuss2023distracting}
\bibfield{author}{\bibinfo{person}{Jenny Schmalfuss}, \bibinfo{person}{Lukas
  Mehl}, {and} \bibinfo{person}{Andr{\'e}s Bruhn}.}
  \bibinfo{year}{2023}\natexlab{}.
\newblock \showarticletitle{Distracting Downpour: Adversarial Weather Attacks
  for Motion Estimation}.
\newblock \bibinfo{journal}{\emph{arXiv preprint arXiv:2305.06716}}
  (\bibinfo{year}{2023}).
\newblock


\bibitem[Serban et~al\mbox{.}(2020)]%
        {serban2020adversarial}
\bibfield{author}{\bibinfo{person}{Alex Serban}, \bibinfo{person}{Erik Poll},
  {and} \bibinfo{person}{Joost Visser}.} \bibinfo{year}{2020}\natexlab{}.
\newblock \showarticletitle{Adversarial examples on object recognition: A
  comprehensive survey}.
\newblock \bibinfo{journal}{\emph{ACM Computing Surveys (CSUR)}}
  \bibinfo{volume}{53}, \bibinfo{number}{3} (\bibinfo{year}{2020}),
  \bibinfo{pages}{1--38}.
\newblock


\bibitem[Shamsabadi et~al\mbox{.}(2020a)]%
        {shamsabadi2020edgefool}
\bibfield{author}{\bibinfo{person}{Ali~Shahin Shamsabadi},
  \bibinfo{person}{Changjae Oh}, {and} \bibinfo{person}{Andrea Cavallaro}.}
  \bibinfo{year}{2020}\natexlab{a}.
\newblock \showarticletitle{EdgeFool: an adversarial image enhancement filter}.
  In \bibinfo{booktitle}{\emph{ICASSP 2020-2020 IEEE International Conference
  on Acoustics, Speech and Signal Processing (ICASSP)}}. IEEE,
  \bibinfo{pages}{1898--1902}.
\newblock


\bibitem[Shamsabadi et~al\mbox{.}(2021)]%
        {shamsabadi2021semantically}
\bibfield{author}{\bibinfo{person}{Ali~Shahin Shamsabadi},
  \bibinfo{person}{Changjae Oh}, {and} \bibinfo{person}{Andrea Cavallaro}.}
  \bibinfo{year}{2021}\natexlab{}.
\newblock \showarticletitle{Semantically adversarial learnable filters}.
\newblock \bibinfo{journal}{\emph{IEEE Transactions on Image Processing}}
  \bibinfo{volume}{30} (\bibinfo{year}{2021}), \bibinfo{pages}{8075--8087}.
\newblock


\bibitem[Shamsabadi et~al\mbox{.}(2020b)]%
        {shamsabadi2020colorfool}
\bibfield{author}{\bibinfo{person}{Ali~Shahin Shamsabadi},
  \bibinfo{person}{Ricardo Sanchez-Matilla}, {and} \bibinfo{person}{Andrea
  Cavallaro}.} \bibinfo{year}{2020}\natexlab{b}.
\newblock \showarticletitle{Colorfool: Semantic adversarial colorization}. In
  \bibinfo{booktitle}{\emph{Proceedings of the IEEE/CVF Conference on Computer
  Vision and Pattern Recognition}}. \bibinfo{pages}{1151--1160}.
\newblock


\bibitem[Sharif et~al\mbox{.}(2016)]%
        {sharif2016accessorize}
\bibfield{author}{\bibinfo{person}{Mahmood Sharif}, \bibinfo{person}{Sruti
  Bhagavatula}, \bibinfo{person}{Lujo Bauer}, {and} \bibinfo{person}{Michael~K
  Reiter}.} \bibinfo{year}{2016}\natexlab{}.
\newblock \showarticletitle{Accessorize to a crime: Real and stealthy attacks
  on state-of-the-art face recognition}. In
  \bibinfo{booktitle}{\emph{Proceedings of the 2016 acm sigsac conference on
  computer and communications security}}. \bibinfo{pages}{1528--1540}.
\newblock


\bibitem[Sharif et~al\mbox{.}(2019)]%
        {sharif2019general}
\bibfield{author}{\bibinfo{person}{Mahmood Sharif}, \bibinfo{person}{Sruti
  Bhagavatula}, \bibinfo{person}{Lujo Bauer}, {and} \bibinfo{person}{Michael~K
  Reiter}.} \bibinfo{year}{2019}\natexlab{}.
\newblock \showarticletitle{A general framework for adversarial examples with
  objectives}.
\newblock \bibinfo{journal}{\emph{ACM Transactions on Privacy and Security
  (TOPS)}} \bibinfo{volume}{22}, \bibinfo{number}{3} (\bibinfo{year}{2019}),
  \bibinfo{pages}{1--30}.
\newblock


\bibitem[Shen et~al\mbox{.}(2021)]%
        {shen2021effective}
\bibfield{author}{\bibinfo{person}{Meng Shen}, \bibinfo{person}{Hao Yu},
  \bibinfo{person}{Liehuang Zhu}, \bibinfo{person}{Ke Xu}, \bibinfo{person}{Qi
  Li}, {and} \bibinfo{person}{Jiankun Hu}.} \bibinfo{year}{2021}\natexlab{}.
\newblock \showarticletitle{Effective and robust physical-world attacks on deep
  learning face recognition systems}.
\newblock \bibinfo{journal}{\emph{IEEE Transactions on Information Forensics
  and Security}}  \bibinfo{volume}{16} (\bibinfo{year}{2021}),
  \bibinfo{pages}{4063--4077}.
\newblock


\bibitem[Shokri et~al\mbox{.}(2017)]%
        {shokri2017membership}
\bibfield{author}{\bibinfo{person}{Reza Shokri}, \bibinfo{person}{Marco
  Stronati}, \bibinfo{person}{Congzheng Song}, {and} \bibinfo{person}{Vitaly
  Shmatikov}.} \bibinfo{year}{2017}\natexlab{}.
\newblock \showarticletitle{Membership inference attacks against machine
  learning models}. In \bibinfo{booktitle}{\emph{2017 IEEE symposium on
  security and privacy (SP)}}. IEEE, \bibinfo{pages}{3--18}.
\newblock


\bibitem[Shukla et~al\mbox{.}(2021)]%
        {shukla2021simple}
\bibfield{author}{\bibinfo{person}{Satya~Narayan Shukla},
  \bibinfo{person}{Anit~Kumar Sahu}, \bibinfo{person}{Devin Willmott}, {and}
  \bibinfo{person}{Zico Kolter}.} \bibinfo{year}{2021}\natexlab{}.
\newblock \showarticletitle{Simple and efficient hard label black-box
  adversarial attacks in low query budget regimes}. In
  \bibinfo{booktitle}{\emph{Proceedings of the 27th ACM SIGKDD Conference on
  Knowledge Discovery \& Data Mining}}. \bibinfo{pages}{1461--1469}.
\newblock


\bibitem[Song et~al\mbox{.}(2018)]%
        {song2018constructing}
\bibfield{author}{\bibinfo{person}{Yang Song}, \bibinfo{person}{Rui Shu},
  \bibinfo{person}{Nate Kushman}, {and} \bibinfo{person}{Stefano Ermon}.}
  \bibinfo{year}{2018}\natexlab{}.
\newblock \showarticletitle{Constructing unrestricted adversarial examples with
  generative models}.
\newblock \bibinfo{journal}{\emph{Advances in Neural Information Processing
  Systems}}  \bibinfo{volume}{31} (\bibinfo{year}{2018}).
\newblock


\bibitem[Su et~al\mbox{.}(2019)]%
        {su2019one}
\bibfield{author}{\bibinfo{person}{Jiawei Su},
  \bibinfo{person}{Danilo~Vasconcellos Vargas}, {and} \bibinfo{person}{Kouichi
  Sakurai}.} \bibinfo{year}{2019}\natexlab{}.
\newblock \showarticletitle{One pixel attack for fooling deep neural networks}.
\newblock \bibinfo{journal}{\emph{IEEE Transactions on Evolutionary
  Computation}} \bibinfo{volume}{23}, \bibinfo{number}{5}
  (\bibinfo{year}{2019}), \bibinfo{pages}{828--841}.
\newblock


\bibitem[Sun et~al\mbox{.}(2020)]%
        {sun2020towards}
\bibfield{author}{\bibinfo{person}{Jiachen Sun}, \bibinfo{person}{Yulong Cao},
  \bibinfo{person}{Qi~Alfred Chen}, {and} \bibinfo{person}{Z~Morley Mao}.}
  \bibinfo{year}{2020}\natexlab{}.
\newblock \showarticletitle{Towards robust $\{$LiDAR-based$\}$ perception in
  autonomous driving: General black-box adversarial sensor attack and
  countermeasures}. In \bibinfo{booktitle}{\emph{29th USENIX Security Symposium
  (USENIX Security 20)}}. \bibinfo{pages}{877--894}.
\newblock


\bibitem[Suryanto et~al\mbox{.}(2022)]%
        {suryanto2022dta}
\bibfield{author}{\bibinfo{person}{Naufal Suryanto}, \bibinfo{person}{Yongsu
  Kim}, \bibinfo{person}{Hyoeun Kang}, \bibinfo{person}{Harashta~Tatimma
  Larasati}, \bibinfo{person}{Youngyeo Yun}, \bibinfo{person}{Thi-Thu-Huong
  Le}, \bibinfo{person}{Hunmin Yang}, \bibinfo{person}{Se-Yoon Oh}, {and}
  \bibinfo{person}{Howon Kim}.} \bibinfo{year}{2022}\natexlab{}.
\newblock \showarticletitle{DTA: Physical Camouflage Attacks using
  Differentiable Transformation Network}. In
  \bibinfo{booktitle}{\emph{Proceedings of the IEEE/CVF Conference on Computer
  Vision and Pattern Recognition}}. \bibinfo{pages}{15305--15314}.
\newblock


\bibitem[Szegedy et~al\mbox{.}(2013)]%
        {szegedy2013intriguing}
\bibfield{author}{\bibinfo{person}{Christian Szegedy},
  \bibinfo{person}{Wojciech Zaremba}, \bibinfo{person}{Ilya Sutskever},
  \bibinfo{person}{Joan Bruna}, \bibinfo{person}{Dumitru Erhan},
  \bibinfo{person}{Ian Goodfellow}, {and} \bibinfo{person}{Rob Fergus}.}
  \bibinfo{year}{2013}\natexlab{}.
\newblock \showarticletitle{Intriguing properties of neural networks}.
\newblock \bibinfo{journal}{\emph{arXiv preprint arXiv:1312.6199}}
  (\bibinfo{year}{2013}).
\newblock


\bibitem[Tao et~al\mbox{.}(2023)]%
        {tao20233dhacker}
\bibfield{author}{\bibinfo{person}{Yunbo Tao}, \bibinfo{person}{Daizong Liu},
  \bibinfo{person}{Pan Zhou}, \bibinfo{person}{Yulai Xie}, \bibinfo{person}{Wei
  Du}, {and} \bibinfo{person}{Wei Hu}.} \bibinfo{year}{2023}\natexlab{}.
\newblock \showarticletitle{3DHacker: Spectrum-based Decision Boundary
  Generation for Hard-label 3D Point Cloud Attack}.
\newblock \bibinfo{journal}{\emph{arXiv preprint arXiv:2308.07546}}
  (\bibinfo{year}{2023}).
\newblock


\bibitem[Tram{\`e}r et~al\mbox{.}(2017a)]%
        {tramer2017ensemble}
\bibfield{author}{\bibinfo{person}{Florian Tram{\`e}r}, \bibinfo{person}{Alexey
  Kurakin}, \bibinfo{person}{Nicolas Papernot}, \bibinfo{person}{Ian
  Goodfellow}, \bibinfo{person}{Dan Boneh}, {and} \bibinfo{person}{Patrick
  McDaniel}.} \bibinfo{year}{2017}\natexlab{a}.
\newblock \showarticletitle{Ensemble adversarial training: Attacks and
  defenses}.
\newblock \bibinfo{journal}{\emph{arXiv preprint arXiv:1705.07204}}
  (\bibinfo{year}{2017}).
\newblock


\bibitem[Tram{\`e}r et~al\mbox{.}(2017b)]%
        {tramer2017space}
\bibfield{author}{\bibinfo{person}{Florian Tram{\`e}r},
  \bibinfo{person}{Nicolas Papernot}, \bibinfo{person}{Ian Goodfellow},
  \bibinfo{person}{Dan Boneh}, {and} \bibinfo{person}{Patrick McDaniel}.}
  \bibinfo{year}{2017}\natexlab{b}.
\newblock \showarticletitle{The space of transferable adversarial examples}.
\newblock \bibinfo{journal}{\emph{arXiv preprint arXiv:1704.03453}}
  (\bibinfo{year}{2017}).
\newblock


\bibitem[Tsai et~al\mbox{.}(2020)]%
        {tsai2020robust}
\bibfield{author}{\bibinfo{person}{Tzungyu Tsai}, \bibinfo{person}{Kaichen
  Yang}, \bibinfo{person}{Tsung-Yi Ho}, {and} \bibinfo{person}{Yier Jin}.}
  \bibinfo{year}{2020}\natexlab{}.
\newblock \showarticletitle{Robust adversarial objects against deep learning
  models}. In \bibinfo{booktitle}{\emph{Proceedings of the AAAI Conference on
  Artificial Intelligence}}, Vol.~\bibinfo{volume}{34}.
  \bibinfo{pages}{954--962}.
\newblock


\bibitem[Tu et~al\mbox{.}(2019)]%
        {tu2019autozoom}
\bibfield{author}{\bibinfo{person}{Chun-Chen Tu}, \bibinfo{person}{Paishun
  Ting}, \bibinfo{person}{Pin-Yu Chen}, \bibinfo{person}{Sijia Liu},
  \bibinfo{person}{Huan Zhang}, \bibinfo{person}{Jinfeng Yi},
  \bibinfo{person}{Cho-Jui Hsieh}, {and} \bibinfo{person}{Shin-Ming Cheng}.}
  \bibinfo{year}{2019}\natexlab{}.
\newblock \showarticletitle{Autozoom: Autoencoder-based zeroth order
  optimization method for attacking black-box neural networks}. In
  \bibinfo{booktitle}{\emph{Proceedings of the AAAI Conference on Artificial
  Intelligence}}, Vol.~\bibinfo{volume}{33}. \bibinfo{pages}{742--749}.
\newblock


\bibitem[Tu et~al\mbox{.}(2020)]%
        {tu2020physically}
\bibfield{author}{\bibinfo{person}{James Tu}, \bibinfo{person}{Mengye Ren},
  \bibinfo{person}{Sivabalan Manivasagam}, \bibinfo{person}{Ming Liang},
  \bibinfo{person}{Bin Yang}, \bibinfo{person}{Richard Du},
  \bibinfo{person}{Frank Cheng}, {and} \bibinfo{person}{Raquel Urtasun}.}
  \bibinfo{year}{2020}\natexlab{}.
\newblock \showarticletitle{Physically realizable adversarial examples for
  lidar object detection}. In \bibinfo{booktitle}{\emph{Proceedings of the
  IEEE/CVF Conference on Computer Vision and Pattern Recognition}}.
  \bibinfo{pages}{13716--13725}.
\newblock


\bibitem[Vaswani et~al\mbox{.}(2017)]%
        {vaswani2017attention}
\bibfield{author}{\bibinfo{person}{Ashish Vaswani}, \bibinfo{person}{Noam
  Shazeer}, \bibinfo{person}{Niki Parmar}, \bibinfo{person}{Jakob Uszkoreit},
  \bibinfo{person}{Llion Jones}, \bibinfo{person}{Aidan~N Gomez},
  \bibinfo{person}{{\L}ukasz Kaiser}, {and} \bibinfo{person}{Illia
  Polosukhin}.} \bibinfo{year}{2017}\natexlab{}.
\newblock \showarticletitle{Attention is all you need}.
\newblock \bibinfo{journal}{\emph{Advances in neural information processing
  systems}}  \bibinfo{volume}{30} (\bibinfo{year}{2017}).
\newblock


\bibitem[Wang et~al\mbox{.}(2022a)]%
        {wang2022fca}
\bibfield{author}{\bibinfo{person}{Donghua Wang}, \bibinfo{person}{Tingsong
  Jiang}, \bibinfo{person}{Jialiang Sun}, \bibinfo{person}{Weien Zhou},
  \bibinfo{person}{Zhiqiang Gong}, \bibinfo{person}{Xiaoya Zhang},
  \bibinfo{person}{Wen Yao}, {and} \bibinfo{person}{Xiaoqian Chen}.}
  \bibinfo{year}{2022}\natexlab{a}.
\newblock \showarticletitle{Fca: Learning a 3d full-coverage vehicle camouflage
  for multi-view physical adversarial attack}. In
  \bibinfo{booktitle}{\emph{Proceedings of the AAAI Conference on Artificial
  Intelligence}}, Vol.~\bibinfo{volume}{36}. \bibinfo{pages}{2414--2422}.
\newblock


\bibitem[Wang et~al\mbox{.}(2021b)]%
        {wang2021dual}
\bibfield{author}{\bibinfo{person}{Jiakai Wang}, \bibinfo{person}{Aishan Liu},
  \bibinfo{person}{Zixin Yin}, \bibinfo{person}{Shunchang Liu},
  \bibinfo{person}{Shiyu Tang}, {and} \bibinfo{person}{Xianglong Liu}.}
  \bibinfo{year}{2021}\natexlab{b}.
\newblock \showarticletitle{Dual attention suppression attack: Generate
  adversarial camouflage in physical world}. In
  \bibinfo{booktitle}{\emph{Proceedings of the IEEE/CVF Conference on Computer
  Vision and Pattern Recognition}}. \bibinfo{pages}{8565--8574}.
\newblock


\bibitem[Wang and He(2021)]%
        {wang2021enhancing}
\bibfield{author}{\bibinfo{person}{Xiaosen Wang} {and} \bibinfo{person}{Kun
  He}.} \bibinfo{year}{2021}\natexlab{}.
\newblock \showarticletitle{Enhancing the transferability of adversarial
  attacks through variance tuning}. In \bibinfo{booktitle}{\emph{Proceedings of
  the IEEE/CVF Conference on Computer Vision and Pattern Recognition}}.
  \bibinfo{pages}{1924--1933}.
\newblock


\bibitem[Wang et~al\mbox{.}(2022b)]%
        {wang2022triangle}
\bibfield{author}{\bibinfo{person}{Xiaosen Wang}, \bibinfo{person}{Zeliang
  Zhang}, \bibinfo{person}{Kangheng Tong}, \bibinfo{person}{Dihong Gong},
  \bibinfo{person}{Kun He}, \bibinfo{person}{Zhifeng Li}, {and}
  \bibinfo{person}{Wei Liu}.} \bibinfo{year}{2022}\natexlab{b}.
\newblock \showarticletitle{Triangle attack: A query-efficient decision-based
  adversarial attack}. In \bibinfo{booktitle}{\emph{European Conference on
  Computer Vision}}. Springer, \bibinfo{pages}{156--174}.
\newblock


\bibitem[Wang et~al\mbox{.}(2019)]%
        {wang2019dynamic}
\bibfield{author}{\bibinfo{person}{Yue Wang}, \bibinfo{person}{Yongbin Sun},
  \bibinfo{person}{Ziwei Liu}, \bibinfo{person}{Sanjay~E Sarma},
  \bibinfo{person}{Michael~M Bronstein}, {and} \bibinfo{person}{Justin~M
  Solomon}.} \bibinfo{year}{2019}\natexlab{}.
\newblock \showarticletitle{Dynamic graph cnn for learning on point clouds}.
\newblock \bibinfo{journal}{\emph{Acm Transactions On Graphics (tog)}}
  \bibinfo{volume}{38}, \bibinfo{number}{5} (\bibinfo{year}{2019}),
  \bibinfo{pages}{1--12}.
\newblock


\bibitem[Wang et~al\mbox{.}(2021c)]%
        {wang2021demiguise}
\bibfield{author}{\bibinfo{person}{Yajie Wang}, \bibinfo{person}{Shangbo Wu},
  \bibinfo{person}{Wenyi Jiang}, \bibinfo{person}{Shengang Hao},
  \bibinfo{person}{Yu-an Tan}, {and} \bibinfo{person}{Quanxin Zhang}.}
  \bibinfo{year}{2021}\natexlab{c}.
\newblock \showarticletitle{Demiguise attack: Crafting invisible semantic
  adversarial perturbations with perceptual similarity}.
\newblock \bibinfo{journal}{\emph{arXiv preprint arXiv:2107.01396}}
  (\bibinfo{year}{2021}).
\newblock


\bibitem[Wang et~al\mbox{.}(2021a)]%
        {wang2021feature}
\bibfield{author}{\bibinfo{person}{Zhibo Wang}, \bibinfo{person}{Hengchang
  Guo}, \bibinfo{person}{Zhifei Zhang}, \bibinfo{person}{Wenxin Liu},
  \bibinfo{person}{Zhan Qin}, {and} \bibinfo{person}{Kui Ren}.}
  \bibinfo{year}{2021}\natexlab{a}.
\newblock \showarticletitle{Feature importance-aware transferable adversarial
  attacks}. In \bibinfo{booktitle}{\emph{Proceedings of the IEEE/CVF
  International Conference on Computer Vision}}. \bibinfo{pages}{7639--7648}.
\newblock


\bibitem[Wei et~al\mbox{.}(2022)]%
        {wei2022cross}
\bibfield{author}{\bibinfo{person}{Zhipeng Wei}, \bibinfo{person}{Jingjing
  Chen}, \bibinfo{person}{Zuxuan Wu}, {and} \bibinfo{person}{Yu-Gang Jiang}.}
  \bibinfo{year}{2022}\natexlab{}.
\newblock \showarticletitle{Cross-Modal Transferable Adversarial Attacks from
  Images to Videos}. In \bibinfo{booktitle}{\emph{Proceedings of the IEEE/CVF
  Conference on Computer Vision and Pattern Recognition}}.
  \bibinfo{pages}{15064--15073}.
\newblock


\bibitem[Wen et~al\mbox{.}(2020)]%
        {wen2020geometry}
\bibfield{author}{\bibinfo{person}{Yuxin Wen}, \bibinfo{person}{Jiehong Lin},
  \bibinfo{person}{Ke Chen}, \bibinfo{person}{CL~Philip Chen}, {and}
  \bibinfo{person}{Kui Jia}.} \bibinfo{year}{2020}\natexlab{}.
\newblock \showarticletitle{Geometry-aware generation of adversarial point
  clouds}.
\newblock \bibinfo{journal}{\emph{IEEE Transactions on Pattern Analysis and
  Machine Intelligence}} (\bibinfo{year}{2020}).
\newblock


\bibitem[Wicker and Kwiatkowska(2019)]%
        {wicker2019robustness}
\bibfield{author}{\bibinfo{person}{Matthew Wicker} {and} \bibinfo{person}{Marta
  Kwiatkowska}.} \bibinfo{year}{2019}\natexlab{}.
\newblock \showarticletitle{Robustness of 3d deep learning in an adversarial
  setting}. In \bibinfo{booktitle}{\emph{Proceedings of the IEEE/CVF Conference
  on Computer Vision and Pattern Recognition}}. \bibinfo{pages}{11767--11775}.
\newblock


\bibitem[Wong et~al\mbox{.}(2019)]%
        {wong2019wasserstein}
\bibfield{author}{\bibinfo{person}{Eric Wong}, \bibinfo{person}{Frank Schmidt},
  {and} \bibinfo{person}{Zico Kolter}.} \bibinfo{year}{2019}\natexlab{}.
\newblock \showarticletitle{Wasserstein adversarial examples via projected
  sinkhorn iterations}. In \bibinfo{booktitle}{\emph{International Conference
  on Machine Learning}}. PMLR, \bibinfo{pages}{6808--6817}.
\newblock


\bibitem[Wu et~al\mbox{.}(2018)]%
        {wu2018understanding}
\bibfield{author}{\bibinfo{person}{Lei Wu}, \bibinfo{person}{Zhanxing Zhu},
  \bibinfo{person}{Cheng Tai}, {et~al\mbox{.}}}
  \bibinfo{year}{2018}\natexlab{}.
\newblock \showarticletitle{Understanding and enhancing the transferability of
  adversarial examples}.
\newblock \bibinfo{journal}{\emph{arXiv preprint arXiv:1802.09707}}
  (\bibinfo{year}{2018}).
\newblock


\bibitem[Wu et~al\mbox{.}(2021)]%
        {wu2021improving}
\bibfield{author}{\bibinfo{person}{Weibin Wu}, \bibinfo{person}{Yuxin Su},
  \bibinfo{person}{Michael~R Lyu}, {and} \bibinfo{person}{Irwin King}.}
  \bibinfo{year}{2021}\natexlab{}.
\newblock \showarticletitle{Improving the transferability of adversarial
  samples with adversarial transformations}. In
  \bibinfo{booktitle}{\emph{Proceedings of the IEEE/CVF Conference on Computer
  Vision and Pattern Recognition}}. \bibinfo{pages}{9024--9033}.
\newblock


\bibitem[Xiao et~al\mbox{.}(2018a)]%
        {xiao2018generating}
\bibfield{author}{\bibinfo{person}{Chaowei Xiao}, \bibinfo{person}{Bo Li},
  \bibinfo{person}{Jun~Yan Zhu}, \bibinfo{person}{Warren He},
  \bibinfo{person}{Mingyan Liu}, {and} \bibinfo{person}{Dawn Song}.}
  \bibinfo{year}{2018}\natexlab{a}.
\newblock \showarticletitle{Generating adversarial examples with adversarial
  networks}. In \bibinfo{booktitle}{\emph{27th International Joint Conference
  on Artificial Intelligence, IJCAI 2018}}. International Joint Conferences on
  Artificial Intelligence, \bibinfo{pages}{3905--3911}.
\newblock


\bibitem[Xiao et~al\mbox{.}(2019)]%
        {xiao2019meshadv}
\bibfield{author}{\bibinfo{person}{Chaowei Xiao}, \bibinfo{person}{Dawei Yang},
  \bibinfo{person}{Bo Li}, \bibinfo{person}{Jia Deng}, {and}
  \bibinfo{person}{Mingyan Liu}.} \bibinfo{year}{2019}\natexlab{}.
\newblock \showarticletitle{Meshadv: Adversarial meshes for visual
  recognition}. In \bibinfo{booktitle}{\emph{Proceedings of the IEEE/CVF
  Conference on Computer Vision and Pattern Recognition}}.
  \bibinfo{pages}{6898--6907}.
\newblock


\bibitem[Xiao et~al\mbox{.}(2018b)]%
        {xiao2018spatially}
\bibfield{author}{\bibinfo{person}{Chaowei Xiao}, \bibinfo{person}{Jun~Yan
  Zhu}, \bibinfo{person}{Bo Li}, \bibinfo{person}{Warren He},
  \bibinfo{person}{Mingyan Liu}, {and} \bibinfo{person}{Dawn Song}.}
  \bibinfo{year}{2018}\natexlab{b}.
\newblock \showarticletitle{Spatially transformed adversarial examples}. In
  \bibinfo{booktitle}{\emph{6th International Conference on Learning
  Representations, ICLR 2018}}.
\newblock


\bibitem[Xiao et~al\mbox{.}(2021)]%
        {xiao2021improving}
\bibfield{author}{\bibinfo{person}{Zihao Xiao}, \bibinfo{person}{Xianfeng Gao},
  \bibinfo{person}{Chilin Fu}, \bibinfo{person}{Yinpeng Dong},
  \bibinfo{person}{Wei Gao}, \bibinfo{person}{Xiaolu Zhang},
  \bibinfo{person}{Jun Zhou}, {and} \bibinfo{person}{Jun Zhu}.}
  \bibinfo{year}{2021}\natexlab{}.
\newblock \showarticletitle{Improving transferability of adversarial patches on
  face recognition with generative models}. In
  \bibinfo{booktitle}{\emph{Proceedings of the IEEE/CVF Conference on Computer
  Vision and Pattern Recognition}}. \bibinfo{pages}{11845--11854}.
\newblock


\bibitem[Xie et~al\mbox{.}(2017)]%
        {xie2017adversarial}
\bibfield{author}{\bibinfo{person}{Cihang Xie}, \bibinfo{person}{Jianyu Wang},
  \bibinfo{person}{Zhishuai Zhang}, \bibinfo{person}{Yuyin Zhou},
  \bibinfo{person}{Lingxi Xie}, {and} \bibinfo{person}{Alan Yuille}.}
  \bibinfo{year}{2017}\natexlab{}.
\newblock \showarticletitle{Adversarial examples for semantic segmentation and
  object detection}. In \bibinfo{booktitle}{\emph{Proceedings of the IEEE
  international conference on computer vision}}. \bibinfo{pages}{1369--1378}.
\newblock


\bibitem[Xie et~al\mbox{.}(2019)]%
        {xie2019improving}
\bibfield{author}{\bibinfo{person}{Cihang Xie}, \bibinfo{person}{Zhishuai
  Zhang}, \bibinfo{person}{Yuyin Zhou}, \bibinfo{person}{Song Bai},
  \bibinfo{person}{Jianyu Wang}, \bibinfo{person}{Zhou Ren}, {and}
  \bibinfo{person}{Alan~L Yuille}.} \bibinfo{year}{2019}\natexlab{}.
\newblock \showarticletitle{Improving transferability of adversarial examples
  with input diversity}. In \bibinfo{booktitle}{\emph{Proceedings of the
  IEEE/CVF Conference on Computer Vision and Pattern Recognition}}.
  \bibinfo{pages}{2730--2739}.
\newblock


\bibitem[Yang et~al\mbox{.}(2020)]%
        {yang2020learning}
\bibfield{author}{\bibinfo{person}{Jiancheng Yang}, \bibinfo{person}{Yangzhou
  Jiang}, \bibinfo{person}{Xiaoyang Huang}, \bibinfo{person}{Bingbing Ni},
  {and} \bibinfo{person}{Chenglong Zhao}.} \bibinfo{year}{2020}\natexlab{}.
\newblock \showarticletitle{Learning black-box attackers with transferable
  priors and query feedback}.
\newblock \bibinfo{journal}{\emph{Advances in Neural Information Processing
  Systems}}  \bibinfo{volume}{33} (\bibinfo{year}{2020}),
  \bibinfo{pages}{12288--12299}.
\newblock


\bibitem[Yin et~al\mbox{.}(2021)]%
        {yin2021adv}
\bibfield{author}{\bibinfo{person}{Bangjie Yin}, \bibinfo{person}{Wenxuan
  Wang}, \bibinfo{person}{Taiping Yao}, \bibinfo{person}{Junfeng Guo},
  \bibinfo{person}{Zelun Kong}, \bibinfo{person}{Shouhong Ding},
  \bibinfo{person}{Jilin Li}, {and} \bibinfo{person}{Cong Liu}.}
  \bibinfo{year}{2021}\natexlab{}.
\newblock \showarticletitle{Adv-Makeup: A New Imperceptible and Transferable
  Attack on Face Recognition}.
\newblock \bibinfo{journal}{\emph{International Joint Conferences on Artificial
  Intelligence (IJCAI)}} (\bibinfo{year}{2021}).
\newblock


\bibitem[Yu et~al\mbox{.}(2022)]%
        {yu2022availability}
\bibfield{author}{\bibinfo{person}{Da Yu}, \bibinfo{person}{Huishuai Zhang},
  \bibinfo{person}{Wei Chen}, \bibinfo{person}{Jian Yin}, {and}
  \bibinfo{person}{Tie-Yan Liu}.} \bibinfo{year}{2022}\natexlab{}.
\newblock \showarticletitle{Availability attacks create shortcuts}. In
  \bibinfo{booktitle}{\emph{Proceedings of the 28th ACM SIGKDD Conference on
  Knowledge Discovery and Data Mining}}. \bibinfo{pages}{2367--2376}.
\newblock


\bibitem[Zeng et~al\mbox{.}(2019)]%
        {zeng2019adversarial}
\bibfield{author}{\bibinfo{person}{Xiaohui Zeng}, \bibinfo{person}{Chenxi Liu},
  \bibinfo{person}{Yu-Siang Wang}, \bibinfo{person}{Weichao Qiu},
  \bibinfo{person}{Lingxi Xie}, \bibinfo{person}{Yu-Wing Tai},
  \bibinfo{person}{Chi-Keung Tang}, {and} \bibinfo{person}{Alan~L Yuille}.}
  \bibinfo{year}{2019}\natexlab{}.
\newblock \showarticletitle{Adversarial attacks beyond the image space}. In
  \bibinfo{booktitle}{\emph{Proceedings of the IEEE/CVF Conference on Computer
  Vision and Pattern Recognition}}. \bibinfo{pages}{4302--4311}.
\newblock


\bibitem[Zhang et~al\mbox{.}(2020b)]%
        {zhang2020understanding}
\bibfield{author}{\bibinfo{person}{Chaoning Zhang}, \bibinfo{person}{Philipp
  Benz}, \bibinfo{person}{Tooba Imtiaz}, {and} \bibinfo{person}{In~So Kweon}.}
  \bibinfo{year}{2020}\natexlab{b}.
\newblock \showarticletitle{Understanding adversarial examples from the mutual
  influence of images and perturbations}. In
  \bibinfo{booktitle}{\emph{Proceedings of the IEEE/CVF Conference on Computer
  Vision and Pattern Recognition}}. \bibinfo{pages}{14521--14530}.
\newblock


\bibitem[Zhang et~al\mbox{.}(2021)]%
        {zhang2021universal}
\bibfield{author}{\bibinfo{person}{Chaoning Zhang}, \bibinfo{person}{Philipp
  Benz}, \bibinfo{person}{Adil Karjauv}, {and} \bibinfo{person}{In~So Kweon}.}
  \bibinfo{year}{2021}\natexlab{}.
\newblock \showarticletitle{Universal adversarial perturbations through the
  lens of deep steganography: Towards a fourier perspective}. In
  \bibinfo{booktitle}{\emph{Proceedings of the AAAI Conference on Artificial
  Intelligence}}, Vol.~\bibinfo{volume}{35}. \bibinfo{pages}{3296--3304}.
\newblock


\bibitem[Zhang et~al\mbox{.}(2020a)]%
        {zhang2020smooth}
\bibfield{author}{\bibinfo{person}{Hanwei Zhang}, \bibinfo{person}{Yannis
  Avrithis}, \bibinfo{person}{Teddy Furon}, {and} \bibinfo{person}{Laurent
  Amsaleg}.} \bibinfo{year}{2020}\natexlab{a}.
\newblock \showarticletitle{Smooth adversarial examples}.
\newblock \bibinfo{journal}{\emph{EURASIP Journal on Information Security}}
  \bibinfo{volume}{2020}, \bibinfo{number}{1} (\bibinfo{year}{2020}),
  \bibinfo{pages}{1--12}.
\newblock


\bibitem[Zhang et~al\mbox{.}(2018a)]%
        {zhang2018limitations}
\bibfield{author}{\bibinfo{person}{Huan Zhang}, \bibinfo{person}{Hongge Chen},
  \bibinfo{person}{Zhao Song}, \bibinfo{person}{Duane Boning},
  \bibinfo{person}{Inderjit~S Dhillon}, {and} \bibinfo{person}{Cho-Jui Hsieh}.}
  \bibinfo{year}{2018}\natexlab{a}.
\newblock \showarticletitle{The Limitations of Adversarial Training and the
  Blind-Spot Attack}. In \bibinfo{booktitle}{\emph{International Conference on
  Learning Representations}}.
\newblock


\bibitem[Zhang et~al\mbox{.}(2022b)]%
        {zhang2022improving}
\bibfield{author}{\bibinfo{person}{Jianping Zhang}, \bibinfo{person}{Weibin
  Wu}, \bibinfo{person}{Jen-tse Huang}, \bibinfo{person}{Yizhan Huang},
  \bibinfo{person}{Wenxuan Wang}, \bibinfo{person}{Yuxin Su}, {and}
  \bibinfo{person}{Michael~R Lyu}.} \bibinfo{year}{2022}\natexlab{b}.
\newblock \showarticletitle{Improving Adversarial Transferability via Neuron
  Attribution-Based Attacks}. In \bibinfo{booktitle}{\emph{Proceedings of the
  IEEE/CVF Conference on Computer Vision and Pattern Recognition}}.
  \bibinfo{pages}{14993--15002}.
\newblock


\bibitem[Zhang et~al\mbox{.}(2022a)]%
        {zhang2022adversarial}
\bibfield{author}{\bibinfo{person}{Qingzhao Zhang}, \bibinfo{person}{Shengtuo
  Hu}, \bibinfo{person}{Jiachen Sun}, \bibinfo{person}{Qi~Alfred Chen}, {and}
  \bibinfo{person}{Z~Morley Mao}.} \bibinfo{year}{2022}\natexlab{a}.
\newblock \showarticletitle{On adversarial robustness of trajectory prediction
  for autonomous vehicles}. In \bibinfo{booktitle}{\emph{Proceedings of the
  IEEE/CVF Conference on Computer Vision and Pattern Recognition}}.
  \bibinfo{pages}{15159--15168}.
\newblock


\bibitem[Zhang et~al\mbox{.}(2016)]%
        {zhang2016colorful}
\bibfield{author}{\bibinfo{person}{Richard Zhang}, \bibinfo{person}{Phillip
  Isola}, {and} \bibinfo{person}{Alexei~A Efros}.}
  \bibinfo{year}{2016}\natexlab{}.
\newblock \showarticletitle{Colorful image colorization}. In
  \bibinfo{booktitle}{\emph{European conference on computer vision}}. Springer,
  \bibinfo{pages}{649--666}.
\newblock


\bibitem[Zhang et~al\mbox{.}(2018b)]%
        {zhang2018camou}
\bibfield{author}{\bibinfo{person}{Yang Zhang}, \bibinfo{person}{Hassan
  Foroosh}, \bibinfo{person}{Philip David}, {and} \bibinfo{person}{Boqing
  Gong}.} \bibinfo{year}{2018}\natexlab{b}.
\newblock \showarticletitle{CAMOU: Learning physical vehicle camouflages to
  adversarially attack detectors in the wild}. In
  \bibinfo{booktitle}{\emph{International Conference on Learning
  Representations}}.
\newblock


\bibitem[Zhao et~al\mbox{.}(2021a)]%
        {zhao2021point}
\bibfield{author}{\bibinfo{person}{Hengshuang Zhao}, \bibinfo{person}{Li
  Jiang}, \bibinfo{person}{Jiaya Jia}, \bibinfo{person}{Philip~HS Torr}, {and}
  \bibinfo{person}{Vladlen Koltun}.} \bibinfo{year}{2021}\natexlab{a}.
\newblock \showarticletitle{Point transformer}. In
  \bibinfo{booktitle}{\emph{Proceedings of the IEEE/CVF international
  conference on computer vision}}. \bibinfo{pages}{16259--16268}.
\newblock


\bibitem[Zhao et~al\mbox{.}(2019)]%
        {zhao2019perturbations}
\bibfield{author}{\bibinfo{person}{He Zhao}, \bibinfo{person}{Trung Le},
  \bibinfo{person}{Paul Montague}, \bibinfo{person}{Olivier De~Vel},
  \bibinfo{person}{Tamas Abraham}, {and} \bibinfo{person}{Dinh Phung}.}
  \bibinfo{year}{2019}\natexlab{}.
\newblock \showarticletitle{Perturbations are not enough: Generating
  adversarial examples with spatial distortions}.
\newblock \bibinfo{journal}{\emph{arXiv preprint arXiv:1910.01329}}
  (\bibinfo{year}{2019}).
\newblock


\bibitem[Zhao et~al\mbox{.}(2020b)]%
        {zhao2020isometry}
\bibfield{author}{\bibinfo{person}{Yue Zhao}, \bibinfo{person}{Yuwei Wu},
  \bibinfo{person}{Caihua Chen}, {and} \bibinfo{person}{Andrew Lim}.}
  \bibinfo{year}{2020}\natexlab{b}.
\newblock \showarticletitle{On isometry robustness of deep 3d point cloud
  models under adversarial attacks}. In \bibinfo{booktitle}{\emph{Proceedings
  of the IEEE/CVF Conference on Computer Vision and Pattern Recognition}}.
  \bibinfo{pages}{1201--1210}.
\newblock


\bibitem[Zhao et~al\mbox{.}(2018)]%
        {zhao2018generating}
\bibfield{author}{\bibinfo{person}{Zhengli Zhao}, \bibinfo{person}{Dheeru Dua},
  {and} \bibinfo{person}{Sameer Singh}.} \bibinfo{year}{2018}\natexlab{}.
\newblock \showarticletitle{Generating Natural Adversarial Examples}. In
  \bibinfo{booktitle}{\emph{International Conference on Learning
  Representations}}.
\newblock


\bibitem[Zhao et~al\mbox{.}(2020a)]%
        {zhao2020towards}
\bibfield{author}{\bibinfo{person}{Zhengyu Zhao}, \bibinfo{person}{Zhuoran
  Liu}, {and} \bibinfo{person}{Martha Larson}.}
  \bibinfo{year}{2020}\natexlab{a}.
\newblock \showarticletitle{Towards large yet imperceptible adversarial image
  perturbations with perceptual color distance}. In
  \bibinfo{booktitle}{\emph{Proceedings of the IEEE/CVF Conference on Computer
  Vision and Pattern Recognition}}. \bibinfo{pages}{1039--1048}.
\newblock


\bibitem[Zhao et~al\mbox{.}(2021b)]%
        {zhao2021success}
\bibfield{author}{\bibinfo{person}{Zhengyu Zhao}, \bibinfo{person}{Zhuoran
  Liu}, {and} \bibinfo{person}{Martha Larson}.}
  \bibinfo{year}{2021}\natexlab{b}.
\newblock \showarticletitle{On success and simplicity: A second look at
  transferable targeted attacks}.
\newblock \bibinfo{journal}{\emph{Advances in Neural Information Processing
  Systems}}  \bibinfo{volume}{34} (\bibinfo{year}{2021}),
  \bibinfo{pages}{6115--6128}.
\newblock


\bibitem[Zheng et~al\mbox{.}(2019a)]%
        {zheng2019distributionally}
\bibfield{author}{\bibinfo{person}{Tianhang Zheng}, \bibinfo{person}{Changyou
  Chen}, {and} \bibinfo{person}{Kui Ren}.} \bibinfo{year}{2019}\natexlab{a}.
\newblock \showarticletitle{Distributionally adversarial attack}. In
  \bibinfo{booktitle}{\emph{Proceedings of the AAAI Conference on Artificial
  Intelligence}}, Vol.~\bibinfo{volume}{33}. \bibinfo{pages}{2253--2260}.
\newblock


\bibitem[Zheng et~al\mbox{.}(2019b)]%
        {zheng2019pointcloud}
\bibfield{author}{\bibinfo{person}{Tianhang Zheng}, \bibinfo{person}{Changyou
  Chen}, \bibinfo{person}{Junsong Yuan}, \bibinfo{person}{Bo Li}, {and}
  \bibinfo{person}{Kui Ren}.} \bibinfo{year}{2019}\natexlab{b}.
\newblock \showarticletitle{Pointcloud saliency maps}. In
  \bibinfo{booktitle}{\emph{Proceedings of the IEEE/CVF International
  Conference on Computer Vision}}. \bibinfo{pages}{1598--1606}.
\newblock


\bibitem[Zhou et~al\mbox{.}(2020)]%
        {zhou2020lg}
\bibfield{author}{\bibinfo{person}{Hang Zhou}, \bibinfo{person}{Dongdong Chen},
  \bibinfo{person}{Jing Liao}, \bibinfo{person}{Kejiang Chen},
  \bibinfo{person}{Xiaoyi Dong}, \bibinfo{person}{Kunlin Liu},
  \bibinfo{person}{Weiming Zhang}, \bibinfo{person}{Gang Hua}, {and}
  \bibinfo{person}{Nenghai Yu}.} \bibinfo{year}{2020}\natexlab{}.
\newblock \showarticletitle{Lg-gan: Label guided adversarial network for
  flexible targeted attack of point cloud based deep networks}. In
  \bibinfo{booktitle}{\emph{Proceedings of the IEEE/CVF Conference on Computer
  Vision and Pattern Recognition}}. \bibinfo{pages}{10356--10365}.
\newblock


\bibitem[Zhou et~al\mbox{.}(2018)]%
        {zhou2018transferable}
\bibfield{author}{\bibinfo{person}{Wen Zhou}, \bibinfo{person}{Xin Hou},
  \bibinfo{person}{Yongjun Chen}, \bibinfo{person}{Mengyun Tang},
  \bibinfo{person}{Xiangqi Huang}, \bibinfo{person}{Xiang Gan}, {and}
  \bibinfo{person}{Yong Yang}.} \bibinfo{year}{2018}\natexlab{}.
\newblock \showarticletitle{Transferable adversarial perturbations}. In
  \bibinfo{booktitle}{\emph{Proceedings of the European Conference on Computer
  Vision (ECCV)}}. \bibinfo{pages}{452--467}.
\newblock


\bibitem[Zhuang et~al\mbox{.}(2023)]%
        {zhuang2023pilot}
\bibfield{author}{\bibinfo{person}{Haomin Zhuang}, \bibinfo{person}{Yihua
  Zhang}, {and} \bibinfo{person}{Sijia Liu}.} \bibinfo{year}{2023}\natexlab{}.
\newblock \showarticletitle{A pilot study of query-free adversarial attack
  against stable diffusion}. In \bibinfo{booktitle}{\emph{Proceedings of the
  IEEE/CVF Conference on Computer Vision and Pattern Recognition}}.
  \bibinfo{pages}{2384--2391}.
\newblock


\bibitem[Zou et~al\mbox{.}(2020)]%
        {zou2020improving}
\bibfield{author}{\bibinfo{person}{Junhua Zou}, \bibinfo{person}{Zhisong Pan},
  \bibinfo{person}{Junyang Qiu}, \bibinfo{person}{Xin Liu},
  \bibinfo{person}{Ting Rui}, {and} \bibinfo{person}{Wei Li}.}
  \bibinfo{year}{2020}\natexlab{}.
\newblock \showarticletitle{Improving the transferability of adversarial
  examples with resized-diverse-inputs, diversity-ensemble and region fitting}.
  In \bibinfo{booktitle}{\emph{European Conference on Computer Vision}}.
  Springer, \bibinfo{pages}{563--579}.
\newblock


\end{thebibliography}

\end{document}